\documentclass[10pt,twocolumn,letterpaper]{article}
\usepackage[pagenumbers]{cvpr} 
\usepackage{graphicx}
\usepackage{amsmath}
\usepackage{amssymb}
\usepackage{bm}
\usepackage{booktabs}
\usepackage{xcolor}
\usepackage{pifont}% http://ctan.org/pkg/pifont
\usepackage{overpic}
\usepackage{tabu}
\usepackage{url}
\usepackage[normalem]{ulem}
\usepackage{afterpage}
\usepackage{booktabs}
\usepackage{color}
\usepackage{verbatim}
\usepackage{soul}
\usepackage{xspace}
\usepackage{comment}
\usepackage{array}
\usepackage{multirow}
\usepackage{url}
\usepackage{epigraph}
\usepackage[toc,page]{appendix}
\usepackage{microtype}
\usepackage[pagebackref,breaklinks,colorlinks]{hyperref}

\newcommand{\cmark}{\ding{51}}
\newcommand{\xmark}{\ding{55}}

\usepackage[capitalize]{cleveref}
\crefname{section}{Sec.}{Secs.}
\Crefname{section}{Section}{Sections}
\Crefname{table}{Table}{Tables}
\crefname{table}{Tab.}{Tabs.}

\def\cvprPaperID{*****} % *** Enter the CVPR Paper ID here
\def\confName{CVPR}
\def\confYear{2022}

\begin{document}

%%%%%%%%% TITLE - PLEASE UPDATE
\title{Video-Specific Autoencoders for Exploring, Editing and Transmitting Videos}
\author{Kevin Wang \quad\quad Deva Ramanan \quad\quad Aayush Bansal\\
Carnegie Mellon University\\
{\tt \url{https://www.cs.cmu.edu/~aayushb/Video-ViSA/}}
}
\maketitle

%%%%%%%%% ABSTRACT
\begin{abstract}

We study video-specific autoencoders that allow a human user to explore, edit, and efficiently transmit videos. Prior work has independently looked at these problems (and sub-problems) and proposed different formulations. In this work, we train a simple autoencoder (from scratch) on multiple frames of a specific video. We observe: (1) latent codes learned by a video-specific autoencoder capture spatial and temporal properties of that video; and (2) autoencoders can project out-of-sample inputs onto the video-specific manifold. These two properties allow us to explore, edit, and efficiently transmit a video using one learned representation. For e.g., linear operations on latent codes allow users to visualize the contents of a video. Associating latent codes of a video and manifold projection enables users to make desired edits. Interpolating latent codes and manifold projection allows the transmission of sparse low-res frames over a network.

\end{abstract}

%%%%%%%%% Introduction
\section{Introduction}

\begin{figure*}[t]
\centering
\includegraphics[width=\linewidth]{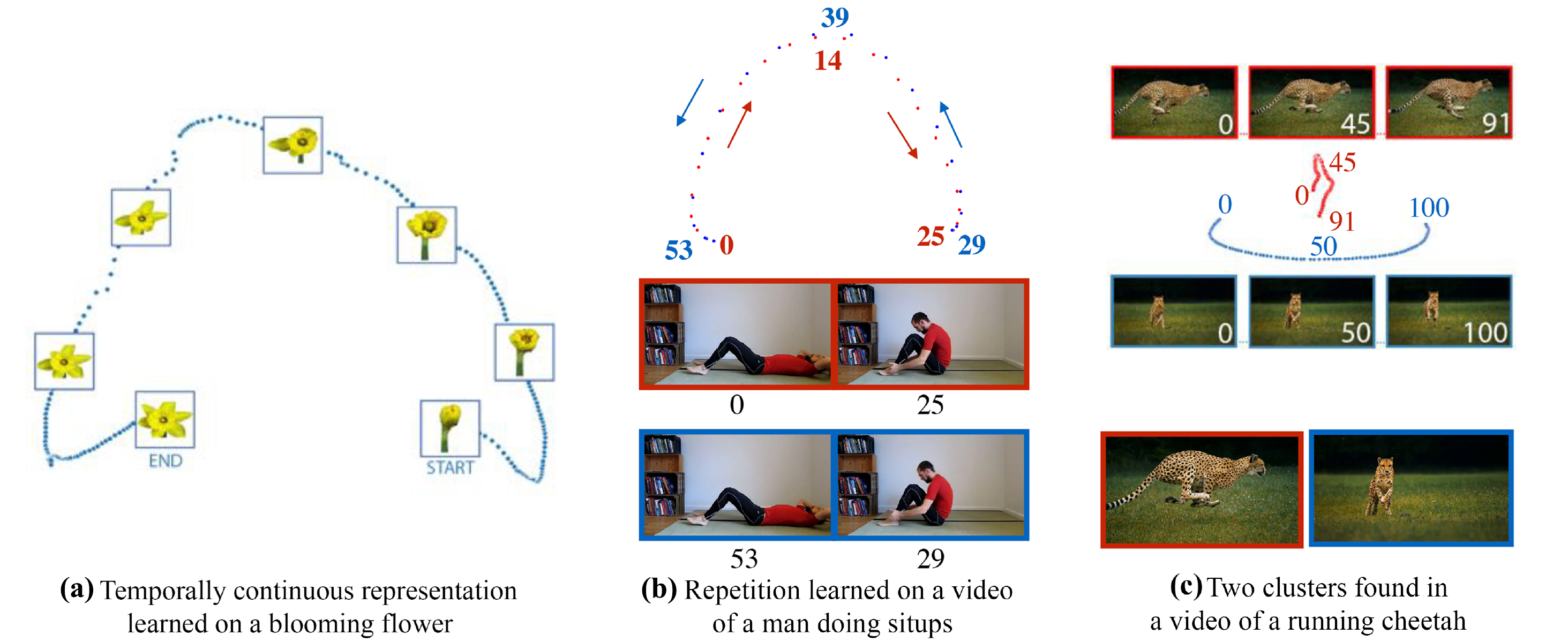}
\caption{\textbf{Representations learned without temporal information using autoencoders: } In each figure, a dot represents the latent code of a frame from a video that the autoencoder is trained on. The latent code is reduced from a high dimension to a 2-dimensional visualisation using PCA. \textbf{(a)} We find that video-specific autoencoders learn a temporally continuous representation without any explicit temporal information, for example on a video of a blooming flower. \textbf{(b)} We observe that latent spaces are able to learn the repetitive motion without using any temporal information, such as the repetitive motion of a situp. \textbf{(c)} Finally, distinct modes emerge when the autoencoder encounters different visual concepts in a video. For this cheetah example, we see that the two modes represent two different running poses in the video. We can also show the average image of each mode, as seen below the example.}
\label{fig:cont}
\end{figure*}

In this work, we demonstrate that simple video-specific autoencoders learn meaningful representations that enable a multitude of video processing tasks, without being optimized for any specific task. An autoencoder trained using individual frames (without any temporal information) of a specific video via a simple reconstruction loss can learn both spatial and temporal aspects of the video. Simple operations on its latent code, encoder, and decoder enables a wide variety of tasks including video exploration, video editing and video transmission.
To the best of our knowledge, we are the first to explore such diverse video processing tasks using a single representation not optimized for any specific task.

\textbf{Contributions}: (1) We introduce a simple, unsupervised approach for learning video-specific exemplar representations without needing large training data. This representation enables us to do a wide variety of the aforementioned video processing tasks that generally require a dedicated approach. (2) Our approach allows for intuitive user interaction, via a low dimensional visualization of latent codes that allow for video exploration and editing; and (3) finally, we demonstrate that the latent codes and manifold created using a single autoencoder allows for the transmission of sparse, low-resolution frames over a network with the ability to reconstruct the hi-res video using the autoencoder.

%%%%%%%%% Related Work
%\subsection{Videos: Opportunities and Conundrums}
\section{Related Work}
\label{sec:background}
There is a large body of work on specialized video processing tasks such as video completion~\cite{Gao-ECCV-FGVC}, video enhancement~\cite{xue2019video}, video inpainting~\cite{chang2019learnable,Huang-SigAsia-2016,Xu_2019_CVPR}, video editing~\cite{BSTSPP14,newson2013towards},  temporal super-resolution~\cite{jiang2018super,liu2017voxelflow,mahajan2009moving}, spatial super-resolution~\cite{haris2019recurrent,xiang2020zooming}, space-time super-resolution~\cite{shechtman2002increasing,shechtman2005space}, removing obstructions~\cite{Liu-CVPR-2020}, varying speed of a video~\cite{benaim2020speednet} or the humans in it~\cite{lu2020layered}, video textures~\cite{Agarwala:2005,Liu:2019,schodl2000video}, finding unintentional events in a video~\cite{Epstein_2020_CVPR}, video prediction~\cite{walker2016uncertain,walker2017pose},  generative modeling for associating two videos~\cite{Recycle-GAN}
discriminative modeling for association~\cite{purushwalkam2020aligning,wang2019learning,Dwibedi_2019_CVPR}, associating multi-view videos~\cite{vo2016spatiotemporal}, or pixel-level correspondences in a video~\cite{baker2011database,yang2019volumetric}. In this work, rather than exploring specialized architectures, we learn a single video-specific representation that  enables many of these tasks. 

\noindent\textbf{Learning from a Single Instance: }  There is plethora of work that has explored representation learned using a single image for various tasks~\cite{bahat2016blind,glasner2009super,michaeli2014blind,shocher2018zero,ulyanov2020deep,wexler2007space}. Recent approaches~\cite{shaham2019singan,shocher2018ingan} have also explored image-specific representation that enables a wide variety of image editing tasks. Often there exists repetitive structure in a signal, such as patch-recurrence in an image~\cite{bahat2016blind}, that allows one to learn meaningful representations for that signal without any additional information. We extend these observations to videos. Our goal is to learn a representation for a video without any additional information. Because we have more informative data, we could learn a simple autoencoder that optimizes the reconstruction of individual video frames. Prior work on video processing~\cite{Recycle-GAN,Gao-ECCV-FGVC,jiang2018super} often encodes spatial-temporal information explicitly. In this work, we considered frames from a video as independent images. Despite this, our video-specific autoencoder learns a continuous representation as shown in Figure~\ref{fig:cont}-(a).

\noindent\textbf{Human-Controllable Representation: } Usually the front-end of an application is designed around the task of interest. For example, prior work~\cite{bahat2020explorable,Fried:2019,zhu2014averageExplorer} on user-control are limited to a task. In this work, we hope to provide users with a simple representation that they can easily work with to design new application without much overhead (such as simple algebraic operations on the latent codes). Our approach also allows a user to see the contents of a video in a glance. Figure~\ref{fig:cont}-(c) shows how a user can explore the contents of a video and see two prominent aspects of this video. Our work also aims at reducing the requirement of application-dependent modules. For e.g., object removal approaches~\cite{Gao-ECCV-FGVC} usually require a user to provide extensive object-level mask across the video or use an off-the-shelf segmentation module trained for a particular object. Our learned representation allows us to both: (1) track the object in the video when marked in a single frame by a user and (2) edit the content.

%%%%%%%%% Method
\section{Video-Specific Autoencoders}
\label{sec:method}
\begin{figure*}[t]
\centering
\includegraphics[width=\linewidth]{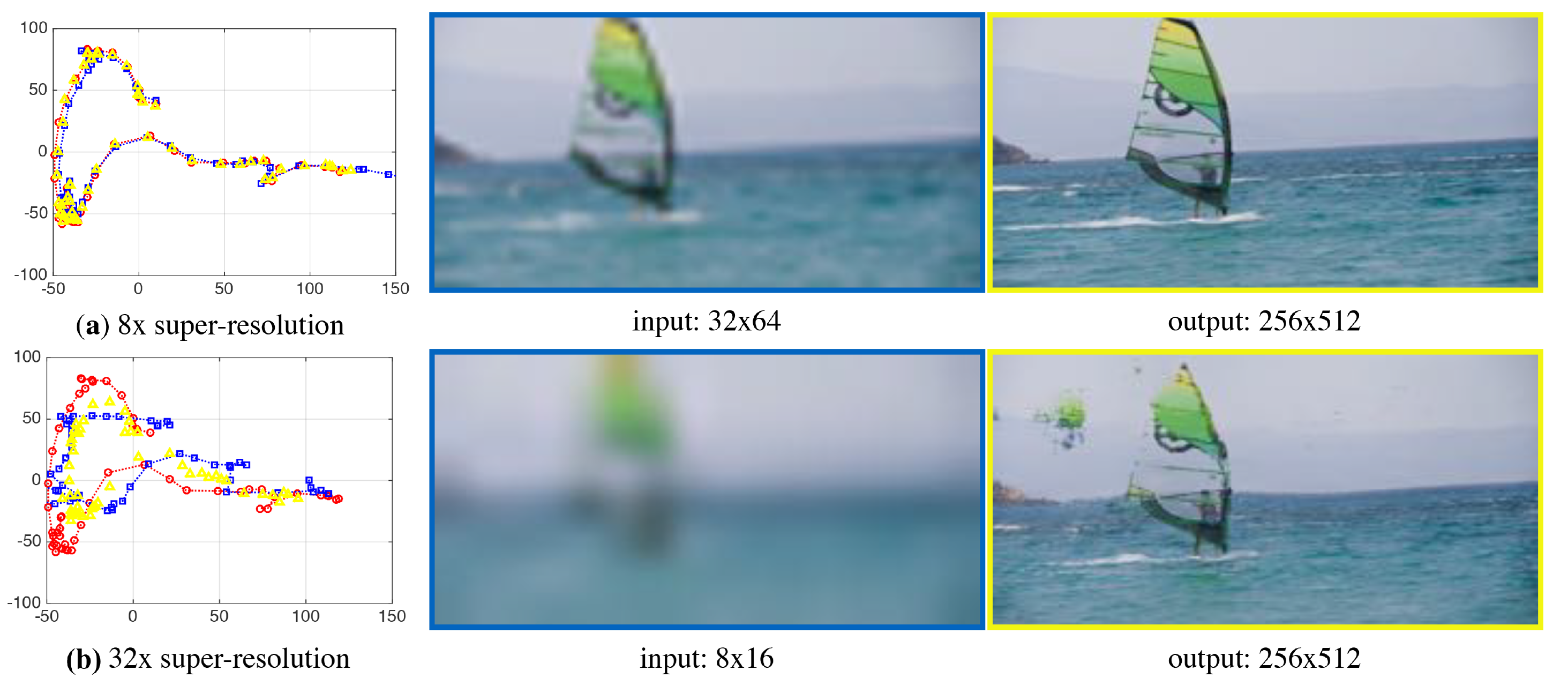}
\caption{\textbf{Manifold Projection with autoencoders}: We train a video-specific autoencoder on $55$ high-res $256\times512$ frames. We visualize the latent codes for the original frames using the {\color{red}{red}} points on the left side. At test time, we input low-res versions of {\em held-out} frames of varying resolution. We visualize the latent codes for the low-res input using the {\color{blue}{blue}} points. We also visualize the latent codes for the output image using the {\color{orange}{yellow}} points. We observe that {\color{red}{red}} points, {\color{blue}{blue}} points, and {\color{orange}{yellow}} points for \textbf{(a)} 8X super-resolution. We observe perfect reconstruction for this input. The results, however, degrade as we further reduce the resolution of images, as seen in \textbf{(b)} 32X super-resolution.}
\label{fig:sres-analysis-02}
\end{figure*}

\begin{figure*}[t]
\centering
\includegraphics[width=\linewidth]{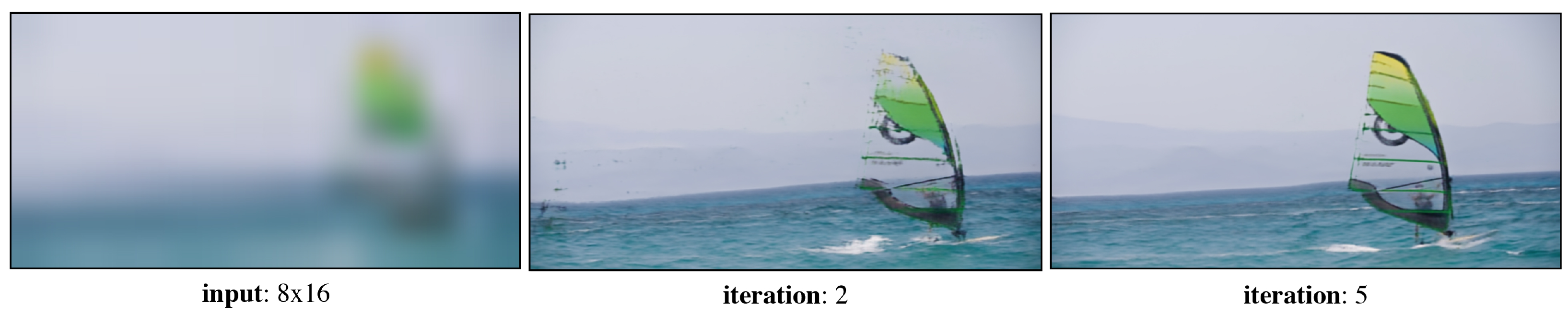}
\caption{\textbf{Iterative Improvement via Reprojection Property}: %We take the same example as in Figure ~\ref{fig:sres-analysis-02}. 
Given a $8\times16$ image, we iteratively improve the quality of the output. The reprojection property allows us to move towards a good solution with every iteration. At the end of the fifth iteration, we get a sharp hi-res ($256\times512$) output despite beginning with an extremely low-res input. We show that iterative projection can lead to quantitatively better exploration and transmission of videos. }
\label{fig:sres-analysis-03}
\end{figure*}

In this section, we review basic properties of autoencoders that we will exploit for various video editing tasks. Recall that an autoencoder~\cite{goodfellow2016deep} compresses the information in a signal via an encoding function.  The compressed signal or latent codes are represented using a few bits of information. Given a set of frames from a video $x \in V$, video-specific autoencoders learn to encode each frame into a low-dimensional latent code $f(x)$ that can be decoded (via a function $g$) back into the high-dimensional input space, so as to minimize the reconstruction error:
\begin{align}
 \min_{f,g} \sum_{x \in V} ||x - g(f(x))||^2 \label{eq:ae}
\end{align}

We use a convolutional feed-forward model that inputs an image and reconstructs it. Importantly, we ensure all operations are convolutional, implying that the size of the latent code scales with the resolution of the video $V$. To ensure that the latent code contains all the information needed to reconstruct a video frame, we do not use skip connections.

\noindent\textbf{Convolutional Encoder ($f$): }The encoder consists of six 2D convolutional layers. We use $5\times5$ kernels for first four layers and a stride of $2$ that downsamples the input by $0.5$ after each convolution. The last two layers have $5\times5$ kernels without any downsampling. The output of each of these layers is max-pooled in a $2\times2$ region with a stride of $2$. Each conv-layer is followed by  batch-normalization~\cite{ioffe2015batch} and a ReLU activation function~\cite{krizhevsky2017imagenet}. The output of last layer of the encoder is used as a latent representation (or also termed as latent code) in this work.

\noindent\textbf{Convolutional Decoder ($g$): } The decoder inputs the latent code to reconstruct the output. It consists of six up-sampling conv-layers with a $4\times4$ kernels and a stride of $2$ that upsamples the input by $2$. Each conv-layer is followed by  batch-normalization and a ReLu activation function.

\noindent\textbf{Latent Codes ($f(x)$):} Given an input image $x$ with shape h$\times$w, the encoder outputs an encoding with the shape $(k * 12) \times \frac{h}{64} \times \frac{w}{64}$, where $k$ is the number of filters in the first layer of encoder. We later show that such codes can be intuitively visualized with low-dimensional projections (via 2D PCA). For most experiments, we fix $k = 64$. This implies that for an input frame of size $h=256$ by $w=512$, latent codes compress inputs by a factor of 16.

\noindent\textbf{Continuous Temporal Spaces: } We study the space of latent codes $f(x)$ to examine the impact of temporal variation of input frames $x_t$, since the autoencoder is learned without any explicit temporal input. We visualize the latent code space via multidimensional scaling with PCA~\cite{bishop2006pattern}, as seen in examples in Figure~\ref{fig:cont}-(a).  An autoencoder trained on a specific video implicitly learns the correlations in the various frames and a continuous temporal space emerges. This property allows us to slow-down or speed-up a video (through latent code resampling). Latent codes can also capture repetitive motion as shown in Figure~\ref{fig:cont}-(b). This property allows us to temporally edit the video.

\begin{figure*}[t]
\includegraphics[width=\linewidth]{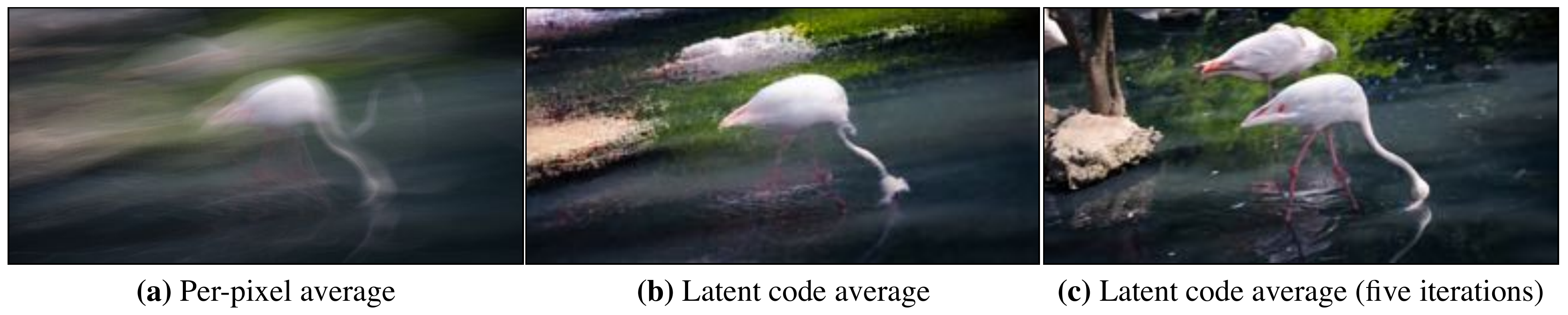}
\caption{\textbf{Video Averages}: We compare a per-pixel average of video frames {\bf (a)} with the decoded average of the latent codes of individual video frames {\bf (b)}, which is far less blurry. We use the iterative reprojection property to sharpen the average image {\bf (c)}.} %\deva{Could add closet training image to right, {\em if} it looks different.}}
\label{fig:avg}
\end{figure*}

\noindent{\bf Video-Specific Manifold ($M$):} We define the {\em manifold} of an autoencoder to be the set of all possible output reconstructions obtainable with {\em any} input:
\begin{align}
M =\{g(f(x)): \forall x\} .%\quad \text{where $f,g$ = argmin of (1)}
%    M = \{g(f(x)): \forall x}\} \quad \text{where} \quad %\text{fg = argmin of \ref{eq:ae}
\end{align}

where $f,g$ are the ``argmin" encoder and decoder learned from \eqref{eq:ae}. In our setting, $M$ corresponds to be a video-specific manifold of potential image/frame reconstructions. It is well-known that feedforward autoencoders, when properly trained, act as projection operators that project out-of-sample inputs $x$ into the manifold set $M$~\cite{goodfellow2016deep}. 
\begin{align}
    ||g(f(x)) - x||^2 &= \min_{m \in M} ||m - x||^2, \quad \forall x\\
    &= \min_{x'} ||g(f(x')) - x||^2, \quad \forall x.
\end{align}
One can build intuition for above by appealing to linear autoencoders, which can be learned with PCA. In this case, the above equations point out the well-known fact that PCA projects out-of-sample inputs into the closet point in the linear subspace spanned by the training data~\cite{bishop2006pattern}. We make use of this property to ``denoise" noisy input images $x \not \in V$ into manifold $M$ (where noisy inputs can consist of e.g., blurry frames).

\noindent\textbf{Manifold Reprojection: } The reprojection property enables the model to map noisy inputs $x$ onto the video-specific manifold $M$ spanned by the training set. We write this as:
\begin{align}
   \text{Project}_0(x) &= g(f(x)).
\end{align}
Figure~\ref{fig:sres-analysis-02} shows that 8X-downsampled inputs can be effectively upsampled by reprojecting a blurry bilinearly-upsampled input $x$ onto the video manifold. While results are nearly perfect for 8X upsampling, reprojected outputs contain visual artifacts for more agressive downsampled inputs.

\noindent\textbf{Iterative Reprojection: } One can iteratively reproject outputs with an autoencoder: 
\begin{align}
    \text{Project}_n(x) &= g(f(\text{Project}_{n-1}(x))).
 \end{align}
For linear autoencoders, one can show that this iteration converges after one step: $\text{Project}_n(x) = \text{Project}_0(x), \forall n$ because a single linear projection ensures the output falls within the linear subspace of the training data~\cite{bishop2006pattern}. For nonlinear autoenoders, convergence may take longer and is not necessarily guaranteed~\cite{goodfellow2016deep}.
Figure~\ref{fig:sres-analysis-03} shows that one can iteratively improve the quality of results for 32X super-resolution when inputting a $8\times16$ image. %Over the iterations, the autoencoder results in a plausible hi-res output. 
Because it is not guaranteed to converge, we use a fixed $n=5$ in our experiments unless otherwise noted. 

\noindent\textbf{Multi-Video Manifolds:} Because our autoencoders are learned with collections of frames, they can easily be trained on frames from multiple videos. We show in our applications that such shared latent representations can be used to compare, cluster, and process collections of videos.

\noindent\textbf{Pixel Codes ($f_i(x)$):} Finally, we can use our encoder to extract {\em pixel}-level representations, similar to past work that extracts such representations from classification networks~\cite{pixelnet, hariharan2015hypercolumns}. Conceptually, one can resize each of the six convolutional layers of the encoder $f$ back to the original image input size, and then extract out the ``hyper"-column of features aligned with pixel $i$. In practice, one can extract features from the image-level encoder $f(x)$ without resizing through judicious bookkeeping. Our final pixel representation, written as $f_i(x)$ is $2176$ dimensional.

\noindent{\bf Training Details:} We train a video-specific autoenocder from scratch using the Adam solver~\cite{kingma2014adam} with a batch-size of $6$. The learning rate is kept constant to $0.0002$ for first $100$ epochs and then linearly decayed to zero over the next $100$ epochs. For larger videos (i.e., more than $3,000$ frames), we reduce the number of epochs to $40$. 

We now study various applications of autoencoders to explore (Section~\ref{sec:explore}), edit (Section~\ref{sec:edit}), and efficiently transmit videos (Section~\ref{sec:transmit}). 

%%%%%%%%% Properties
\section{Exploring Videos}
\label{sec:explore}

\begin{figure*}[t]
\includegraphics[width=\linewidth]{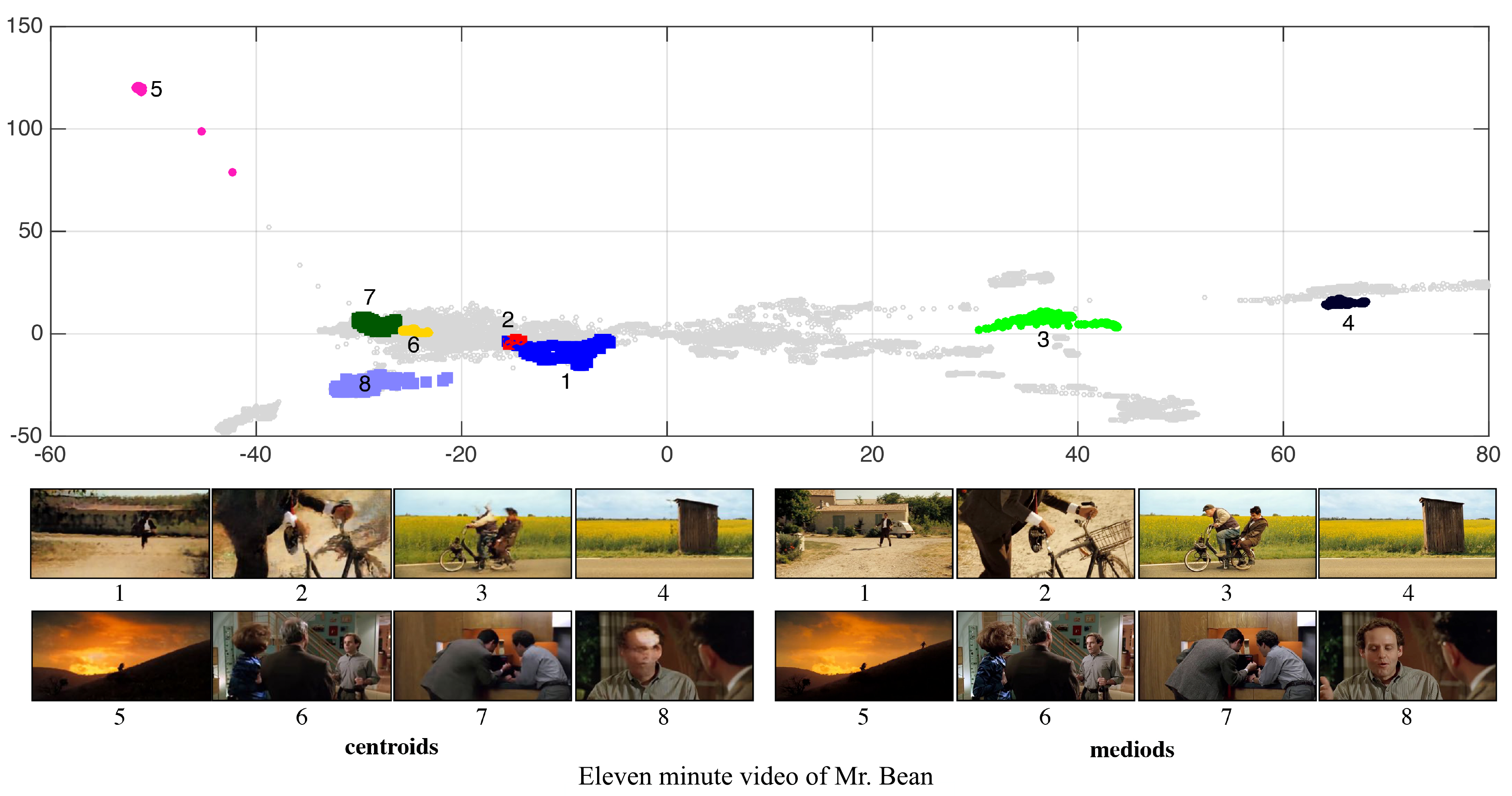}
\caption{\textbf{Video Exploration}: We embed an 11-minute video with a video-specific autoencoder. Our approach allows us to quickly visualize the contents of the video by providing an average summary of the selected region. A user can select the region on the 2D visualization. Our approach generates the centroid of the points in selected region. We also
show the mediod corresponding to each centroid. A user can quickly glance by browsing over the points or selecting the region, and can also stitch representative/selected frames in the video.
}
\label{fig:explore}
\end{figure*}

\noindent{\bf Video Averages:} Arguably the simplest way to summarize a video may be an pixelwise average of its frames~\cite{zhu2014averageExplorer}. Figure~\ref{fig:avg} compares this with a reconstructed average of latent codes from a DAVIS video~\cite{Pont-Tuset_arXiv_2017}. We contrast it with a simple per-pixel mean of the frames, and observe sharper results using our approach. We quantitatively show these improvements in Table~\ref{tab:avg-res}. 

\noindent{\bf Video clustering:} 
Latent codes can be used to discover different visual modes in a video by clustering (e.g., with k-means). Figure~\ref{fig:cont}-(c) shows that two distinctive modes naturally appear for videos with multiple shots. We also use this property to cluster collections of videos by clustering latent codes learned from multi-video manifolds (i.e., from an autoencoder trained on frames from multiple videos; see Appendix~\ref{appd:explore}). 

\begin{table}
\setlength{\tabcolsep}{3pt}
\def\arraystretch{1.3}
\center
\begin{tabular}{@{}l c  c c}
\toprule
 &   \textbf{FID}~\cite{heusel2017fid} $\downarrow$ \\
\midrule
\textbf{DAVIS}~\cite{Pont-Tuset_arXiv_2017}      & \\
Per-pixel Average  & 306.01 \\
Latent Code Average  & 280.15\\
Latent Code Average + iterative  & \textbf{205.96} \\
\bottomrule
\end{tabular}
\vspace{0.2cm}
\caption{\textbf{Video Averages:} We compare approaches for computing an average frame from a video, using 50 videos from the DAVIS dataset. We measure perceptual quality using the FID metric~\cite{heusel2017fid}, which measures the frechet distance between the average frame and all frames of the corresponding video. We see that a simple pixelwise average is not perceptually faithful to the video, while averaging latent codes is more effective. Iteratively reprojecting the latent reconstruction results in an even more faithful average.}
\label{tab:avg-res}
\end{table}

\noindent{\bf Video exploration} allow users to quickly peruse large amounts of video (e.g., consider an analyst who must process large amounts of surveillance video). A user can select a region in the PCA-based 2D visualization that allows for them to quickly visualize summaries by averaging arbitrary subsets of video frames $V_\text{subset} \subseteq V$:
 \begin{align}
    g\bigg(\frac{1}{|V_\text{subset}|}\sum_{x \in V_\text{subset}} f(x)\bigg).
 \end{align}
We give an example of how to explore a video on an 11-minute video in Figure~\ref{fig:explore}.

\section{Editing Videos}
\label{sec:edit}

\begin{figure*}
\includegraphics[width=\linewidth]{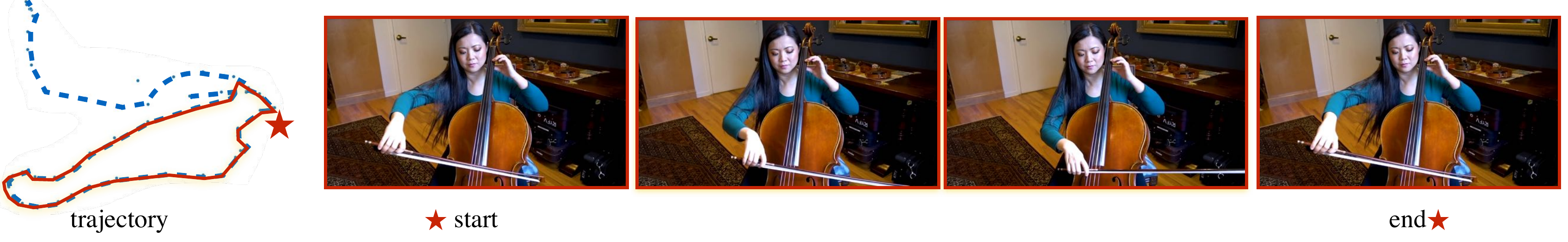}
\caption{\textbf{Video Textures }:  We create arbitrary length video (also known as video textures) from an existing short sequence by making continuous loops. Our ability to associate frames within a video using the latent codes enable us to create infinite loops for repetitive motion. We show many examples in the supp material. 
}
\label{fig:tex}
\end{figure*}

\begin{figure*}
\includegraphics[width=\linewidth]{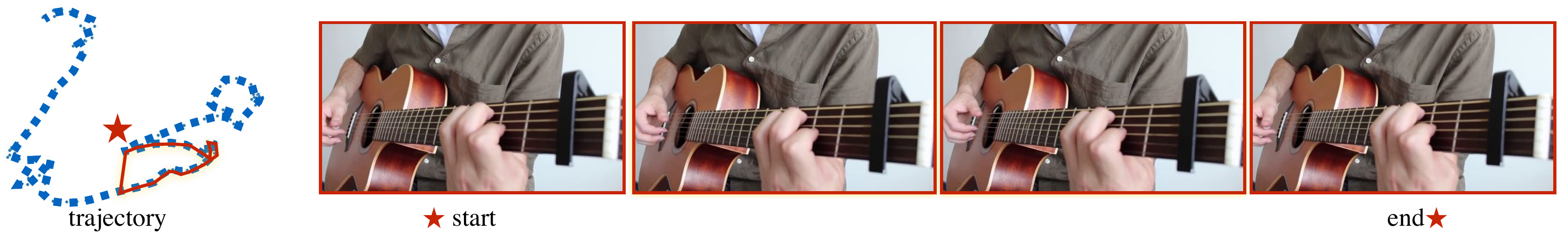}
\caption{\textbf{User-Controllable Trajectories}: A user can create  trajectories they want for various videos using PCA visualization. Prior approaches~\cite{schodl2000video} find close loops to do video texture. It is not a strict requirement for our work as we are able to interpolate new frames between $x_1$ and $x_2$ to smoothly transition between these frames. 
}
\label{fig:tex2}
\end{figure*}

\noindent\textbf{Video Textures:} We can create infinite-loop video textures~\cite{schodl2000video} from a short sequence, as seen in Figure~\ref{fig:tex}. We can find corresponding frames via a cosine similarity on latent codes $f(x_1), f(x_2)$. We emphasize two important distinctions from prior work: (1) PCA visualization enables a user to {\em interactively} define loops and even paths for the target video; and (2) we can close loops without requiring an exact match by interpolating nearby latent codes, which we see in Figure~\ref{fig:tex2}. Table~\ref{tab:texture} shows the benefits of our interpolating nearby latent codes. 

\begin{table}

\setlength{\tabcolsep}{3pt}
\center
\begin{tabular}{@{}l c  c c}
\toprule
 &   \textbf{LPIPS}~\cite{Zhang_2018_CVPR} $\downarrow$ \\
\midrule
\textbf{PENN}~\cite{zhang2013actemes}     & \\
Frame Difference  & 0.136 \\
Texture Difference & \textbf{0.100} \\
\bottomrule
\end{tabular}
\vspace{0.2cm}
\caption{\textbf{Video Textures:} We compare approaches for connecting two similar frames from a video, using 50 videos from the PENN dataset. We evaluate these approaches using LPIPS~\cite{Zhang_2018_CVPR}, which measures the perceptual similarity between two images. Given two similar frames $x_1$ and $x_2$, prior works would simply connect these frames directly. However, these frames may not necessarily have a smooth transition, while we can create arbitrarily long, smooth video textures by continuously connecting frames with interpolation.  For every video, we take two frames $x_1$ and $x_2$ and calculate the LPIPS between them. Our method interpolates a new frame between these two frames, and we calculate the LPIPS between this new frame and $x_2$. \textbf{A lower score is better}.}
\label{tab:texture}
\end{table}

\begin{figure*}[t]
\includegraphics[width=\linewidth]{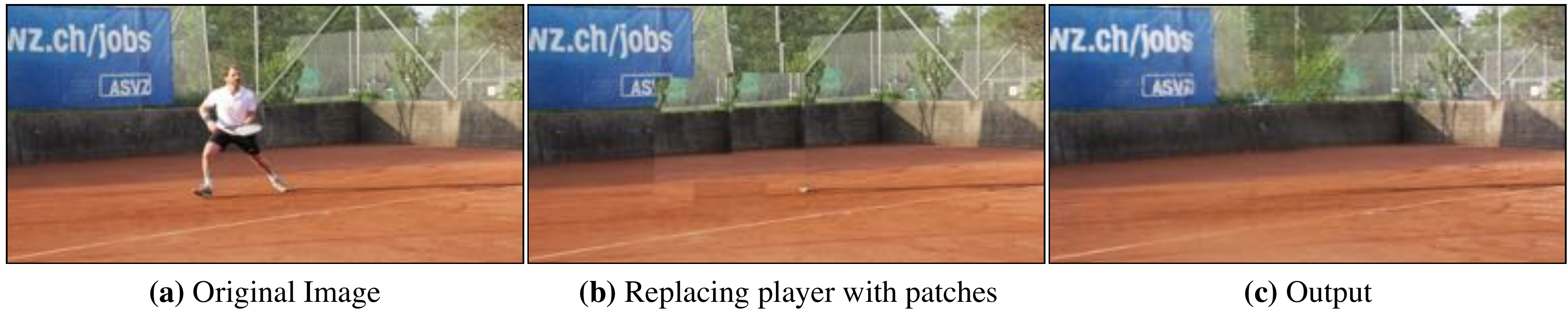}
\caption{\textbf{Spatial Editing: } Given an image, a user can edit it by copy-pasting a patch from the surroundings to the target location and feed it to the video-specific autoencoder. We show an example of spatial editing on a frame in the tennis video from the \textbf{DAVIS}~\cite{Pont-Tuset_arXiv_2017} dataset. We train an autoencoder on the tennis video. Given a frame from the video, we are able to remove the tennis player from the image by replacing the player with surrounding patches in the image and passing it through the autoencoder.}
\label{fig:spt-rem-mp}
\end{figure*}

\begin{figure*}[t]
\includegraphics[width=\linewidth]{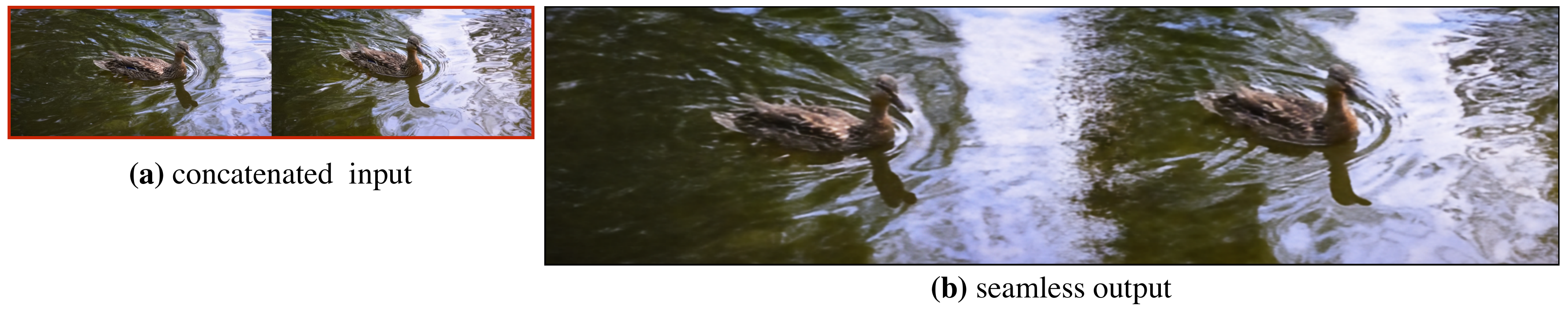}
\caption{\textbf{Seamlessly Stitching Frames: } We naively concatenate the spread-out frames in a video and feed it through the video-specific autoencoder. The learned model generates a seamless output and can capture reflection and ripples in water.
}
\label{fig:stitch}
\end{figure*}

\noindent\textbf{Object Removal and Insertion: } The reprojection property of autoencoder also enables us to learn patch-level statistics in a frame. To remove an object, we copy a patch from surroundings to fill the bounding box. Despite discontinuities, the autoencoder generates a continuous spatial image as shown in Figure~\ref{fig:spt-rem-mp}. We can also use the video-specific autoencoder to insert the known content from the video in a frame. The video-specific autoencoder also allows us to stitch spread-out video frames as shown in Figure~\ref{fig:stitch}.  We naively concatenate the different frames and feed it through the video-specific autoencoder, and the learned model can generate a seamless output. Using the same property, we can also stretch frames and do spatial extrapolation. Examples in Appendix~\ref{appd:edit}.

\begin{figure*}[t]
\centering
\includegraphics[width=1.0\linewidth]{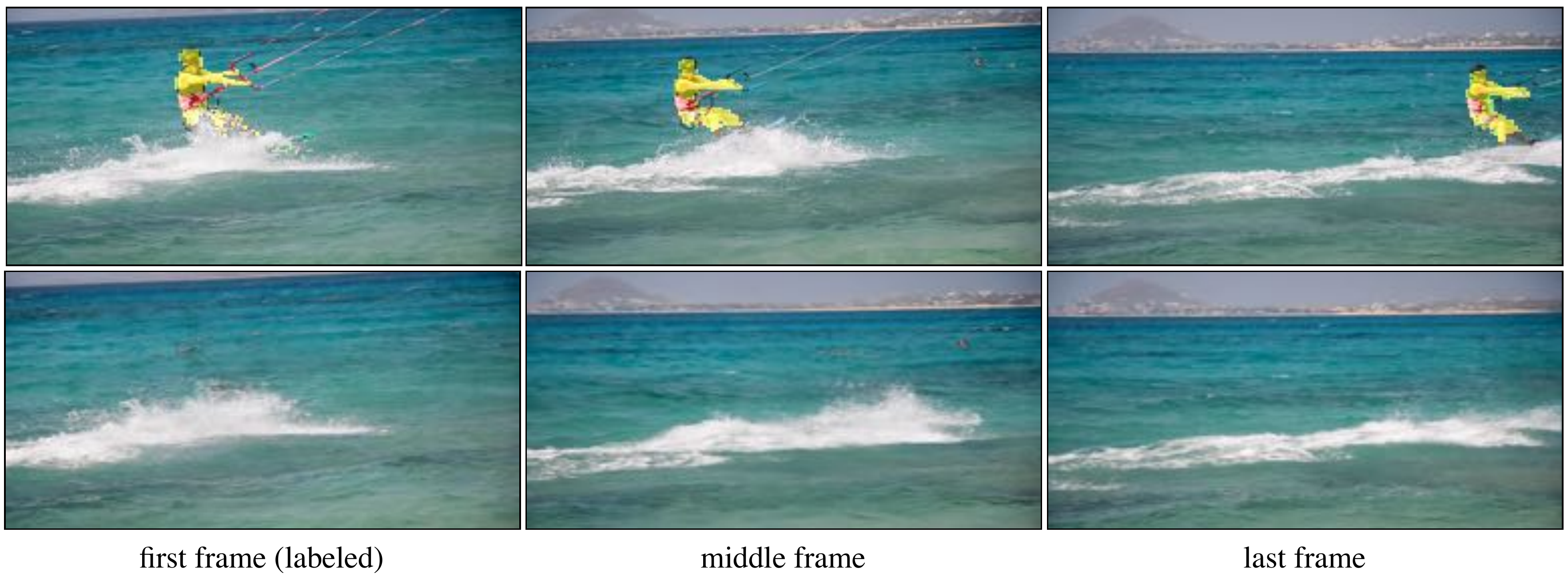}
\caption{\textbf{Instance Mask Propagation:} Given the first labeled frame, we propagate the instance labels using the pixel codes. We show
here the labels in the first frame on the left. We show the propagated labels from the first frame to the middle and last frame of the video.
Using these propagated masks, and the method demonstrated in Fig.~\ref{fig:spt-rem-mp}, we can then perform object removal on the video.
}
\label{fig:mask}
\end{figure*}

\noindent\textbf{Pixel Correspondences and Mask Propagation: } We use our pixel codes to establish a pixel-wise correspondences in the adjacent video frames via a cosine similarity measure. The pixel correspondences allow us to propagate the instance labels from one frame to the rest of video. The mask propagation is especially crucial from user perspective as: (1) it is a tedious task to label entire video; and (2) it helps us to use one less segmentation module in our system. Crucially, we do not need perfect labels for spatial editing as the video-specific autoencoder corrects the imperfections due to its ability to generate continuous spatial outputs.  We show an example of mask propagation in Figure~\ref{fig:mask}, and how we can use masks to edit videos.

\section{Transmitting Videos}
\label{sec:transmit}

\begin{table}
\small{
\setlength{\tabcolsep}{3pt}
\def\arraystretch{1.3}
\center
\begin{tabular}{@{}l c  c c}
\toprule
 & \textbf{Task}  &  \textbf{PSNR}$\uparrow$ & \textbf{SSIM}$\uparrow$  \\
\midrule
\textbf{PENN}~\cite{zhang2013actemes} &     &  &  \\
SuperSlowMo~\cite{jiang2018super} & \cmark   & $31.060\pm2.324$ & $0.951\pm0.018$ \\
Ours & \xmark   & \bm{$33.643\pm4.315$} & \bm{$0.969\pm0.020$}  \\
\midrule
\textbf{Random Web Videos} &     &  &  \\
SuperSlowMo~\cite{jiang2018super} & \cmark   & $34.156\pm1.853$ & $0.721\pm0.100$ \\
Ours & \xmark   & $34.274\pm1.935$ & $0.723\pm0.100$  \\
\bottomrule
\end{tabular}
% \vspace{0.2cm}
\caption{\textbf{Temporal Super-Resolution:} We contrast our approach with an off-the-shelf SuperSlowMo~\cite{jiang2018super} model. We use $150$ videos from Penn Action dataset. We show interpolation between every other frames. The original frames (more than $10,000$ frames in total) are used as a ground-truth for evaluation and not used for training the video-specific autoencoders. We compute PSNR and SSIM scores between the original frames and interpolated frames (\textbf{Higher is Better}). Our approach achieves competitive performance. We also compare our approach with SuperSloMo using random web videos in the same manner.}
\label{tab:visa-tres}
}
\end{table}

\begin{table}
\small{
\setlength{\tabcolsep}{3pt}
\def\arraystretch{1.3}
\center
\begin{tabular}{@{}l c  c c}
\toprule
 & \textbf{Task}  &  \textbf{PSNR}$\uparrow$ & \textbf{SSIM}$\uparrow$  \\
\midrule
\textbf{DAVIS}~\cite{Pont-Tuset_arXiv_2017} &    &  &  \\
Ours (ALT) & \xmark   & $22.065\pm5.205$ & $0.728\pm0.171$  \\
Ours (ALL) & \xmark   & $24.272\pm6.092$ & $0.802\pm0.154$  \\
 Ours (ALL+iterative) & \xmark   & $28.938\pm7.356$ & $0.883\pm0.151$  \\
 \midrule
SuperSlowMo~\cite{jiang2018super} & \cmark   & $27.077\pm5.950$ & $0.871\pm0.134$ \\
\bottomrule
\end{tabular}
% \vspace{0.2cm}
\caption{\textbf{Studying Influence of Sparse Samples on Video-Specific Autoencoders via Temporal Super-Resolution:} We study our approach with sparse frames from $90$ videos of DAVIS dataset~\cite{Pont-Tuset_arXiv_2017}. We show interpolation between every other frames. The original frames are used as a ground-truth for evaluation. We consider two scenarios:  (1) using alternate (\textbf{ALT}) frames from a video for training; and (2) using \textbf{ALL} frames from a video for training the video-specific autoencoders (since we do not use temporal information and optimize for this task). We compute PSNR and SSIM scores between the original frames and interpolated frames (\textbf{Higher is Better}). We observe that performance of video-specific autoencoders degrade when making samples extremely sparse, especially for the video with large dynamics. We, however, observe that performance can be improved by using iterative reprojection property. This means we can sparsely transmit frames over the network and do temporal super-res with iterative reprojection property to get dense outputs at the reception. For reference, we also provide the performance of an off-the-shelf SuperSlowMo~\cite{jiang2018super} model.}
\label{tab:visa-tres-02}
}
\end{table}

\begin{table}[h]
\small{
\setlength{\tabcolsep}{5pt}
\center
\begin{tabular}{@{}l  c  c c}
\toprule
 &  \textbf{Task}  &\textbf{PSNR}$\uparrow$ & \textbf{SSIM}$\uparrow$  \\
\midrule
\textbf{4X Super-Res}   &  & &  \\
ESR-GAN~\cite{wang2018esrgan}  & \cmark &  $26.886\pm3.821$ & $0.847\pm0.087$\\
Ours  & \xmark &   \bm{$31.493\pm3.681$} & \bm{$0.940\pm0.053$}  \\
\midrule
\textbf{16X Super-Res}  &  & &  \\
ESR-GAN~\cite{wang2018esrgan}  & \cmark &   $19.324\pm3.163$ & $0.605\pm0.162$\\
Ours  & \xmark &   \bm{$25.737\pm3.511$} & \bm{$0.856\pm0.086$}  \\
\midrule
\textbf{64X Super-Res} &  &  &  \\
ESR-GAN~\cite{wang2018esrgan}  & \cmark &   $15.441\pm2.752$ & $0.490\pm0.170$\\
Ours  & \xmark &   \bm{$17.862\pm4.789$} & \bm{$0.622\pm0.196$}  \\
\bottomrule
\end{tabular}
% \vspace{0.2cm}
\caption{\textbf{4X, 16X, and 64X Spatial Super-Resolution:} We contrast our approach with an off-the-shelf ESR-GAN model~\cite{wang2018esrgan} trained for 4X super-resolution. For 16X super-resolution, we use the ESR-GAN model twice. For 64X super-resolution, we use it three times iteratively. We use $90$ videos from DAVIS dataset~\cite{Pont-Tuset_arXiv_2017} for this evaluation. The original frames ($256\times512$) are used as a ground-truth for evaluation. We compute PSNR and SSIM scores between the original frames and the outputs from two approaches (\textbf{Higher is Better}). Our approach achieves better performance without being optimized for the task. %Additionally, our approach is trained on a specific video and does not use any extra data.
}
\label{tab:sres}
}
\end{table}

Finally, we examine applications of autoencoders motivated by real-time, low bit-rate video transmission (e.g., video conferencing). 
We envision a setting where one can transmit network parameters off-line along with online transmission of aggressively-subsampled frames, both temporally and spatially. Such samples can then be decoded at the receiver to produce hi-res, hi-frame rate video.

\noindent{\bf Transmitting Sparse Temporal Frames:} We use simple operation on latent-codes to slow-down (known as temporal super-resolution) and speed-up a video. Given a frame $x_t$ and $x_{t+1}$, we can insert an arbitrary number of frames between by linearly interpolating their latent codes:
\begin{align}
g\bigg(\alpha f(x_t) + (1-\alpha) f(x_{t+1})\bigg), \quad \alpha \in [0,1] \label{eq:int}
\end{align}

We quantitatively compare our approach with Super Slowmo~\cite{jiang2018super}\footnote{We use the publicly available model from \url{https://github.com/avinashpaliwal/Super-SloMo}.} in Table~\ref{tab:visa-tres}. We use an off-the-shelf SuperSlowMo~\cite{jiang2018super} model trained on a large dataset in a supervised manner specifically for the task of temporal super-resolution. We use $150$ videos ($10$ videos from each action) from Penn-Action dataset~\cite{zhang2013actemes}. We do 2X interpolation, i.e., interpolate between  every other frame. The original frames that are used as a ground-truth for evaluation and are not used for training the video-specific autoencoder. We compute PSNR and SSIM scores between the original frames and interpolated frames. \textbf{Higher is Better}. Table~\ref{tab:visa-tres} shows the performance of our approach with SuperSlowMo~\cite{jiang2018super}. We also contrast the performance on random web videos in Table~\ref{tab:visa-tres} and observe competitive performance to SuperSlowMo. 

We also study our approach using $90$ videos of DAVIS dataset~\cite{Pont-Tuset_arXiv_2017} in Table~\ref{tab:visa-tres-02}. Sequences in DAVIS dataset consists of $50-80$ frames (roughly sampled at $5$fps)  with substantial dynamics. Training a video-specific autoencoder with alternate frames is challenging as it becomes even more sparse. Since we do not use temporal information and optimize for temporal super-res, we contrast the performance of our model when trained with all frames (ALL) and alternate frames (ALT). We observe that performance of a video-specific autoencoder degrades when trained on extremely sparse frames from a video with large camera and object motion. We, however, observe that performance can be improved by using iterative reprojection. This means we can sparsely transmit frames over the network and do temporal super-res with iterative reprojection property to get dense outputs. Finally, video-specific autoencoders enable arbitrary temporal resampling of a video, in contrast to previous methods that are often trained for a fixed resampling factor without being able to generalize to others~\cite{liu2017voxelflow,long2016learning,niklaus2017video}. Our approach can also benefit from advances in temporal super-resolution approaches~\cite{lee2020adacof,park2020bmbc} using optical flow. However, in this work we wanted to limit the use of an application-dependent module.

\noindent\textbf{Transmitting Low-Res Frames: } Our goal is to transmit minimal bits over network and yet be able to get an original quality original video output at the reception. Spatial super-resolution is enabled by the reprojection property of the autoencoder (discussed in Section~\ref{sec:method}). The convolutional autoencoder allows us to use videos of varying resolution. We get temporally smooth outputs without using any temporal information for all the experiments. We quantitatively compare our approach with an off-the-shelf ESR-GAN~\cite{wang2018esrgan} model for 4X, 16X, and 64X super-resolution on $90$ videos (roughly $6,000$ frames) from DAVIS dataset~\cite{Pont-Tuset_arXiv_2017}. ESR-GAN is trained for 4X super-resolution. We use it twice for 16X super-resolution, and thrice for 64X super-resolution. Table~\ref{tab:sres} contrast our approach with ESR-GAN for these three settings. Our approach achieves better performance without being optimized for the task. We show examples in Appendix~\ref{appd:transmit}.

%%%%%%%%% Discussion

\section{Discussion}
An autoencoder trained using individual frames of a specific video without any temporal information can be used for a wide variety of tasks in video analytics without even optimizing for any of those tasks. This feat could be possible by careful analysis of spatial and temporal properties of a video-specific autoencoder. We hope that an interface based on this simple representation can enable users to easily design new applications (not even explored in this work) without much overhead such as simple algebraic operations on the latent codes. We also hope that every video uploaded on web get its autoencoder. It may not take a long time to train a model but it can drastically reduce the amount of resources required for video transmission, exploration, and processing because once trained, the model can be used for fairly large number of applications without much constraints. 

\section{Limitations and Societal Impact}

$400$ hours of video data is uploaded to YouTube every minute. The rich video data opens up enormous opportunities for exploring the vast visual content. Creating ways that allows a user to explore the contents of a video, edit them, and efficiently transmit the content would provide a comprehensive platform to the users. We take a small step towards this grand goal. Our work requires training a video-specific autoencoder. Training a new model at test time is not instantaneous and it depends on the length of the videos. We need to come up with faster ways to train a new model and use insights from transfer learning such as a simple finetuning. In this work, we also observe that training a reliable model is challenging if the frames of a video are temporally sparse. Such a situation is observed when the cameras are fast moving like a hand-held camera. Our current evaluation is conducted using standard video benchmarks and regular web videos that are properly curated. We plan to collect a wide variety of videos from hand-held devices for better analysis. Finally, tools for editing videos can be misused for spreading misinformation. While videos from any source are not considered as an evidence in the court of law, they are still important in forming public opinion on social media like Twitter, Facebook, YouTube. Every time a viewer see a generated video on these platforms, there needs to be a banner displaying “This is a generated content.” so that they do not make any opinions out of it.

\appendix

\section{Exploring a Video}
\label{appd:explore}

\begin{figure*}
\centering
\includegraphics[width=\linewidth]{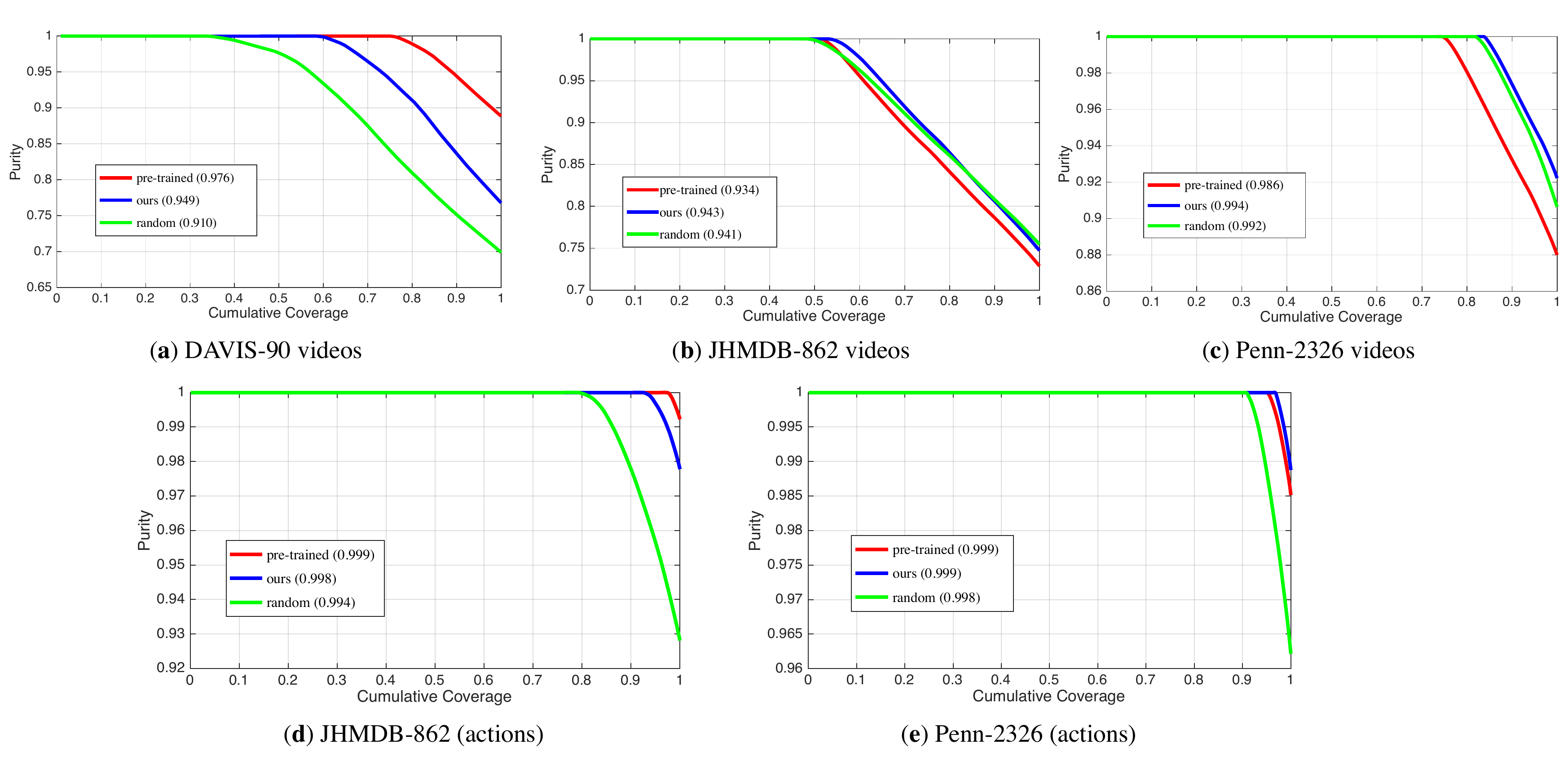}
\caption{\textbf{Visual concept discovery} on three datasets: \textbf{(a)} DAVIS-90 videos~\cite{Pont-Tuset_arXiv_2017}; \textbf{(b)} JHMDB-862 videos~\cite{Jhuang:ICCV:2013}; and \textbf{(c)} Penn-Action 2326 videos~\cite{zhang2013actemes}. We naively concatenate all frames from each dataset and study various methods for recovering the original videos via clustering with k-means (where $K$ is the number of videos in each dataset), measuring purity vs. cumulative coverage. We compare AlexNet~\cite{krizhevsky2017imagenet} \emph{fc-7} features pre-trained on ImageNet~\cite{Russakovsky15}, a randomly initialized autoencoder (i.e., no training), and trained autoencoder. We observe that even random weights of the autoencoder (without any training) can  reliably separate visual concepts. Finally, we observe that both JHMDB and Penn-Action have visually similar videos but with different ids. We, therefore, computed the performance considering the ``action'' class of each cluster instead of their video-id. The respective plots are shown in \textbf{(d)} and \textbf{(e)}. We also show \textbf{area-under-curve} for each method in the legends of each plot.}
\label{fig:purity}
\end{figure*}

Video-specific autoencoders work well on exemplar data distribution. One may wonder their applicability for long and diverse videos. We observe that latent codes of a trained autoencoder can also separate different visual concepts. This allows us to cluster a long video into multiple short videos. In this section, we study this property using the task of video classification for $90$ videos in DAVIS dataset~\cite{Pont-Tuset_arXiv_2017}, $862$ videos of JHMDB dataset~\cite{Jhuang:ICCV:2013}, and $2326$ videos in Penn dataset~\cite{zhang2013actemes}. We learn an autoencoder using all the frames from each of these datasets. Once trained, we do k-means clustering (where $k$ is the number of videos in each dataset) using the latent codes obtained from the trained autoencoder. Figure~\ref{fig:purity}-(a)-(c) demonstrates our ability to separate different visual concepts. Here, we also contrast with \emph{fc-7} features of an AlexNet model~\cite{krizhevsky2017imagenet} trained on the labeled ImageNet dataset~\cite{Russakovsky15}. We observe competitive results. We also study the ability of a \emph{randomly-initialized} autoencoder. We use the latent codes from an \emph{untrained} autoencoder. We observe that even the randomly initialized autoencoder can reliably separate different visual concepts. Finally, we observe that both JHMDB and Penn-Action have visually similar videos but with different ids. We, therefore, computed the performance considering the ``action'' class of each cluster instead of their video-id. The respective plots are shown in Figure~\ref{fig:purity}-(d)-(e).

\begin{figure*}
\centering
\includegraphics[width=\linewidth]{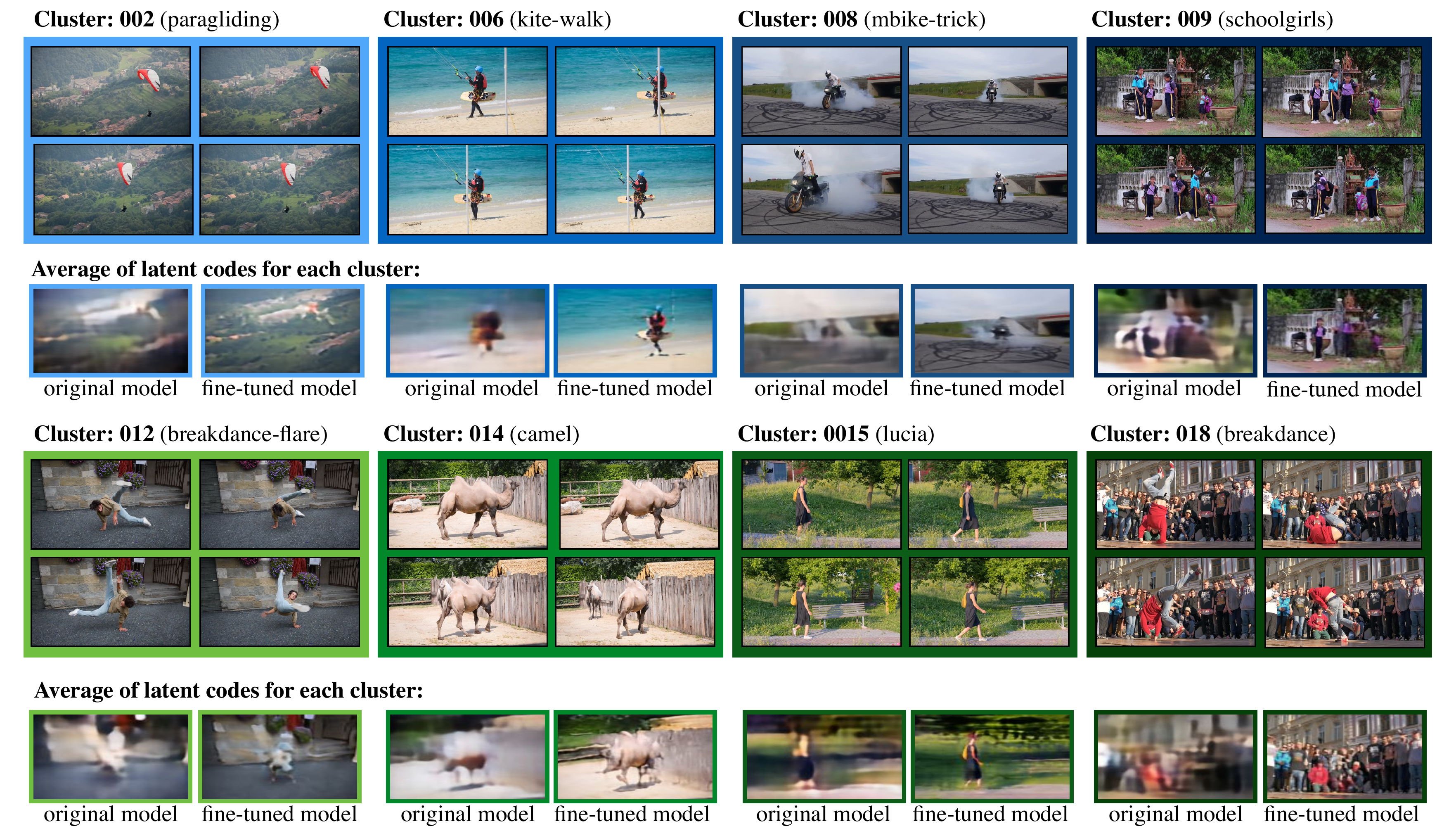}
\includegraphics[width=\linewidth]{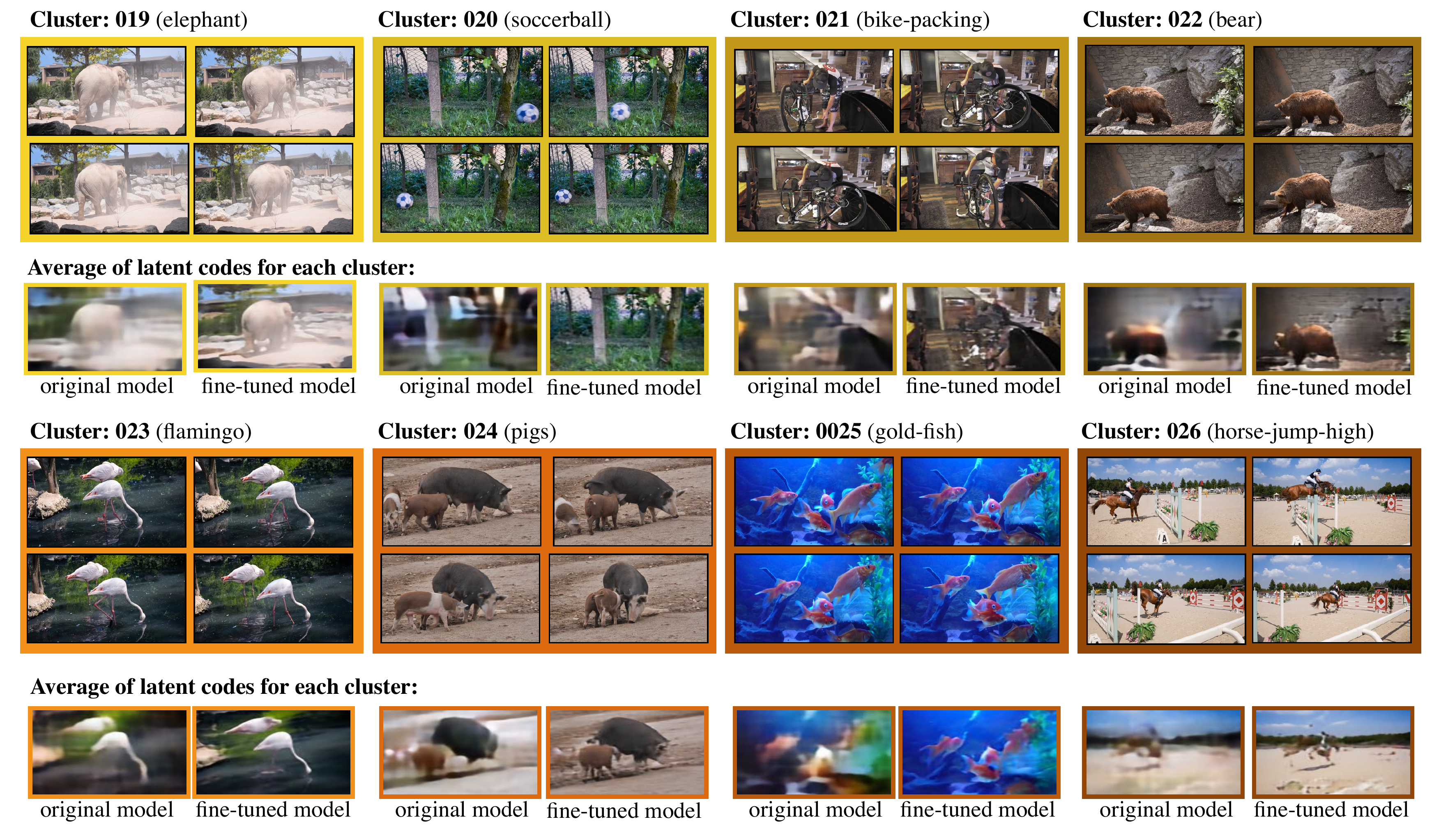}
\caption{\textbf{Leaning a single autoencoder on all the frames of DAVIS dataset:} We show different visual clusters (with purity of 1) when doing k-means on the latent codes for all the frames of DAVIS dataset. We show the average image for each of these clusters using the single model. We can further sharpen the average image by fine-tuning the model for a few iterations on the examples of specific cluster.}
\label{fig:davis_p1}
\end{figure*}

\begin{figure*}
\includegraphics[width=\linewidth]{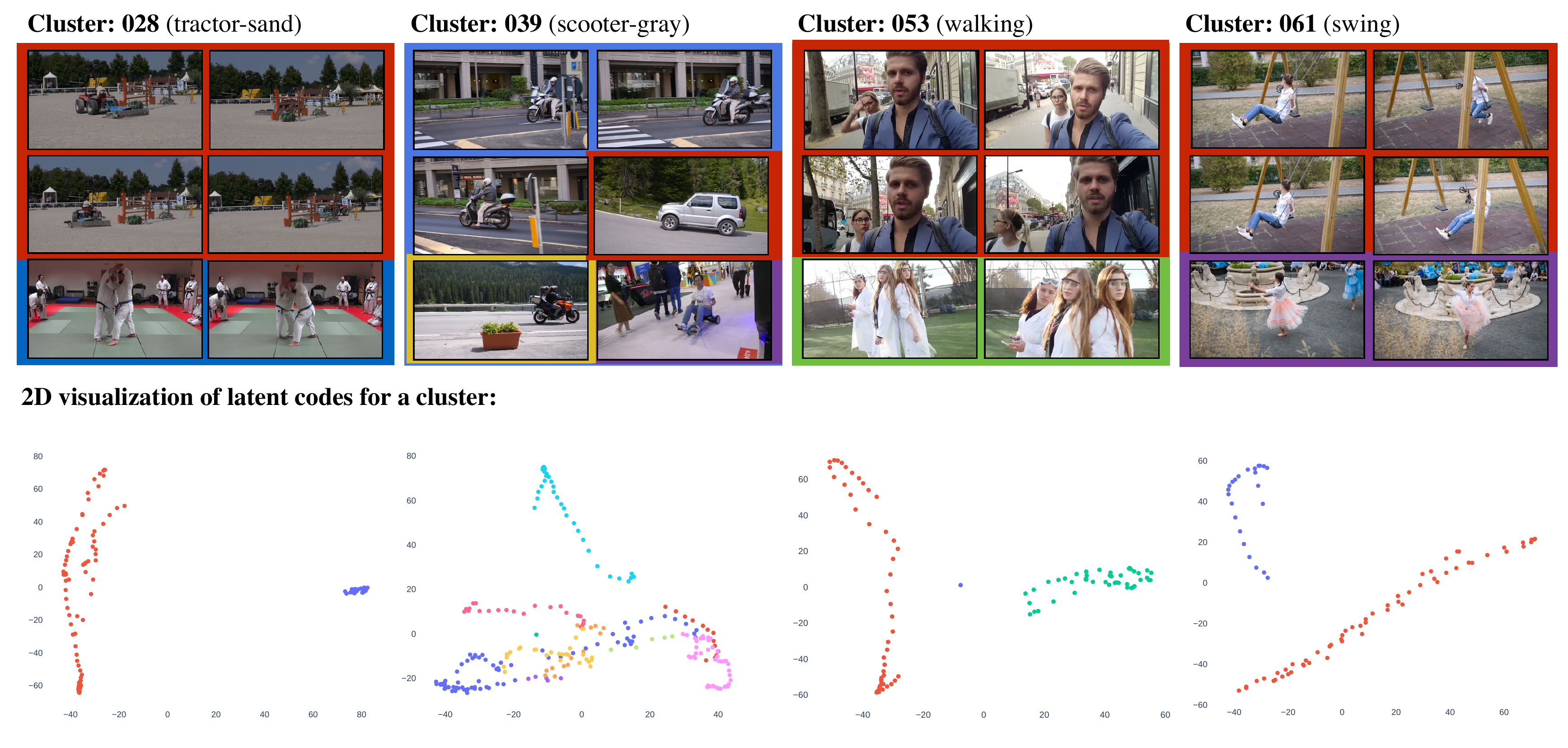}
\caption{\textbf{Impure clusters of DAVIS dataset can be further separated}: We show different clusters with lower purity value. On a closer observation, we find that concepts within most of these clusters can further separate out as shown by the 2D visualization of latent codes.}
\label{fig:davis_p2}
\end{figure*}

\begin{figure*}
\centering
\includegraphics[width=0.96\linewidth]{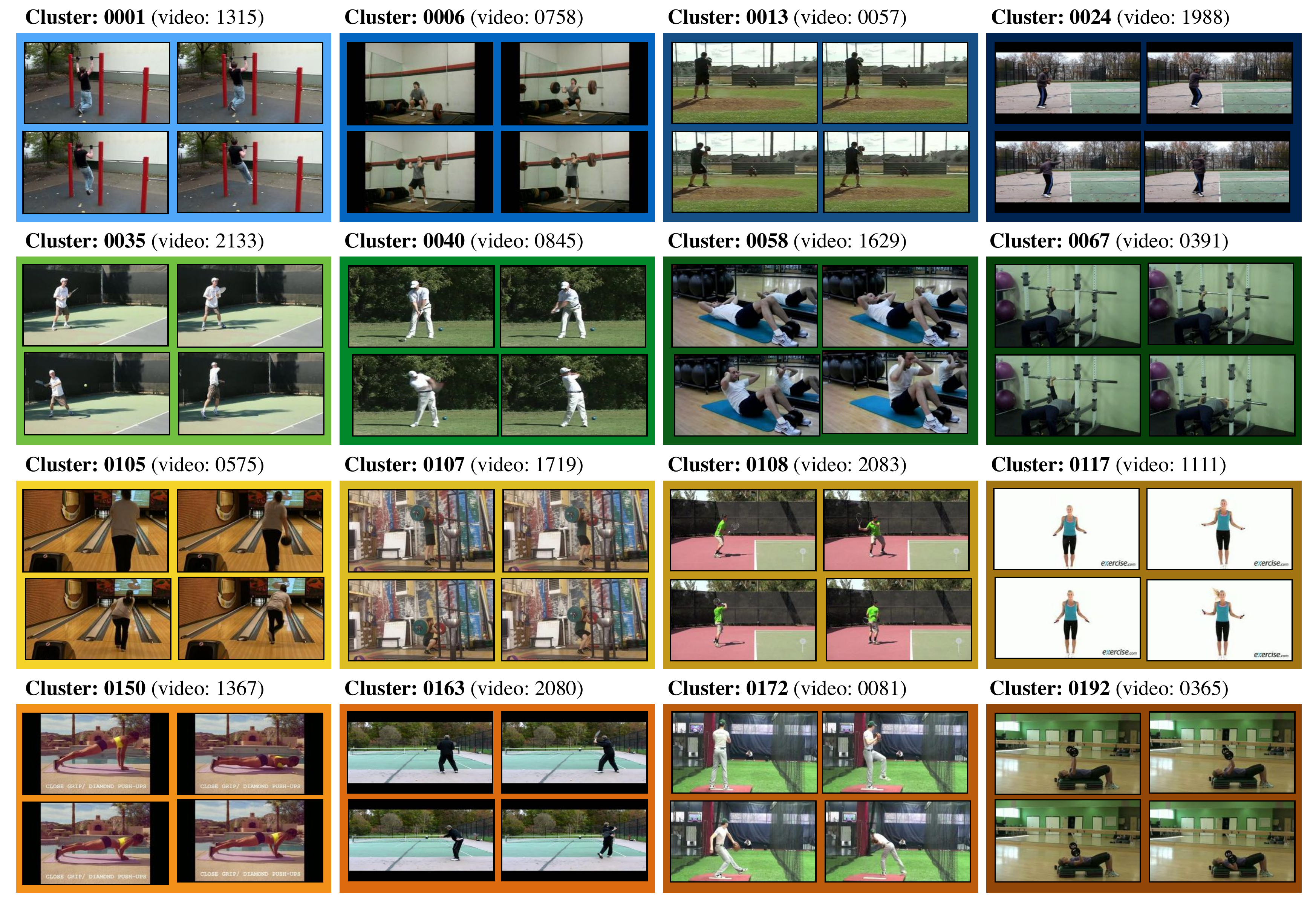}
\caption{\textbf{Pure clusters from the Penn Action dataset} We visualize different clusters (purity of 1) obtained by simple k-means clustering on the latent codes via an autoenoder trained on all the frames on Penn Dataset.}
\label{fig:penn_p1}
\end{figure*}

\begin{figure*}
\centering
\includegraphics[width=\linewidth]{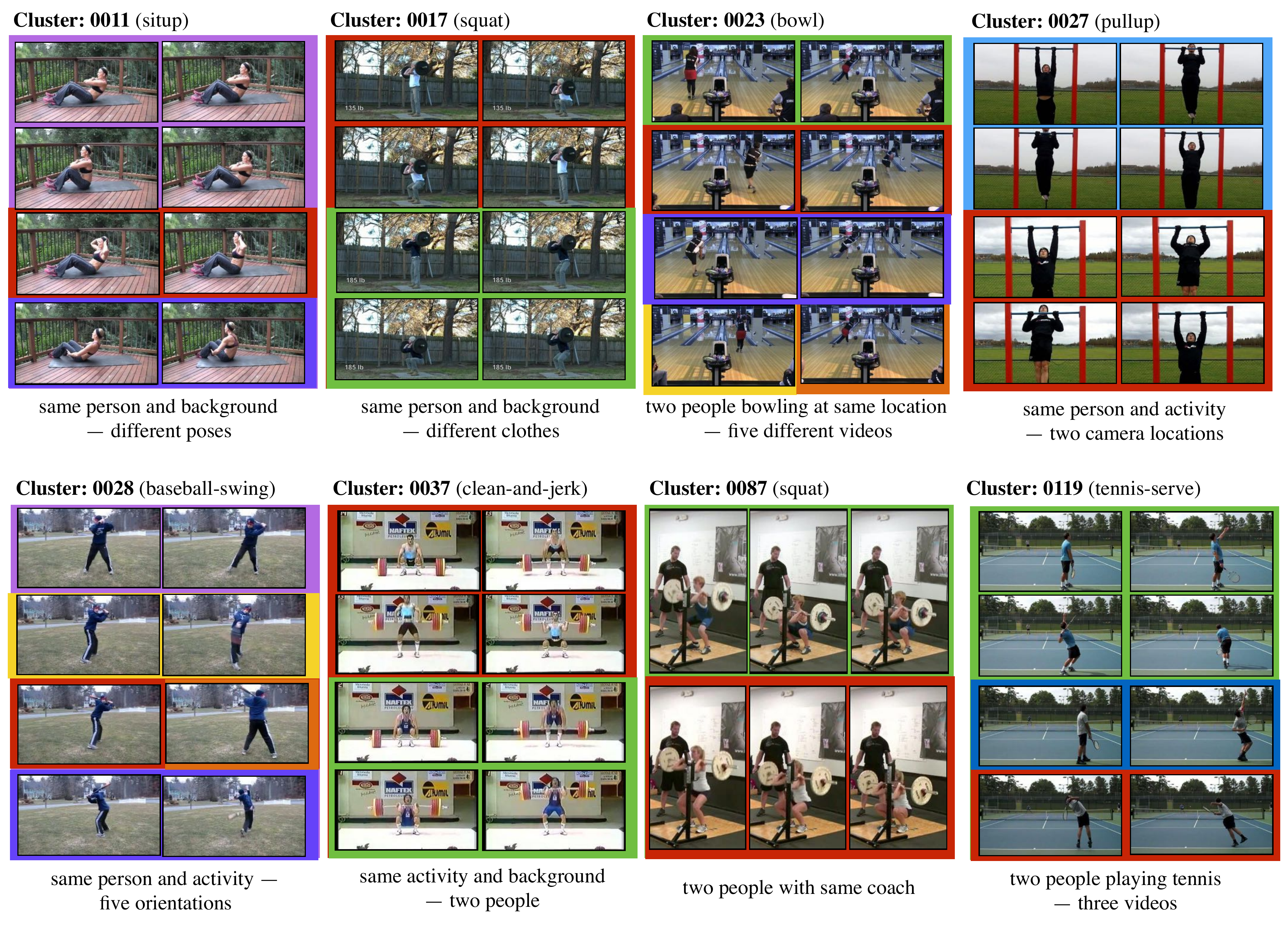}
\caption{\textbf{Interesting clusters from the Penn Action dataset} We show different visual concepts that come from different videos but looks visually similar, i.e., these are the clusters with video purity of less than 1 but action purity of 1. This happens because we have various videos in Penn-Action dataset that has same person and background with slight variation. We show these variations with each cluster.}
\label{fig:penn_p2}
\end{figure*}

\begin{figure*}
\centering
\includegraphics[width=\linewidth]{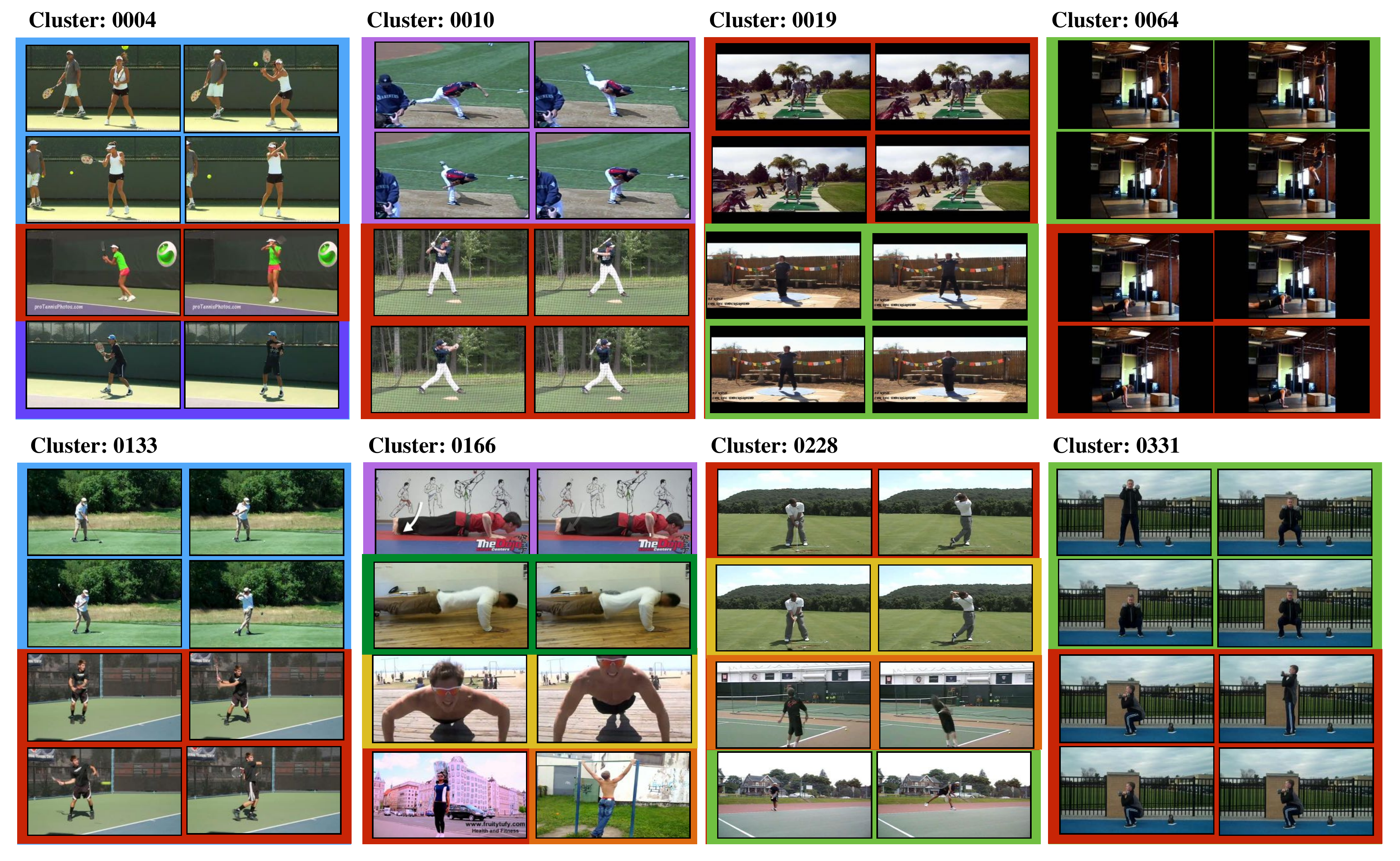}
\caption{\textbf{Impure clusters from the Penn Action dataset:} We now show different clusters with lower purity value. Most of them either capture the same action or are captured in similar background. }
\label{fig:penn_p3}
\end{figure*}

We visualize various visual clusters (with purity of 1) from DAVIS dataset in Figure~\ref{fig:davis_p1}. We show average images for these clusters using the single autoencoder trained on all frames. We can further sharpen these average images by fine-tuning the autoencoder only on the examples in the cluster for $5$ iterations. We also visualize various impure clusters in Figure~\ref{fig:davis_p2} along with 2D visualization of latent codes for the examples in each cluster. We observe that different concepts within an impure cluster can further be separated. We also visualize the clusters of Penn-Action dataset in Figure~\ref{fig:penn_p1}. We observe that both JHMDB and Penn-Action dataset consists of different videos that look similar because of different reasons like same person and background but different orientation, different camera location etc. We visualize these interesting clusters in Figure~\ref{fig:penn_p2}. Finally, we visualize impure clusters in Figure~\ref{fig:penn_p3}. Most of these clusters belong to same action or have a similar background.

With these analysis, we posit that one may also use a simple autoencoder to first cluster the different videos and then use a video-specific autoencoder for each cluster, or fine-tune the model for the cluster. These analysis also shows the potential of autoencoders for average image exploration and unsupervised learning of exemplar visual concepts. We, however, leave them for the future work.

\section{Editing a Video}
\label{appd:edit}

\begin{figure*}[t]
\centering
\includegraphics[width=\linewidth]{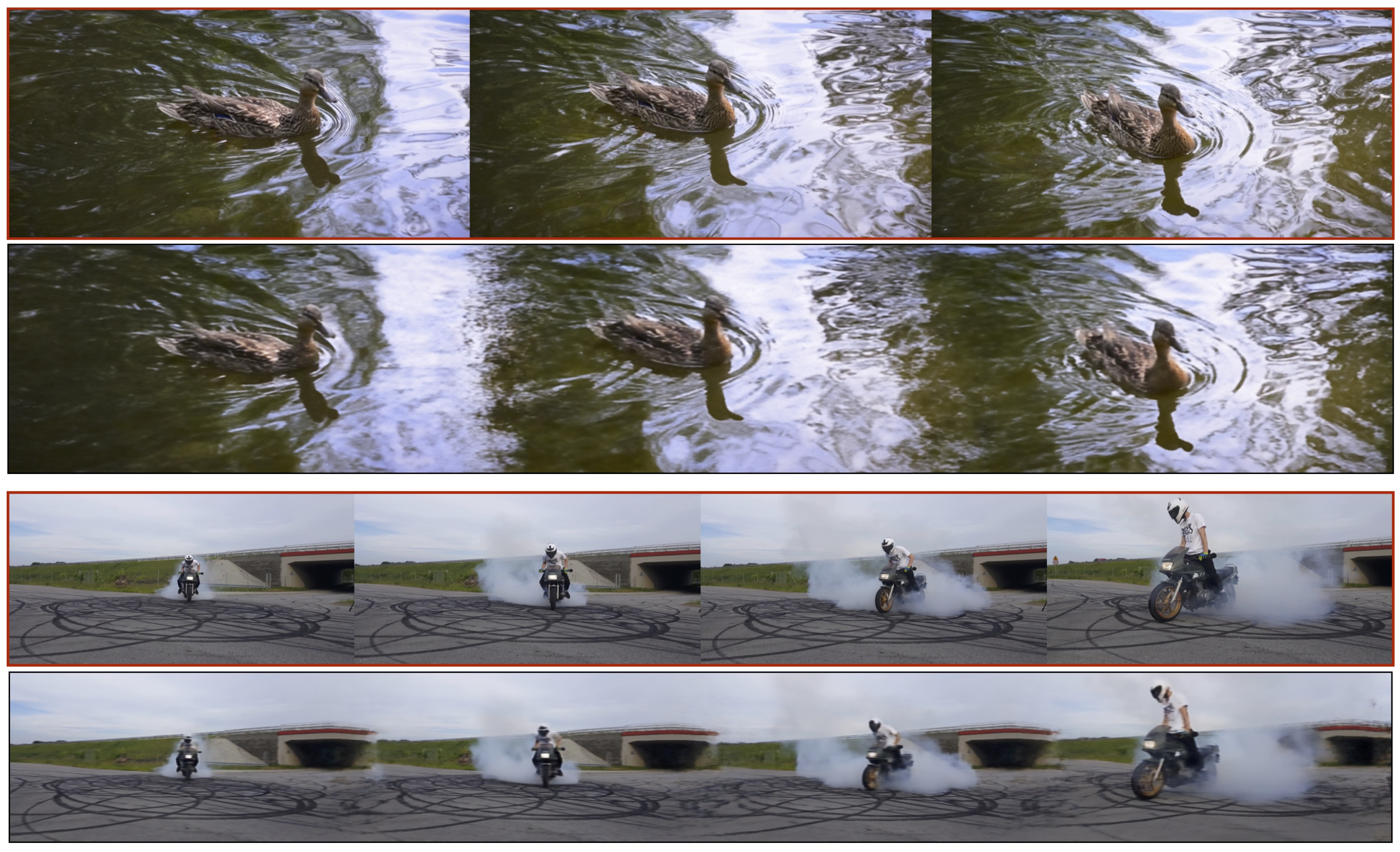}
\includegraphics[width=\linewidth]{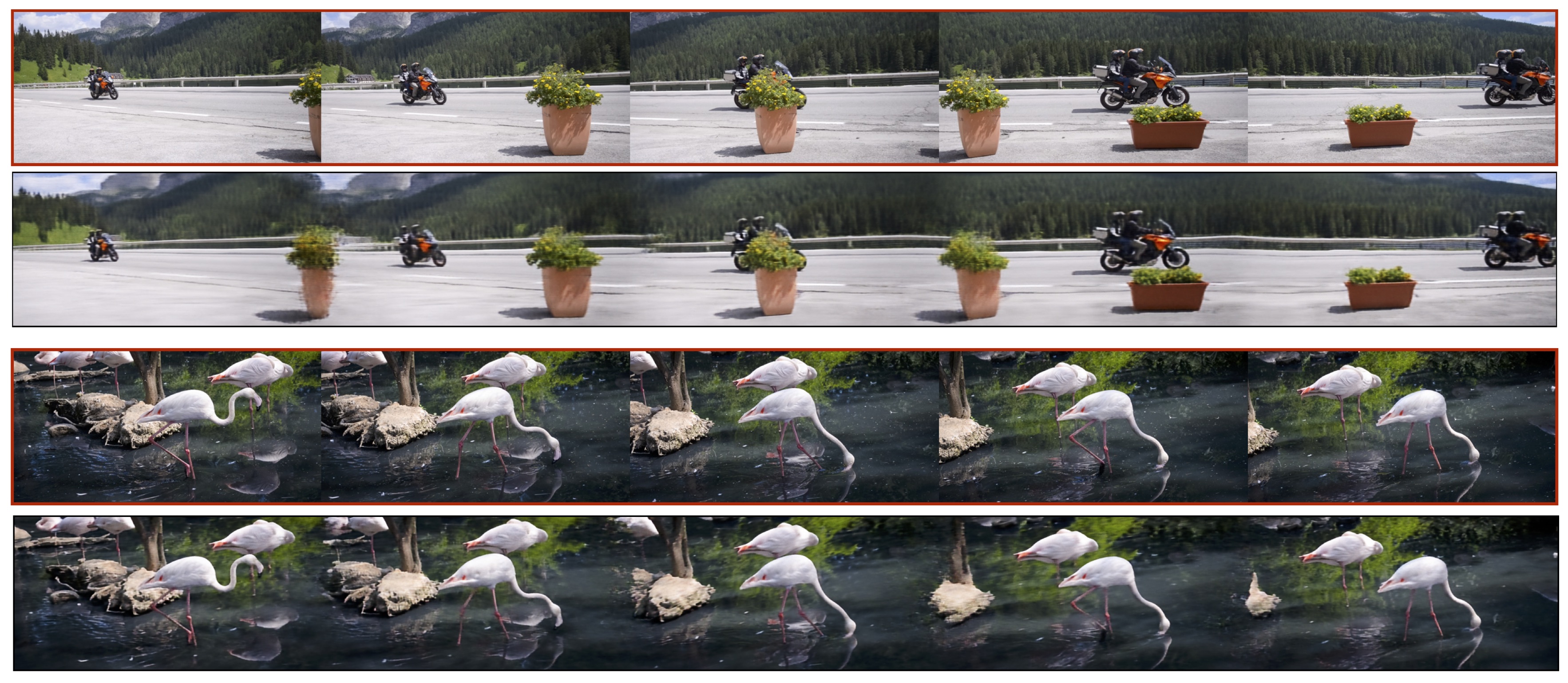}
\caption{\textbf{Stitching Far-Apart Frames in a Video: } We naively concatenate the spread-out frames in a video and feed it through the video-specific autoencoder. The learned model generates a seamless output. The top row in each example shows concatenated frames. The bottom row shows seamless output of video-specific autoencoders. }
\label{fig:stitch-01}
\end{figure*}

\begin{figure*}[t]
\centering
\includegraphics[width=\linewidth]{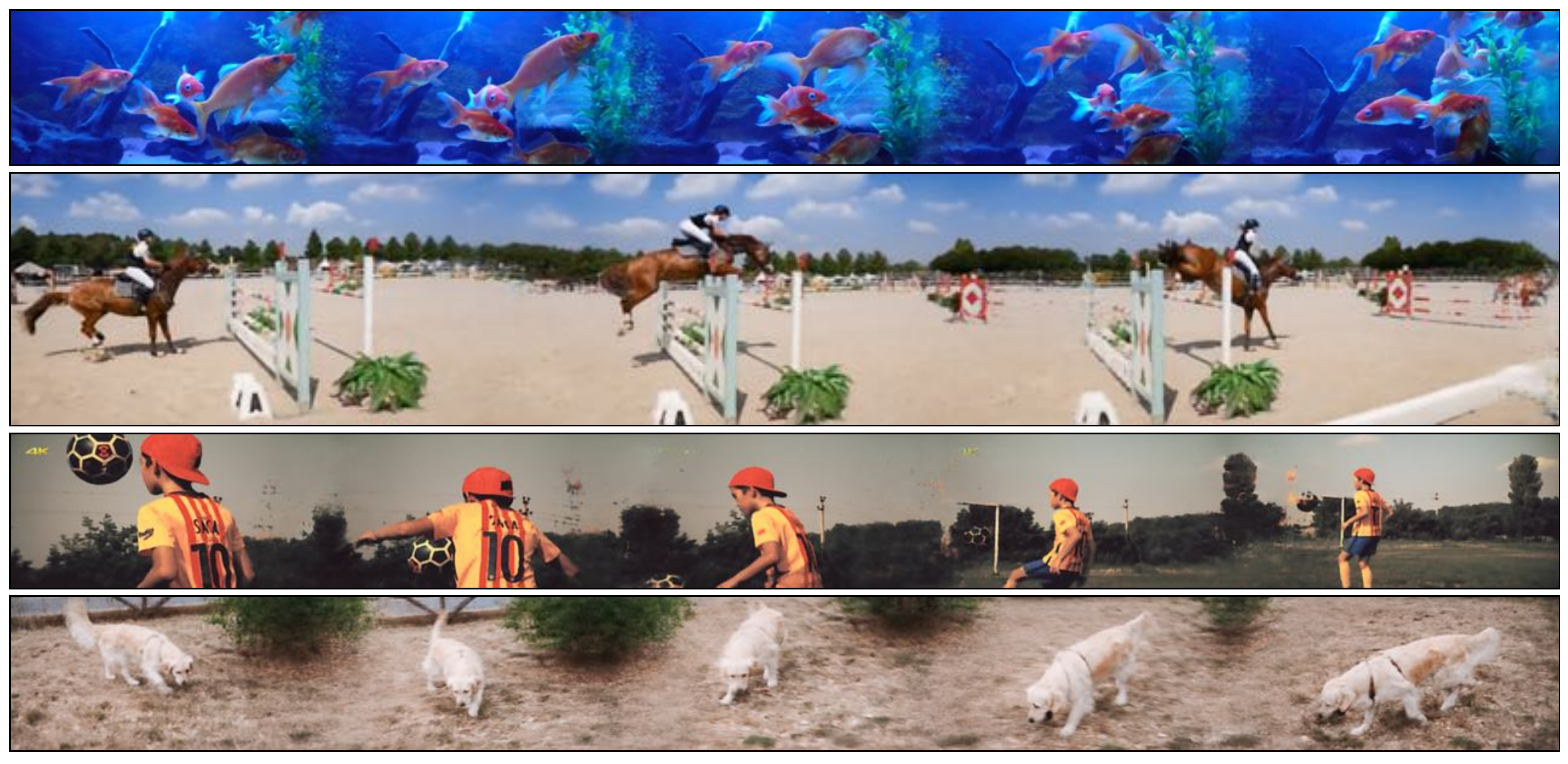}
\caption{\textbf{More examples of stitching far-apart frames in a video: } We show more examples of stitching for different videos and observe continuous seamless outputs. }
\label{fig:stitch-02}
\end{figure*}

\noindent\textbf{Stitching and Stretching Video Frames: }  The ability of the video-specific autoencoder to generate continuous spatial imaging allows us to stitch spread-out video frames. Figure~\ref{fig:stitch-01} shows various examples where we stitch random frames from arbitrary videos with varying texture and content. We naively concatenate the different frames and feed it through the video-specific autoencoder. The learned model generates a seamless output. For e.g., the reflections in water, and circular patterns formed by motorbike. More examples in Figure~\ref{fig:stitch-02}.

\begin{figure*}
\centering
\includegraphics[width=\linewidth]{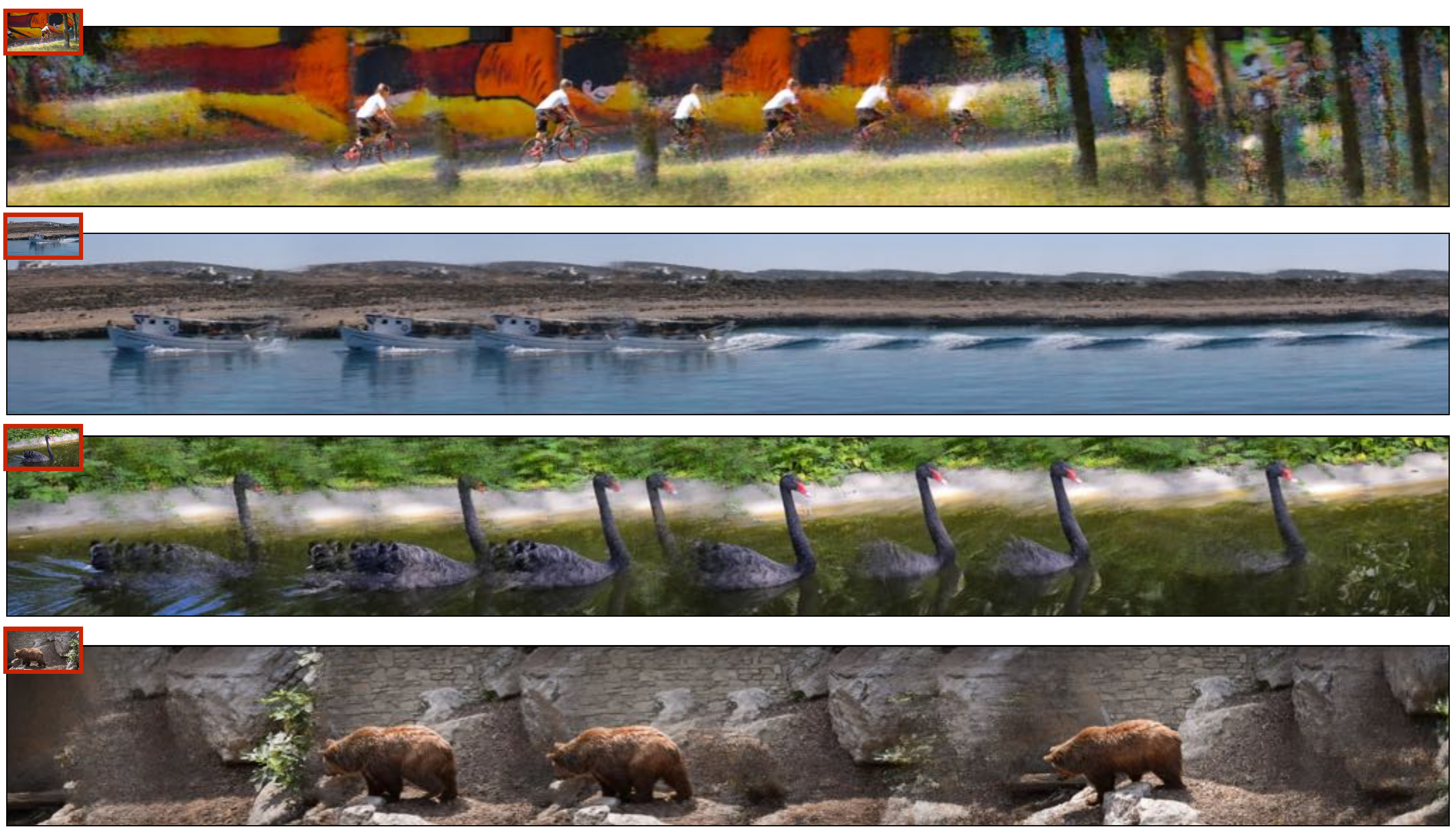}
\includegraphics[width=\linewidth]{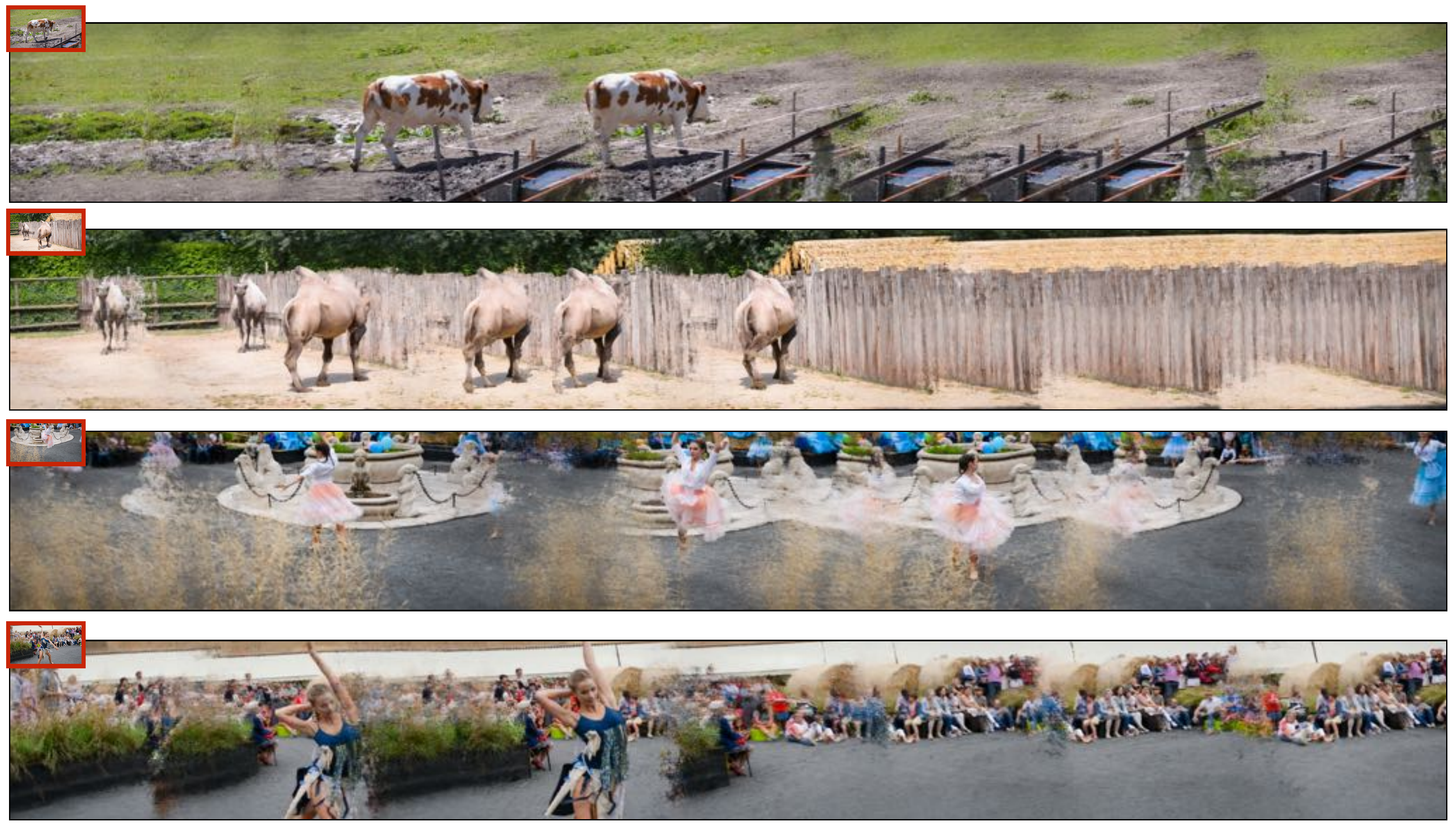}
\caption{\textbf{Stretching a Frame: } Given a $256\times512$ frame of a video (shown in the red box on the left), we stretch it horizontally to $256\times2048$ and feed it to a learned video-specific autoencoder. We iterate $30$ times and get a consistent panorama-type output. We can further use our editing tools to create desirable panoramas as shown in Figure~\ref{fig:spt-rem}.}
\label{fig:stretch-01}
\end{figure*}

The iterative reprojection property of autoencoder and its ability to generate continuous spatial imaging allows us to arbitrarily stretch a video frame. We show various examples of stretching a $256\times512$ video frame to $256\times2048$ in Figure~\ref{fig:stretch-01}.

\begin{figure*}
\centering
\includegraphics[width=0.93\linewidth]{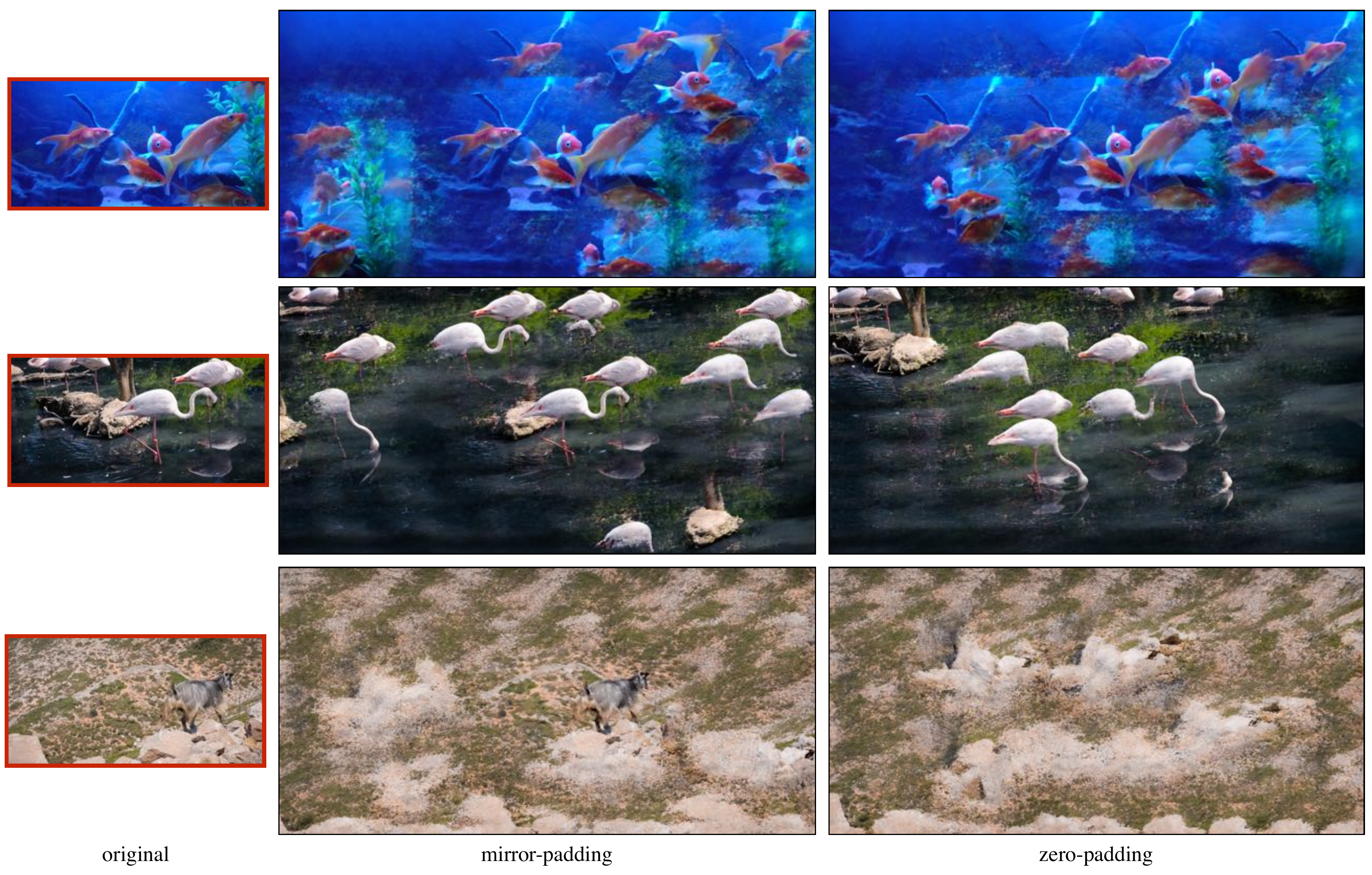}
\includegraphics[width=0.93\linewidth]{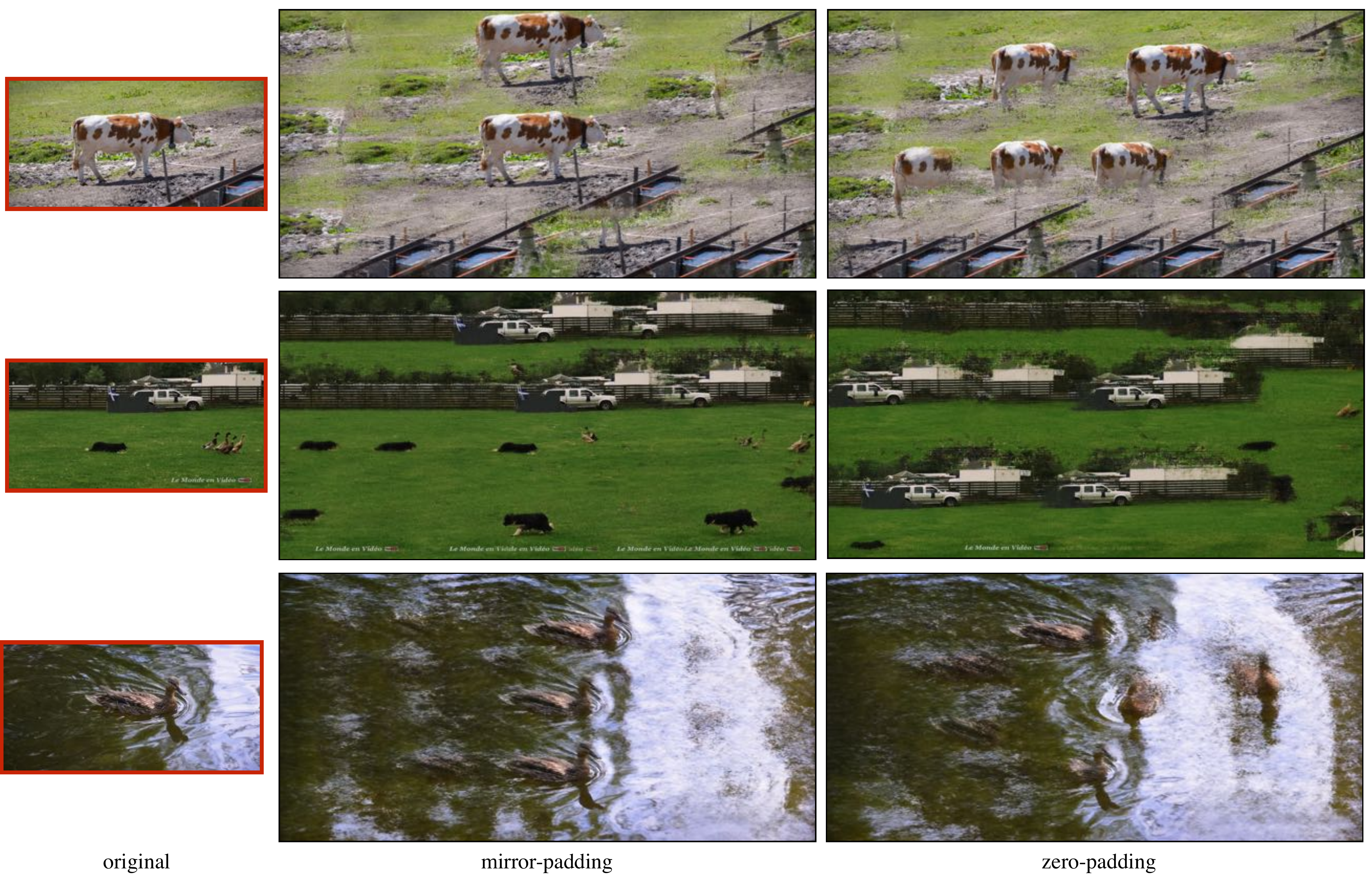}
\caption{\textbf{Spatial Extrapolation: } Given a $256\times512$ image, we spatially extrapolate on its edges to create a $512\times1024$ image. To do this, we \textbf{mirror-pad} the image to the target size and feed it to the video-specific autoencoder. We also show the results of \textbf{zero-padding} here. While mirror-padded input preserves the spatial structure of the central part, zero-padded input leads to a different spatial structure.}
\label{fig:spt-extp-01}
\end{figure*}

\noindent\textbf{Spatial Extrapolation: } Our approach allows us to do a spatial extrapolation as shown in Figure~\ref{fig:spt-extp-01}. Given a $256\times512$ image, we spatially extrapolate on its edges to create a $512\times1024$ image. To do this, we mirror-pad the image to the target size and feed it to the video-specific autoencoder. We use the iterative reprojection property of the autoencoder. After a few iterations, the video-specific autoencoder generates continuous spatial outputs. We also show the results when zero padding the input image. We observe that mirror-padded input preserves the spatial structural of the central part, i.e., the actual input, whereas zero-padded input leads to a different spatial structure. Instead of mirror-padding, one may also place known content from the videos and yet be able to get a valid output.

\begin{figure*}
\includegraphics[width=\linewidth]{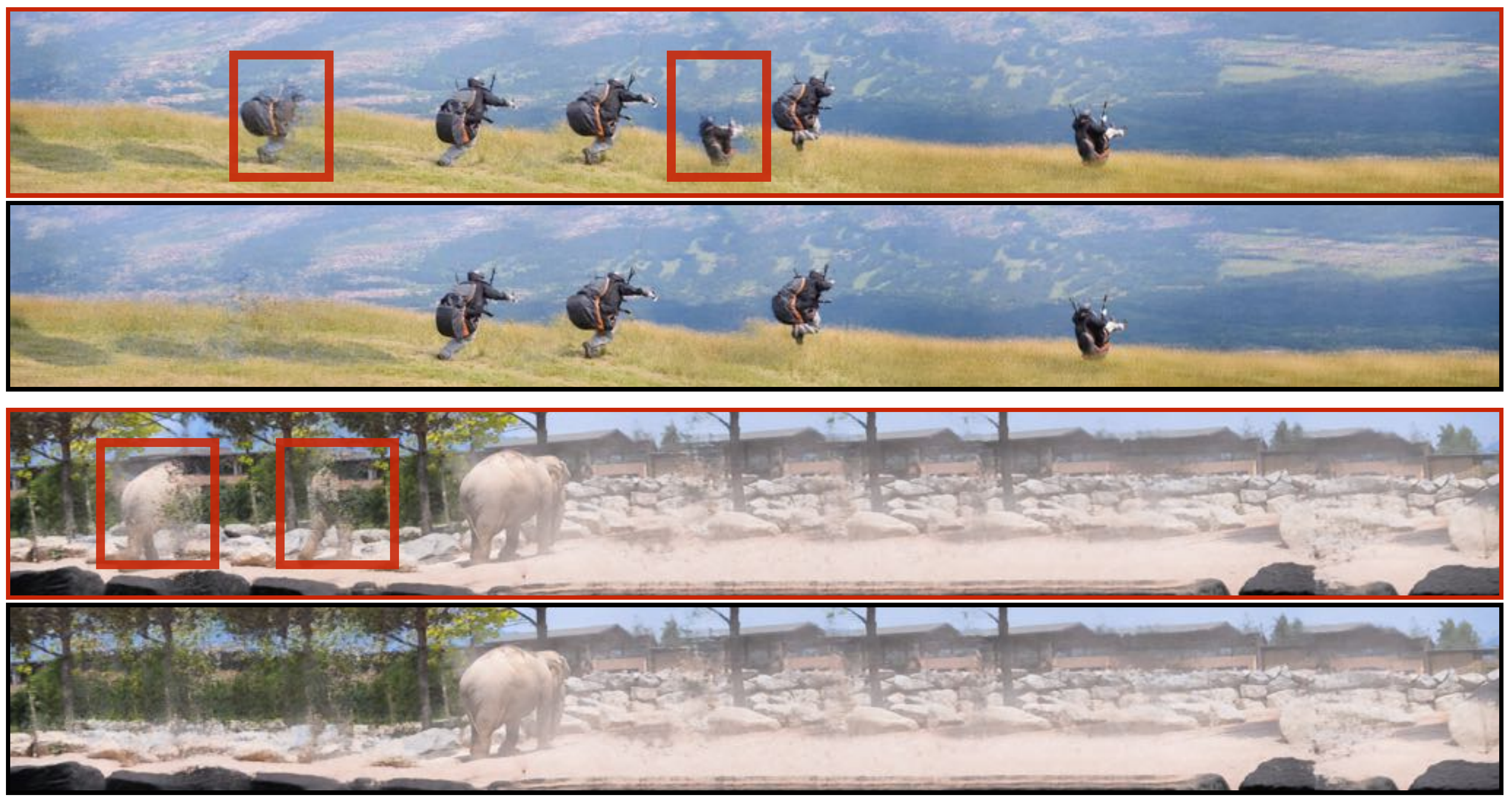}
\includegraphics[width=\linewidth]{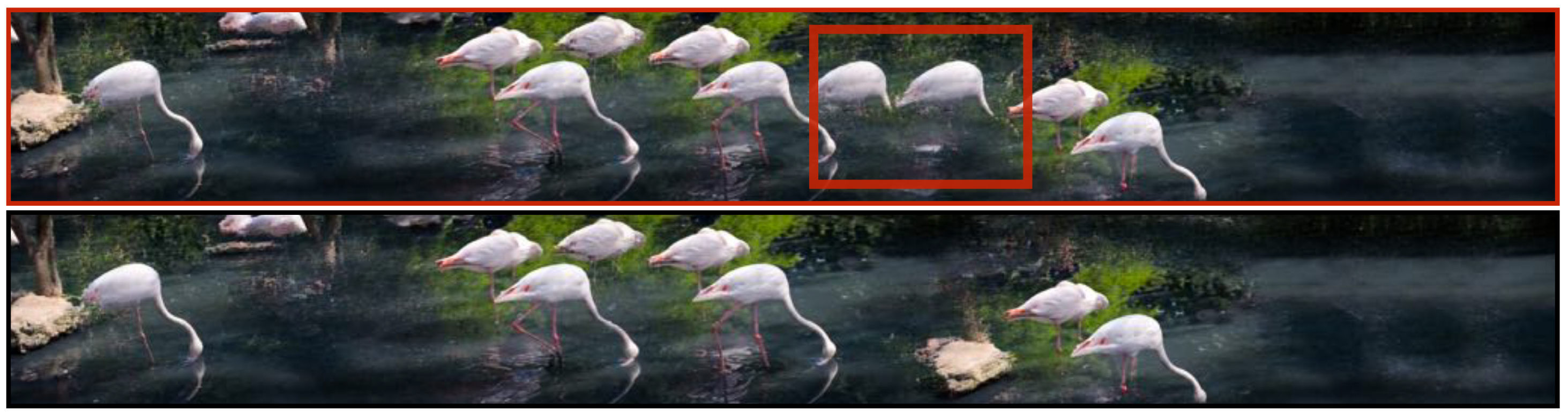}
\caption{\textbf{Spatial Editing (Removal): } Given a image, a user can do editing by copy-pasting a patch from surroundings to the target location and feed it to the video-specific autoencoder. We show examples (input-output pairs
) of spatial editing for stretched out images. The video-specific autoencoder yields a continuous and consistent spatial outputs as shown in the various examples here.}
\label{fig:spt-rem}
\end{figure*}

\noindent\textbf{User-Controlled Editing: } The ability of video-specific autoencoder to generate continuous outputs from a noisy input allows avenues for user-controlled editing. We show this aspect in Figure~\ref{fig:spt-rem} on stretched-out video frames as it allows us to make multiple edits.  Given an image, a user can do the editing by copy-pasting a patch from surroundings to the target location and feed it to the video-specific autoencoder. We can naively replace a forgeround object with a background patch, and the video-specific autoencoder will automatically correct the imperfections. We get consistent results due to the reprojection property of the autoencoder and its ability to generate continuous spatial image.

\begin{figure*}
\includegraphics[width=\linewidth]{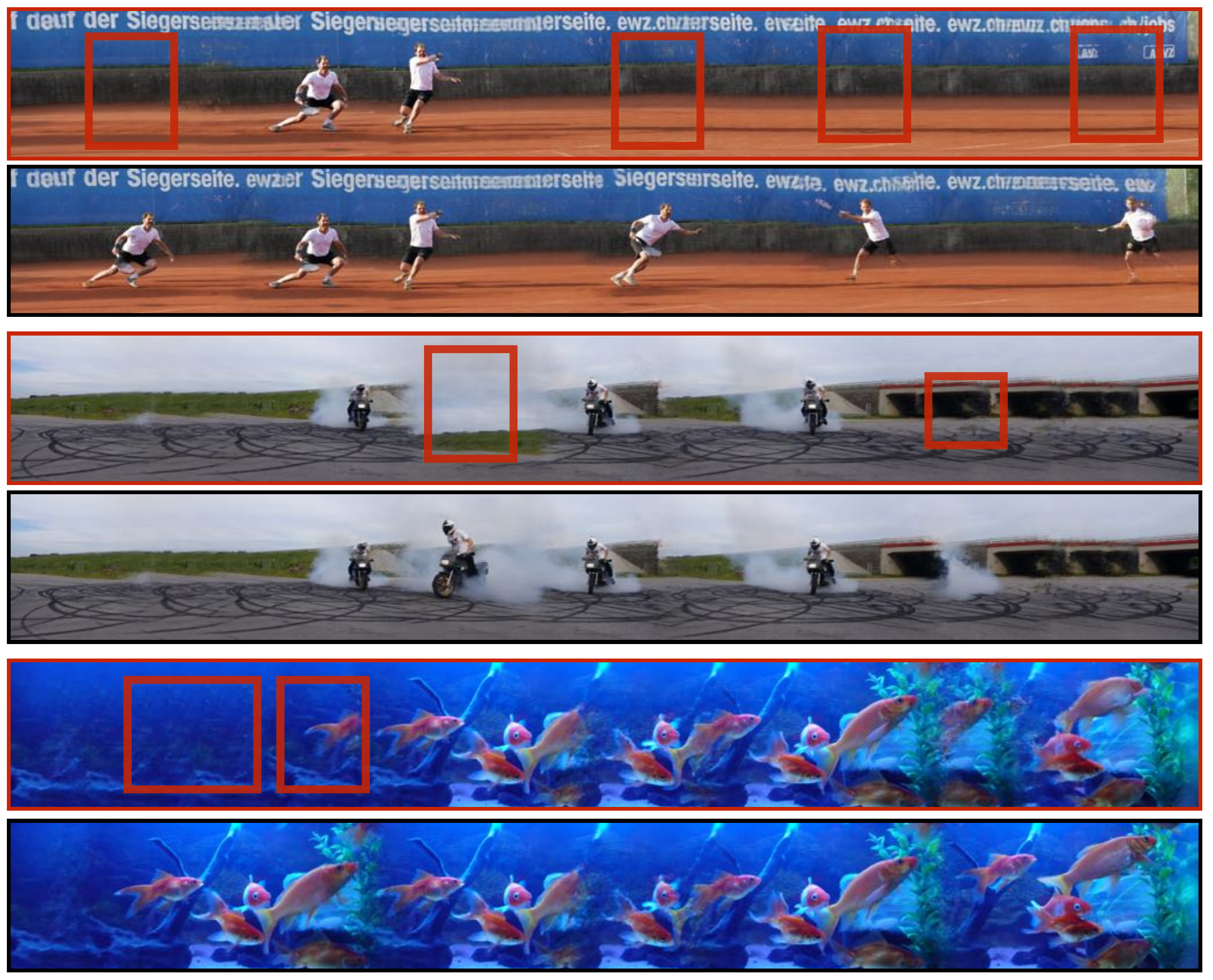}
\caption{\textbf{Spatial Editing (Insertion): } We show more examples (input-output pairs
) of spatial editing for stretched out images. Here a user inserts patches from far-apart frames and feed it to the video-specific autoencoder. The autoencoder generates a continuous spatial output. However, there are no guarantees if the video-specific autoencoder will preserve the input edits. It may generate a completely different output at that location. For e.g., we placed small fishes in the bottom example but the video-specific autoencoder chose to generate different outputs. }
\label{fig:spt-ins}
\end{figure*}

\noindent\textbf{Insertion: } We can also use the video-specific autoencoder to insert the known content from the video in a frame as shown in Figure~\ref{fig:spt-ins}. A user inserts patches from the far-apart frames and feed it to the video-specific autoencoder. We use the iterative reprojection property here. The autoencoder generates a continuous and seamless spatial output. We, however, observe that there are no guarantees if the video-specific autoencoder (trained on a single scale) will preserve the input edits all the times when foreground/moving objects are inserted. It is very likely that it may generate a completely different output at that location. In the bottom row of Figure~\ref{fig:spt-ins}, we inserted a number of small fish. The video-specific autoencoder, however, chose to generate different outputs. This behaviour would require more analysis of the reprojection property and we leave it to the future work.

\begin{figure*}
\centering
\includegraphics[width=\linewidth]{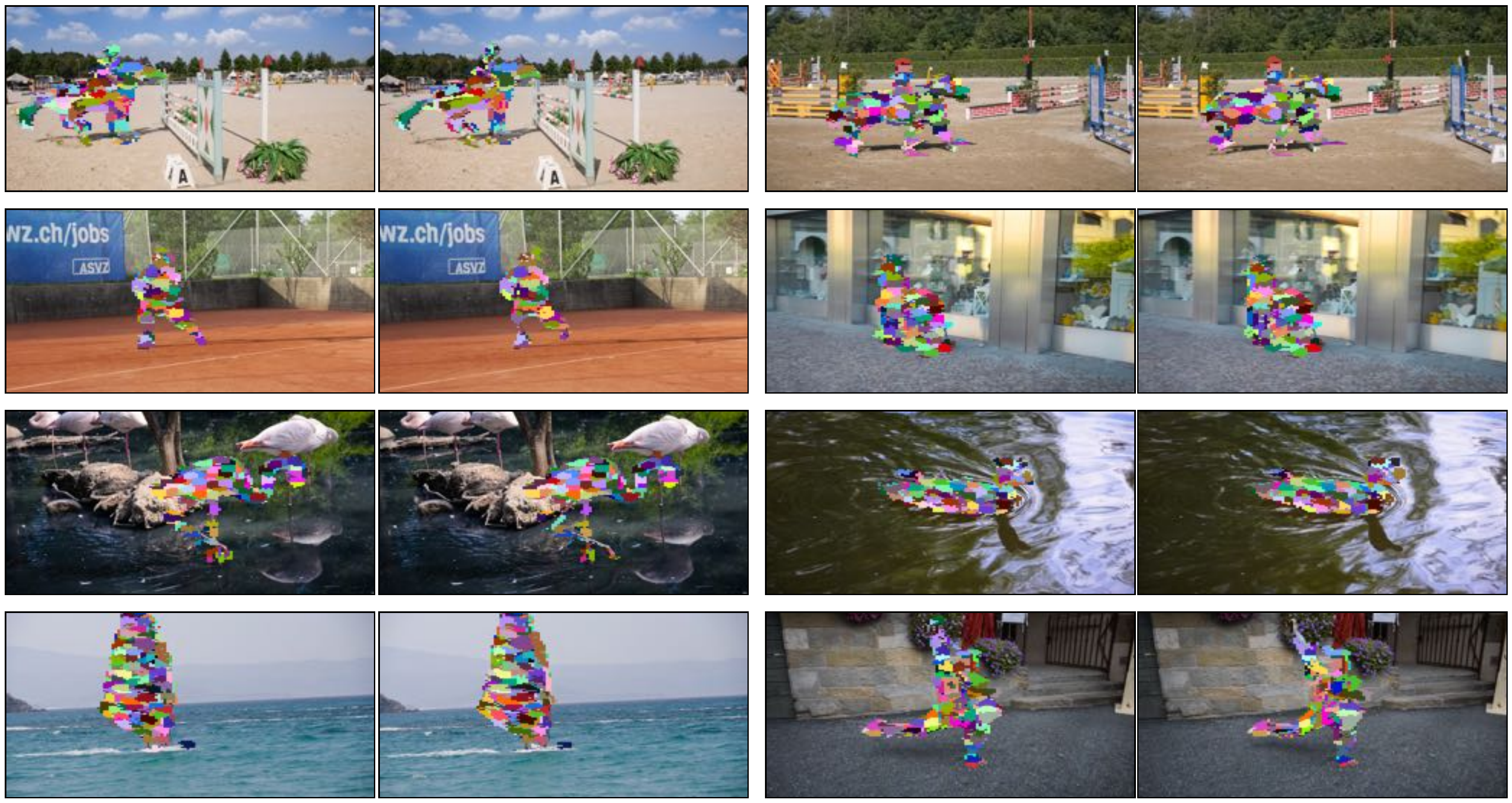}
\includegraphics[width=\linewidth]{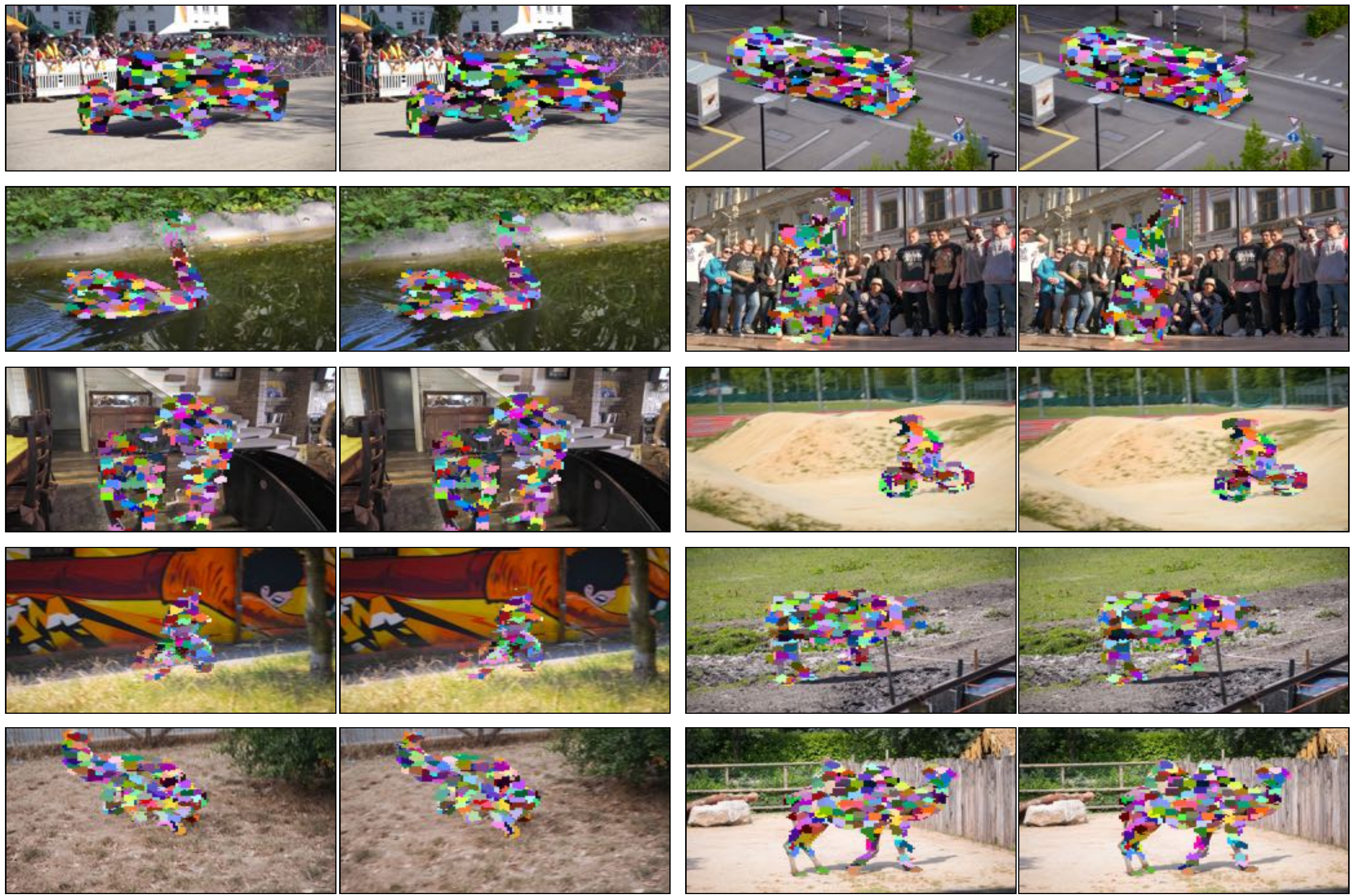}
\caption{\textbf{Correspondences across Adjacent Frames } We show correspondences established in the adjacent frames of a video using pixel-level codes. For the sake of visualization, we select $256$ random regions on the moving objects in a scene.}
\label{fig:pixel-corr}
\end{figure*}

\begin{figure*}
\centering
\includegraphics[width=\linewidth]{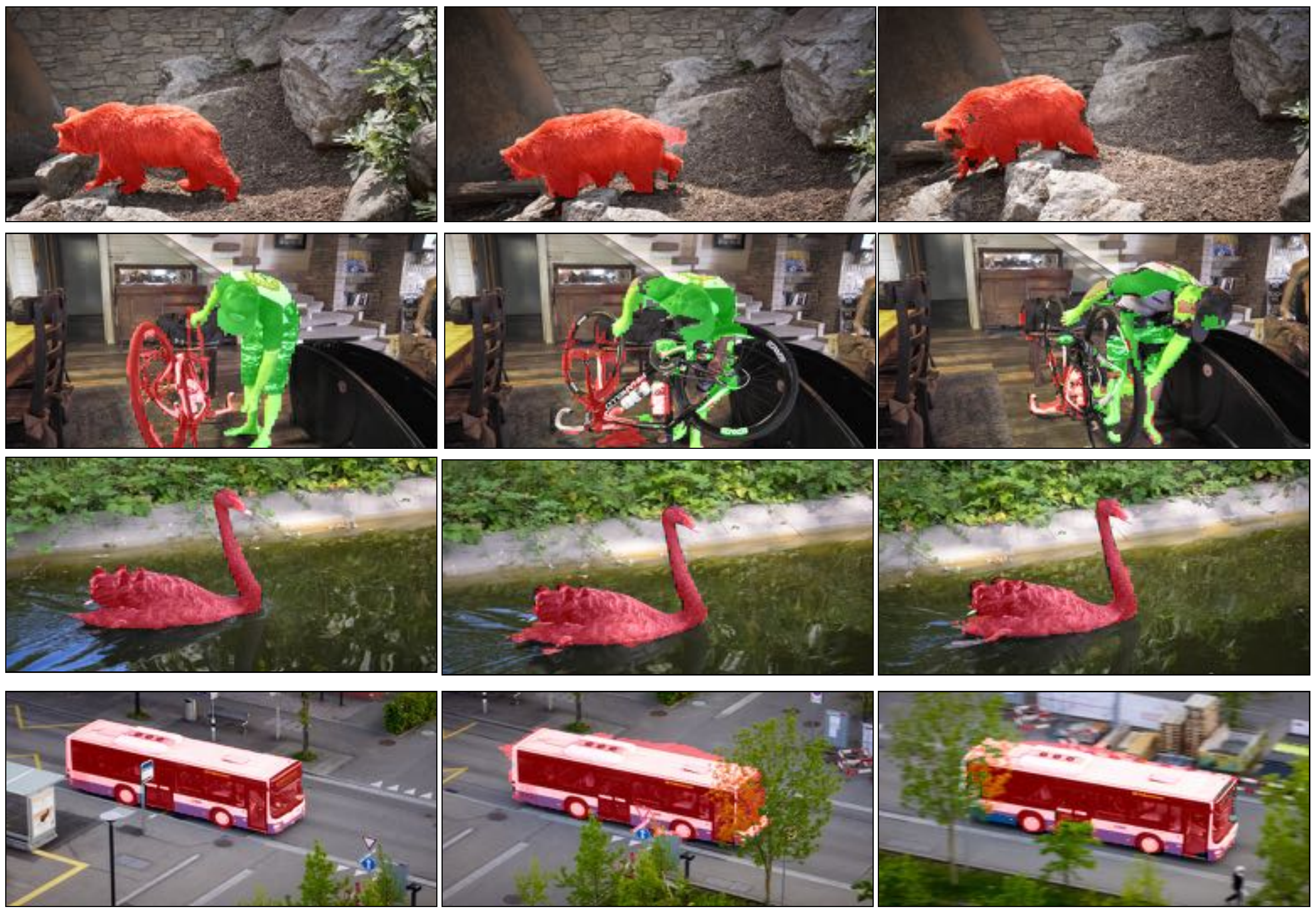}
\includegraphics[width=\linewidth]{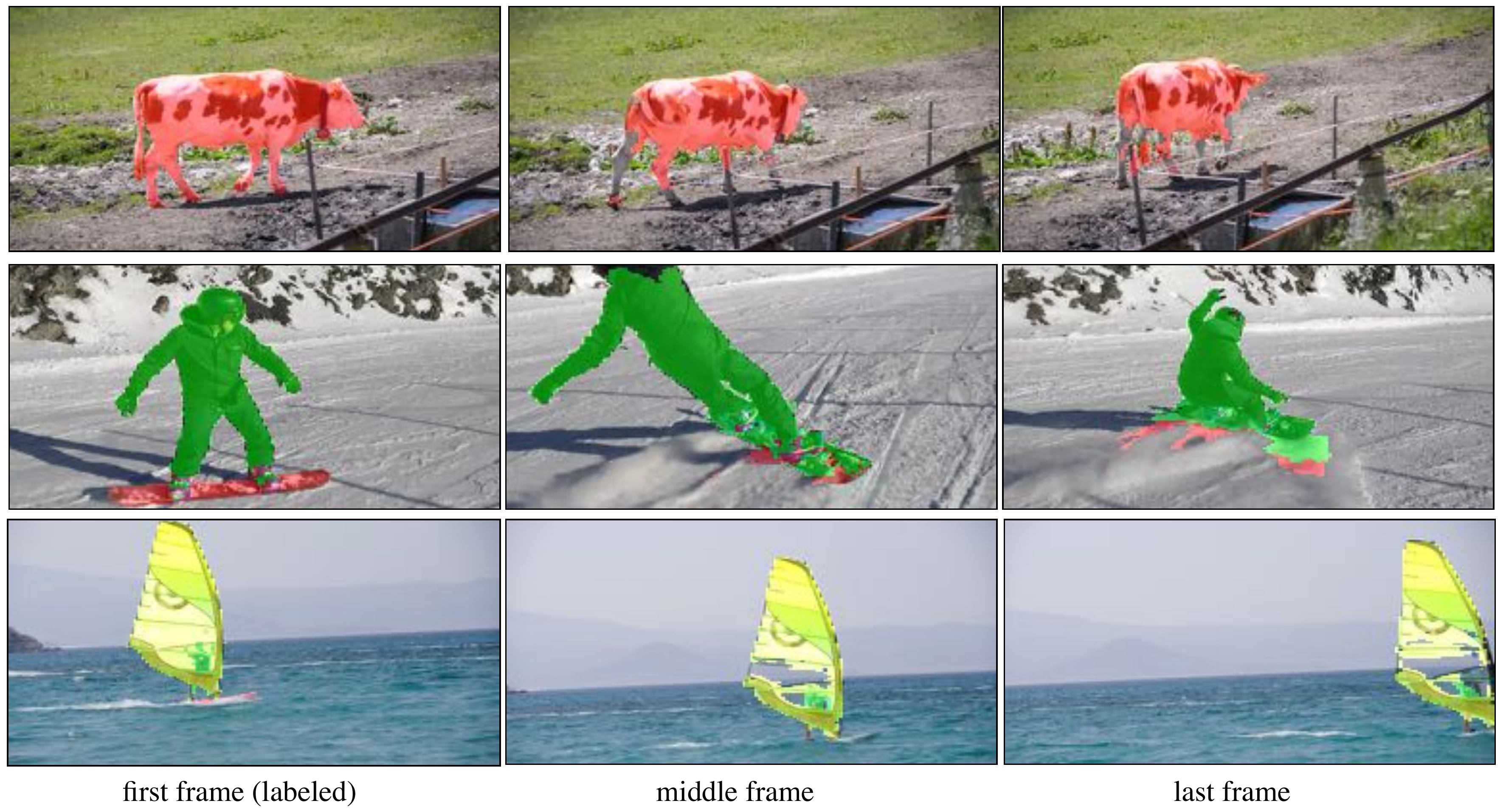}
\caption{\textbf{Instance Mask Propagation: } Given the first labeled frame, we propagate the instance labels using the pixel codes. We show here the labels in the first frame on the left. We show the propagated labels from the first frame to the middle and last frame of the video.}
\label{fig:maskp-01}
\end{figure*}

\begin{figure*}
\centering
\includegraphics[width=\linewidth]{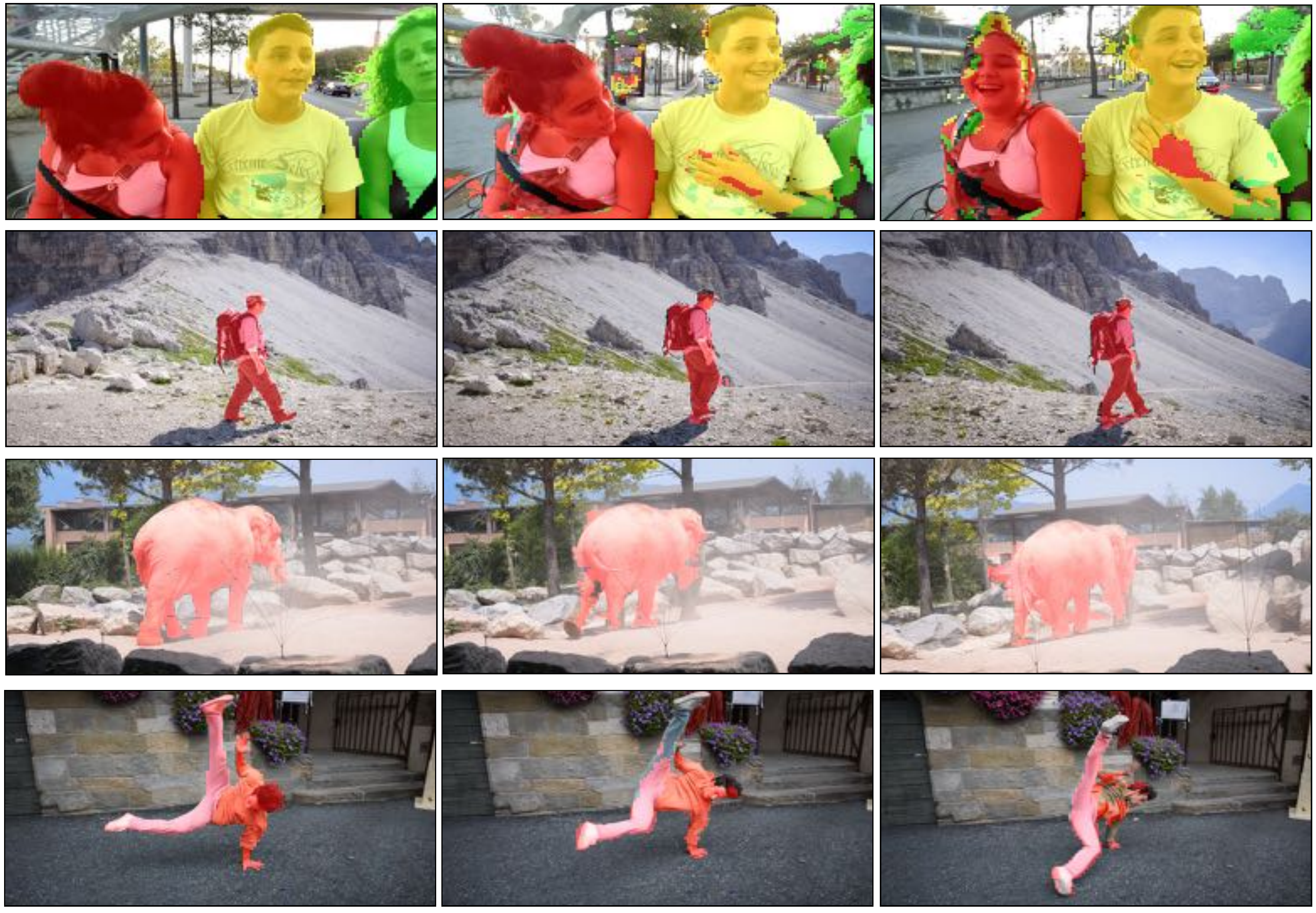}
\includegraphics[width=\linewidth]{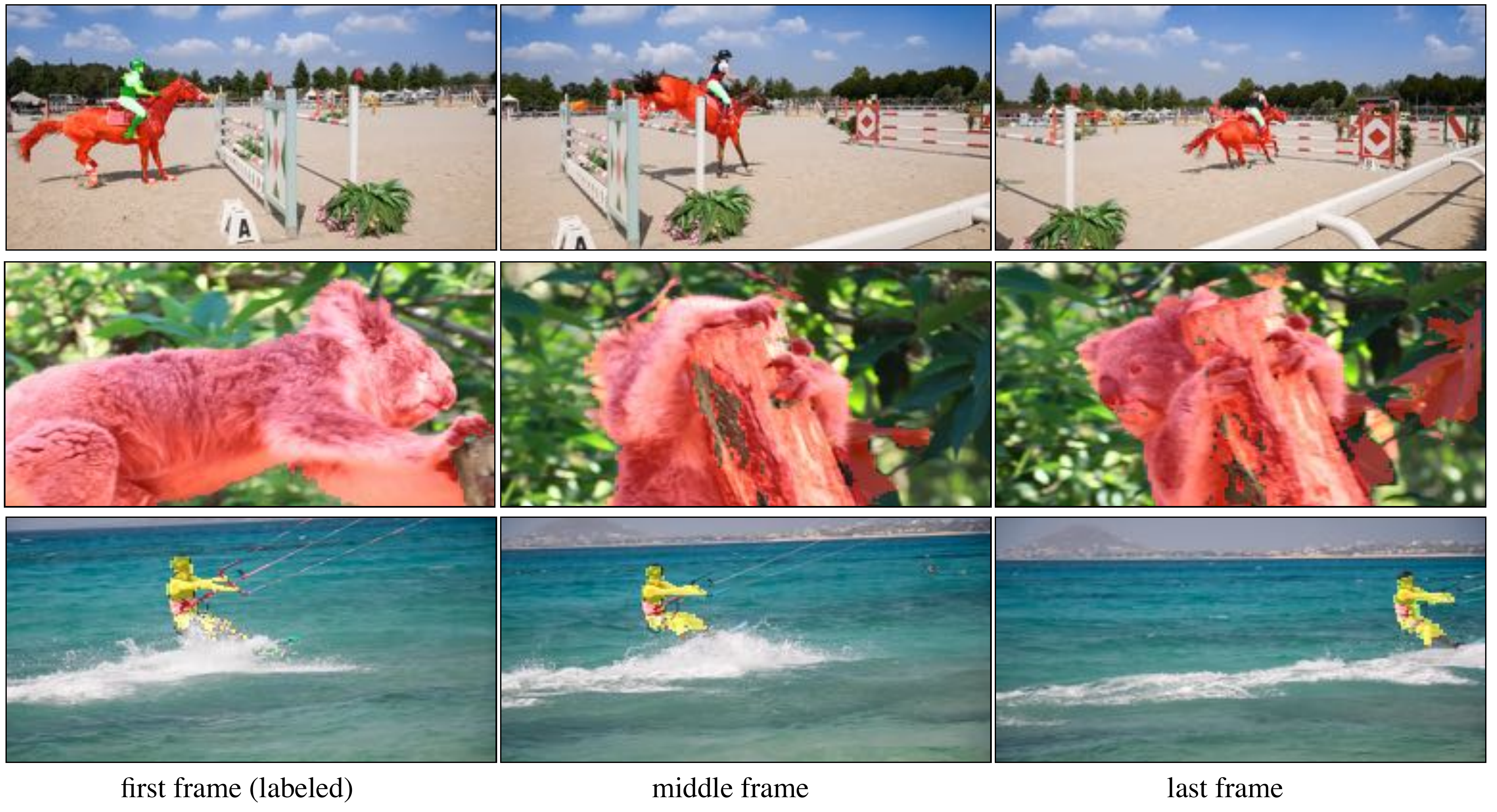}
\caption{\textbf{Instance Mask Propagation: } Given the first labeled frame, we propagate the instance labels using the pixel codes. We show here the labels in the first frame on the left. We show the propagated labels from the first frame to the middle and last frame of the video.}
\label{fig:maskp-02}
\end{figure*}

\noindent\textbf{Pixel Correspondences: } Finally, we show the results of pixel correspondences across adjacent frames in Figure~\ref{fig:pixel-corr}. We show examples of instance mask propagation in Figure~\ref{fig:maskp-01} and Figure~\ref{fig:maskp-02}.

%% --------------------------- %%

%% ---------------------------- %%%

\section{Efficiently Transmitting a Video}
\label{appd:transmit}

\begin{figure*}
\includegraphics[width=\linewidth]{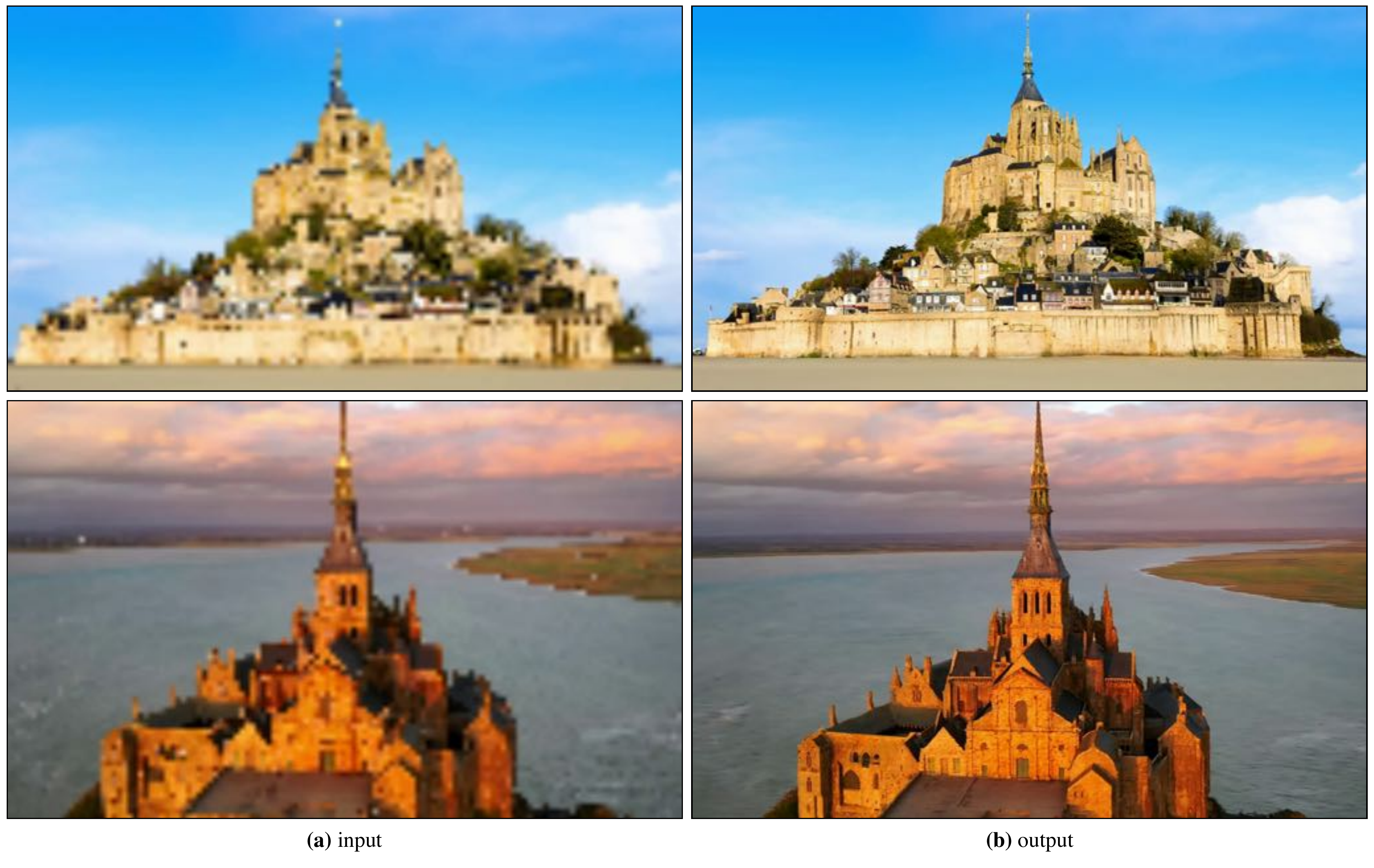}
\caption{\textbf{10X Spatial Super Resolution: } We train video-specific autoencoders using original videos. The reprojection property allows us to get hi-res outputs even when inputting a low-res sample at test time.}
\label{fig:sres-01}
\end{figure*}

\begin{figure*}
\includegraphics[width=\linewidth]{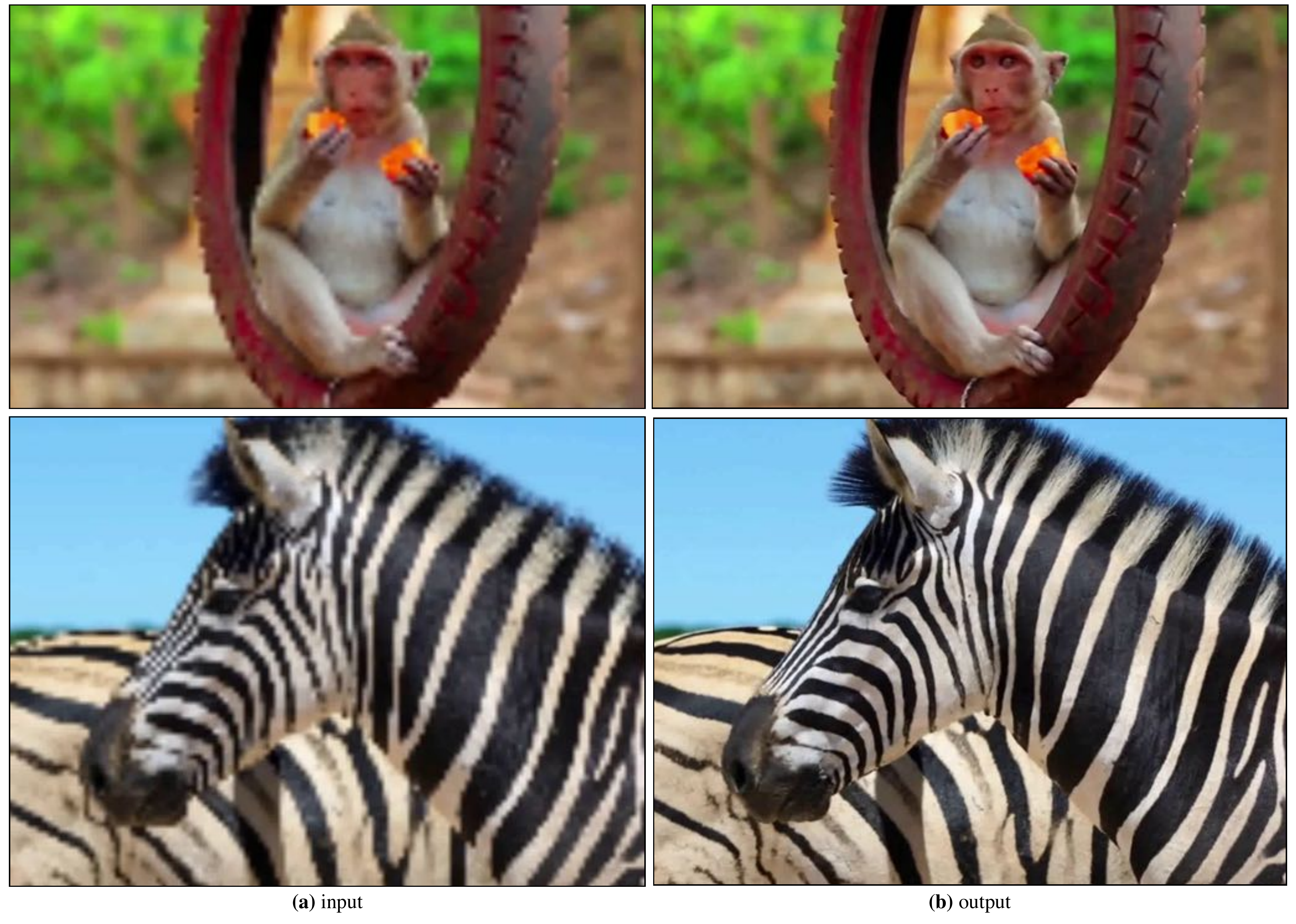}
\caption{\textbf{10X Spatial Super Resolution: } We show more examples of 10X super-resolution with various animal videos. Once trained, the reprojection property of the autoencoder allows us to input low-res samples and get hi-res details in the outputs.}
\label{fig:sres-02}
\end{figure*}

We show examples of 10X super-resolution in Figure~\ref{fig:sres-01}-\ref{fig:sres-02} using this property for hi-res videos. The convolutional autoencoder allows us to use videos of varying resolution. Crucially, we get temporally smooth outputs without using any temporal information for all the experiments.

\begin{figure*}
\includegraphics[width=\linewidth]{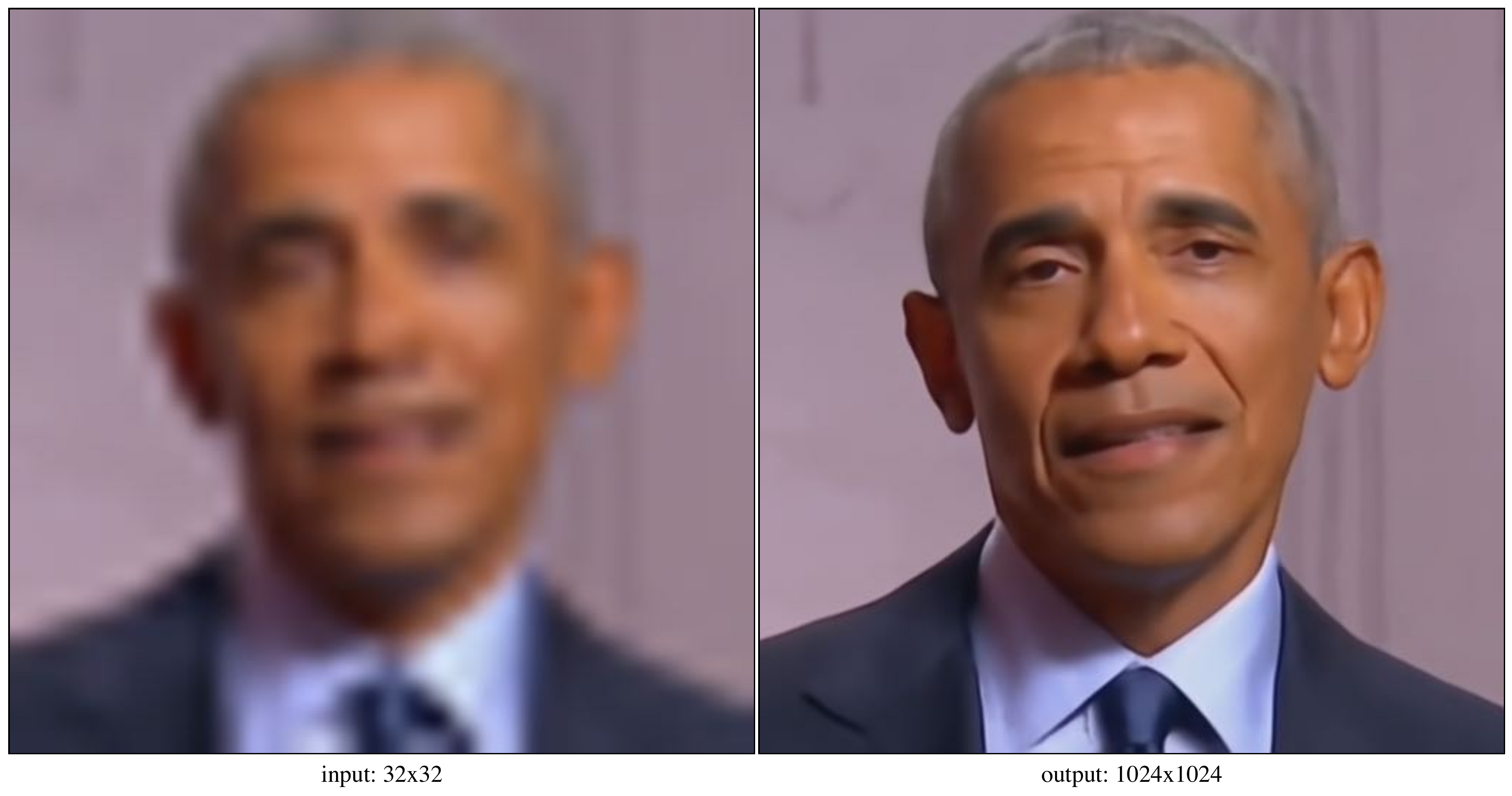}
\caption{\textbf{32X Spatial Super Resolution without Iterative Reprorjection: } We demonstrate that one can also train an autoencoder with noisy or low-res input samples during training. This enables us to input very low resolution inputs at test time, and get a hi-res output in a single iteration. We train an autoencoder using the low-res $32\times32$ input and a hi-res $1024\times1024$ output. This enables the model to be robust to low-res samples. We show the result of our approach on an unseen $32\times32$ input that yields a $1024\times1024$ output.}
\label{fig:sres-03}
\end{figure*}

\noindent\textbf{Robust Space for Low-Res Inputs: } One method to deal with extremely low-res inputs is to use the reprojection property iteratively. We also observe that one can \emph{robustify} the space of video-specific autoencoder by varying the resolution of input images at training time (the resolution of output does not change). This added noise makes the model robust to low-res noise and allows us to get hi-res outputs for a extremely low-res input without iterative reprojection. Figure~\ref{fig:sres-03} shows an example of 32X super-resolution in one iteration. We show example where we input $32\times32$ Barack Obama frames and yet be able to get $1024\times1024$ hi-res output in a single iteration. We have not use this aspect in any other example/evaluation shown in this paper. We leave it to the future work for more exploration of this property.

\begin{figure*}[t]
\centering
\includegraphics[width=\linewidth]{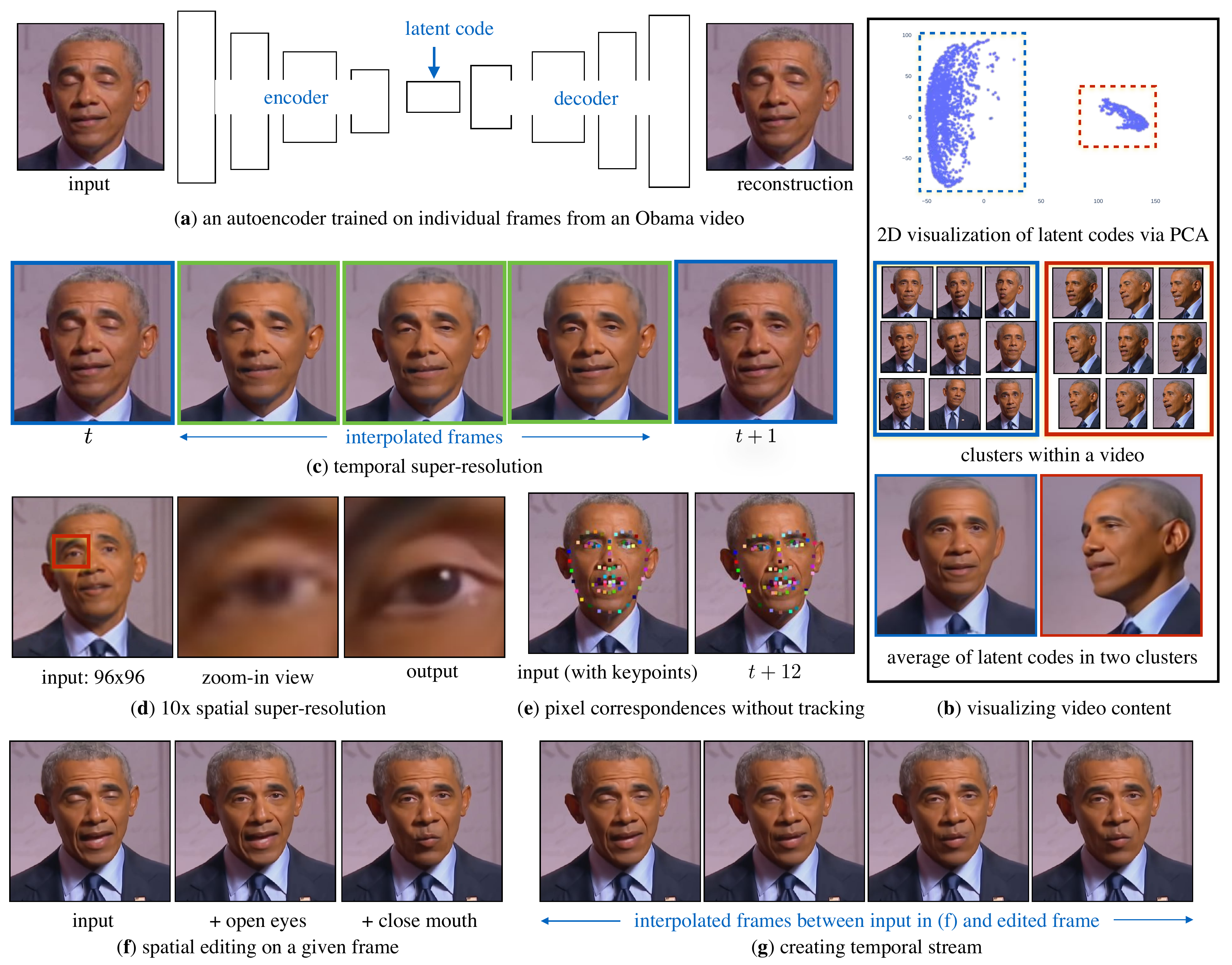}
\caption{\textbf{Summary: What can we do with a video given a single video representation?} \textbf{(a)} We train a video-specific autoencoder on individual frames from a target video (here, 300 unordered $1024\times1024$ frames). \textbf{(b)} We use Principal Component Analysis (PCA) to \textbf{visualize latent codes} of the frames of this video in a 2D space. This 2D visualization shows two different data distributions within the video. We visualize the \textbf{clusters within the video}  and \textbf{average of latent codes} in these two clusters. \textbf{(c)} We interpolate the latent codes of adjacent frames (and decode them) for \textbf{temporal super-resolution}.  \textbf{(d)} By linearly upsampling low-res $96\times96$ image frames to $1024\times1024$ blurry inputs and passing them through the autoencoder, we can ``project" such noisy inputs into the high-res-video-specific manifold, resulting in high quality 10X {\bf super-resolution}, even on subsequent video frames not used for training. The ability to do temporal and spatial super-resolution allows us to transmit sparse low-res frames over the network and still get dense hi-res outputs at the reception. \textbf{(e)} We use hypercolumn features from the encoder and can do \textbf{pixel-level correspondences} between two frames in a video.  \textbf{(f)} We can also do \textbf{spatial editing} on a given frame of video. Shown here is an input frame where eyes are closed. We copy open eyes from another frame and close mouth from a yet another frame, and pass it through the autoencoder to get a consistent output. \textbf{(g)} We \textbf{create temporal stream} between the original frame and edited frame by interpolating the latent code. A single representation learned on unordered video frames allows us to explore the contents of a video, edit them, and efficiently transmit a video over the network. Importantly, we did not optimize for any of the above task. }
\label{fig:teaser-01}
\end{figure*}

\begin{figure*}[t]
\centering
\includegraphics[width=\linewidth]{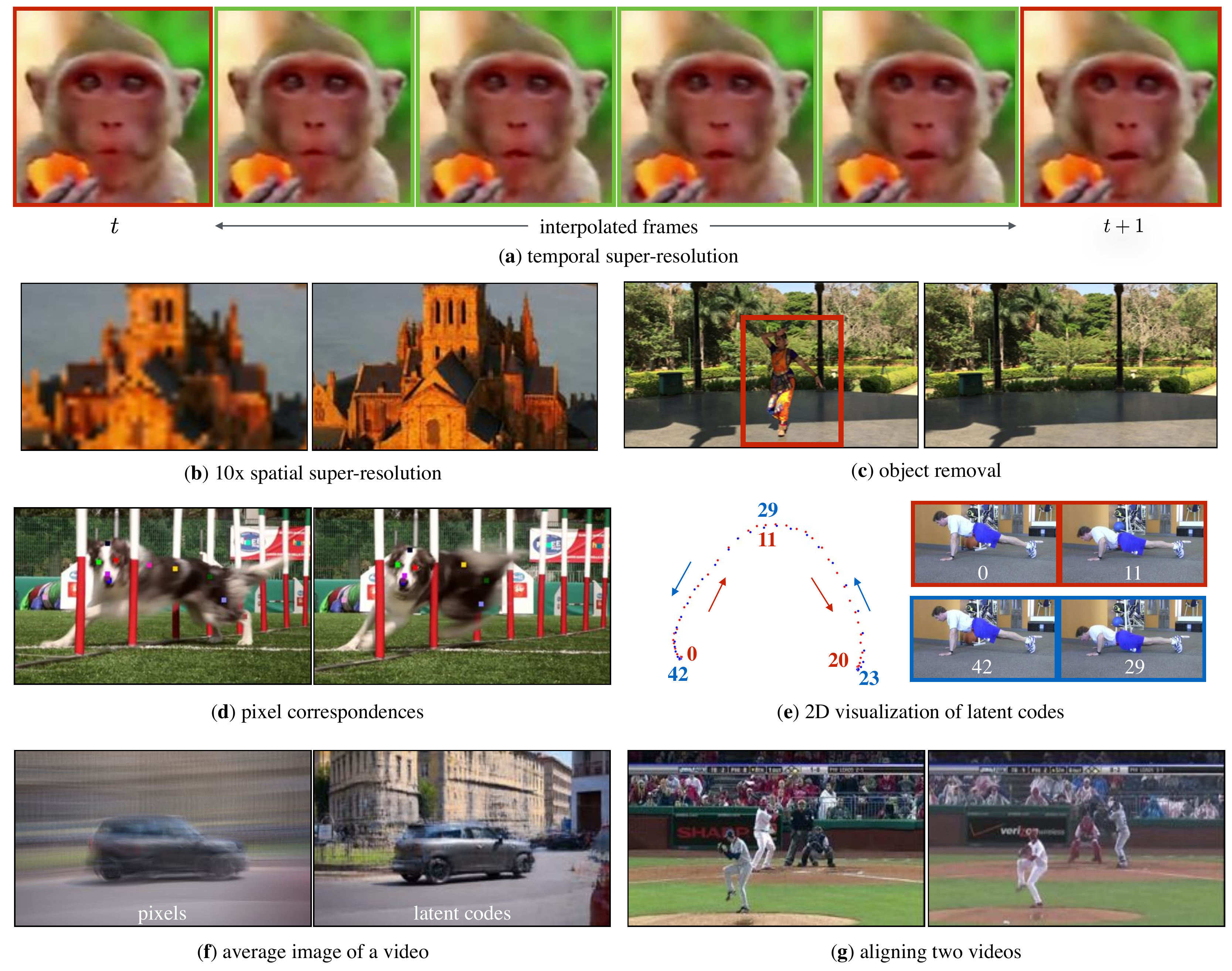}
\caption{\textbf{Different applications on various videos: } We demonstrate a remarkable number of video processing tasks enabled by a simple video-specific representation: an {\em image} autoencoder trained on frames from a target video. \textbf{(a)} By interpolating latent codes of adjacent frames (and decoding them), one can perform \textbf{temporal super-resolution}. \textbf{(b)} By linearly upsampling low-res $96\times96$ image frames to $1024\times1024$ blurry inputs and passing them through the autoencoder, we can {\em project} such blurry inputs into the high-res-video-specific manifold, resulting in high quality \textbf{10X super-resolution}, even on subsequent video frames not used for training. \textbf{(c)} Manifold projection can also be used to \textbf{remove objects} marked with a bounding box. \textbf{(d)} Hypercolumn features of the autoencoder can be used to establish \textbf{per-pixel correspondences} across frames of a video via feature matching. \textbf{(e)} Multidimensional scaling of latent codes (via 2D Principal Component Analysis) allows for ``at-a-glance" interactive video exploration; one can summarize visual modes and discover repeated frames that look similar but are temporally distant. \textbf{(f)} We can {\bf summarize} videos by decoding the average latent codes (of all the frames in a video), which compares favorably to a naive per-pixel average image. \textbf{(g)} Manifold projection allow us to \textbf{align} two semantically similar videos and \textbf{retarget} from one to another (here, two baseball games).}
\label{fig:teaser-end}
\end{figure*}

\noindent\textbf{Summary: } We summarize different operations that we can perform on a video using a single representation in Figure~\ref{fig:teaser-01}. Finally, we show different application on various videos in Figure~\ref{fig:teaser-end}. 

\section{More Discussion on Properties}

\begin{figure*}[t]
\centering
\includegraphics[width=\linewidth]{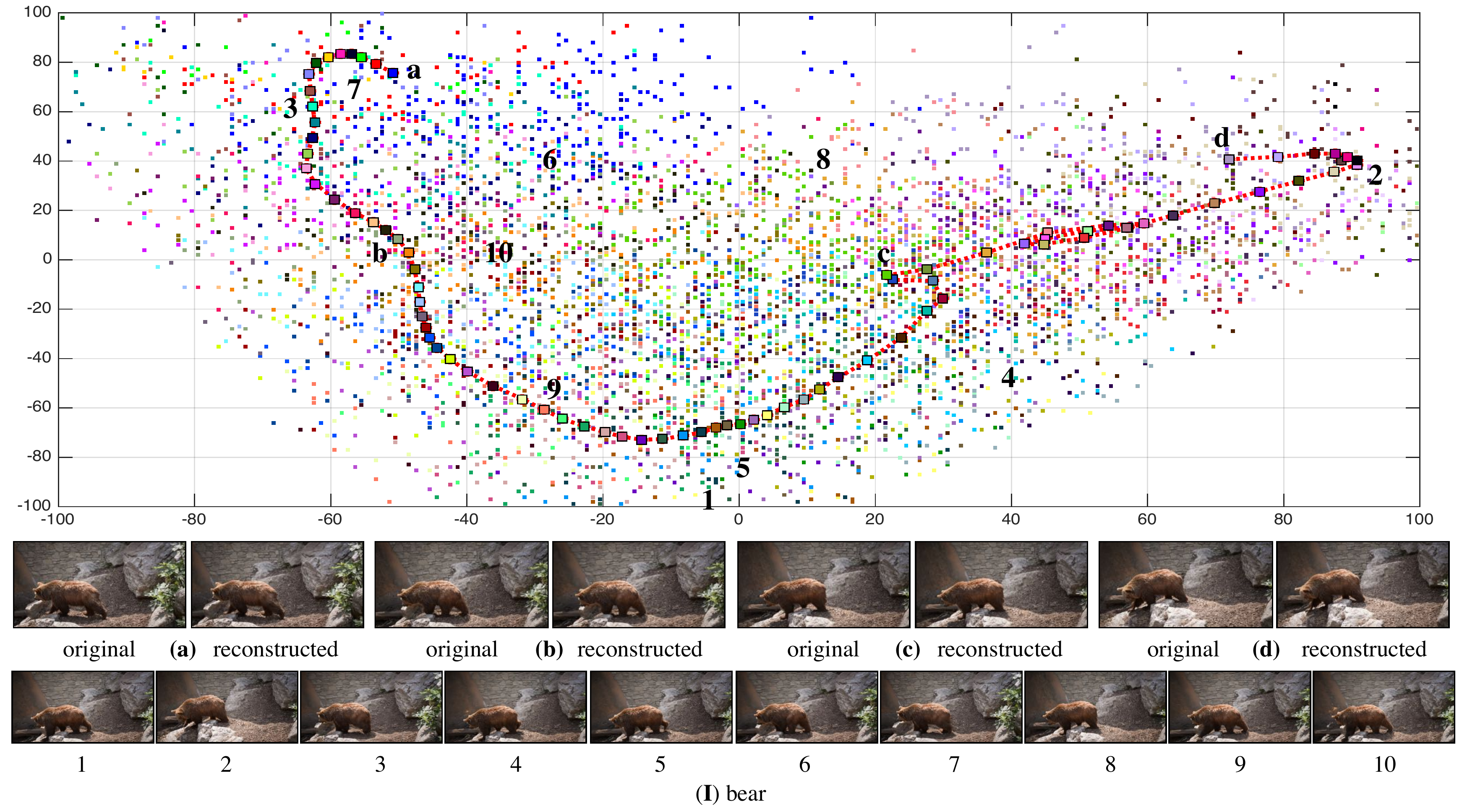}
\includegraphics[width=\linewidth]{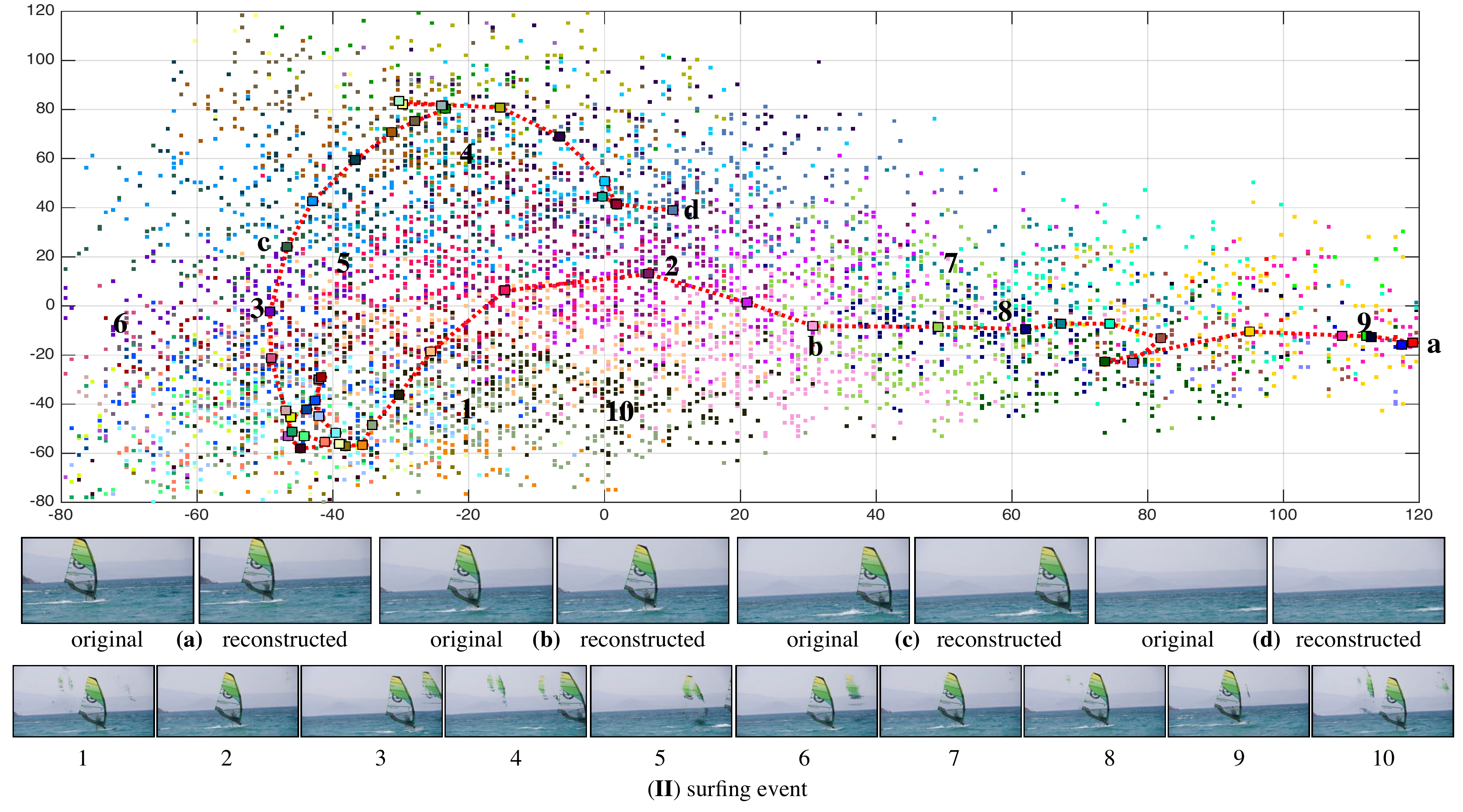}
\caption{\textbf{Exploring the Manifold}:  We train two video instance specific autoencoders on: \textbf{(I)} $82$ individual frames from a bear sequence; and \textbf{(II)} $55$ individual frames from a surfing event. We visualize the latent codes of original points on a 2D plot for two sequences using the bold squares. Each square represents an original frame in the video and is shown using a different color. The red line connecting the squares show the temporal sequence. We show \textbf{original} images and \textbf{reconstructed} images for four points: \textbf{(a)}, \textbf{(b)}, \textbf{(c)}, and \textbf{(d)}. We show random points on manifold $M$ colored by the closest original frame. We visualize image reconstruction of a random subset of 10 in the bottom row. Note that latent coordinates even far away from the original frames tend to produce high-quality image reconstructions.}
\label{fig:manifold-exp}
\end{figure*}

\noindent\textbf{Exploring the Manifold: } Video-specific manifold allows us to move in the 2D space and visualize the characterstics of a video. We explore the video-specific manifolds in Figure~\ref{fig:manifold-exp} via two video instance specific autoencoders on: \textbf{(I)} $82$ individual frames from a bear sequence; and \textbf{(II)} $55$ individual frames from a surfing event. The latent codes of original points on a 2D plot for two sequences are visualized using the bold squares. Each square represents an original frame in the video and is shown using a different color. The red line connecting the squares show the temporal sequence. We show \textbf{original} images and \textbf{reconstructed} images for four points: \textbf{(a)}, \textbf{(b)}, \textbf{(c)}, and \textbf{(d)}. We observe sharp reconstruction. We then show various points on this manifold. The color represents the closest original frame. We also show ten random points around original points in the latent space. We do not see artifacts for the bear sequence as we move away from the original points due to highly correlated frames. We see minor artifacts in the surfing event as we move away from the original points as the frames are sparse and spread-out.

\begin{figure*}
\centering
\includegraphics[width=0.9\linewidth]{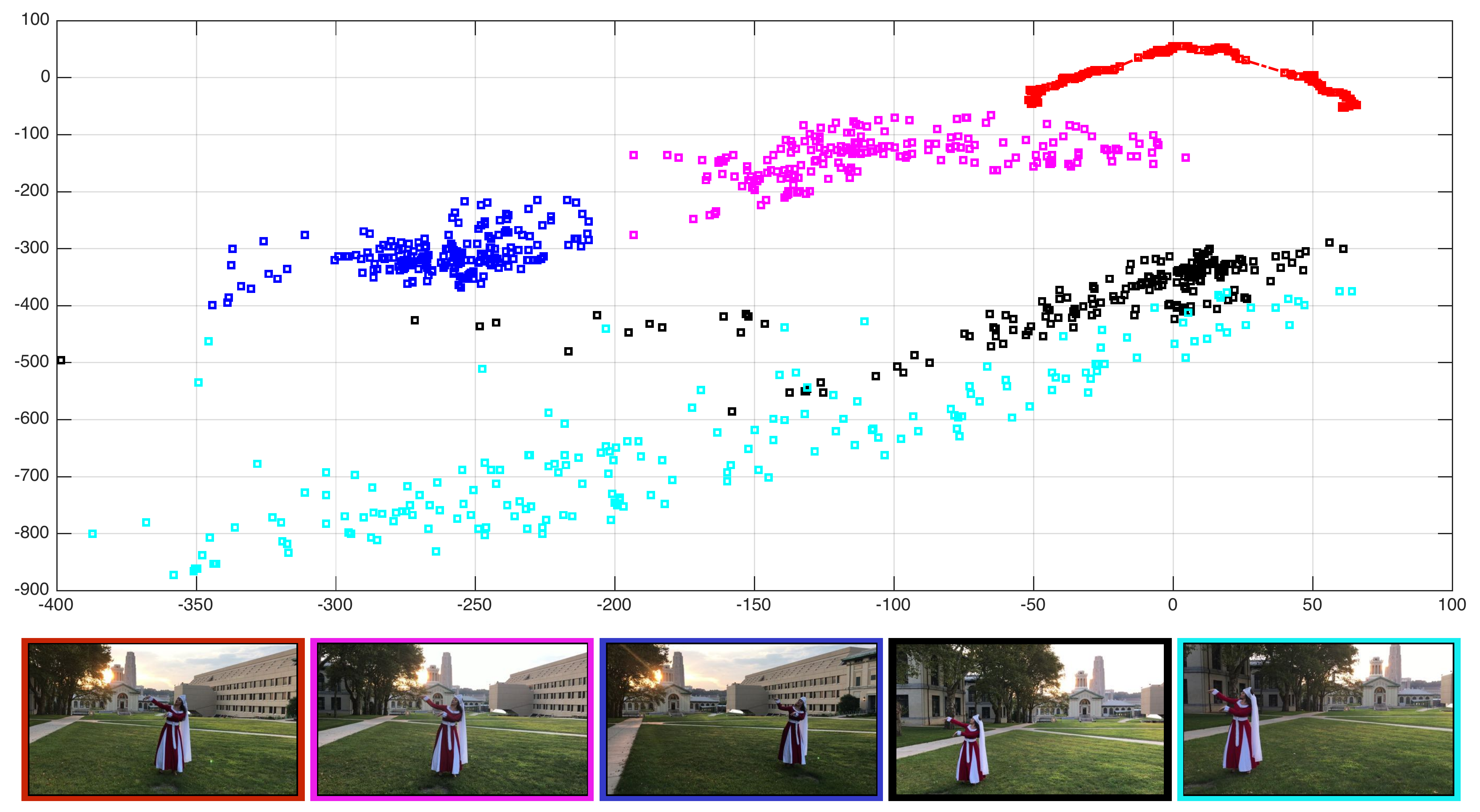}
\caption{\textbf{Perturbation via Multi-Views: } We train a video-specific autoencoder with a sequence captured via a stationary camera (a frame of the sequence as shown in the red box). The 2D visualization of this sequence is shown with the {\color{red}{red points}} in the visualization. Once trained, we take inputs from multi-views (shown by views from other colors). We observe that the points move farther away from the {\color{red}{red points}} as we move away from the original camera/sequence that was used for training the video-specific autoencoder. For e.g., {\color{magenta}{magenta points}} are close to the {\color{red}{red points}} because the frames from two sequences look more similar than the frames representing {\color{cyan}{cyan points}}. }
\label{fig:multi-view-pca}
\end{figure*}

\begin{figure*}
\centering
\includegraphics[width=0.9\linewidth]{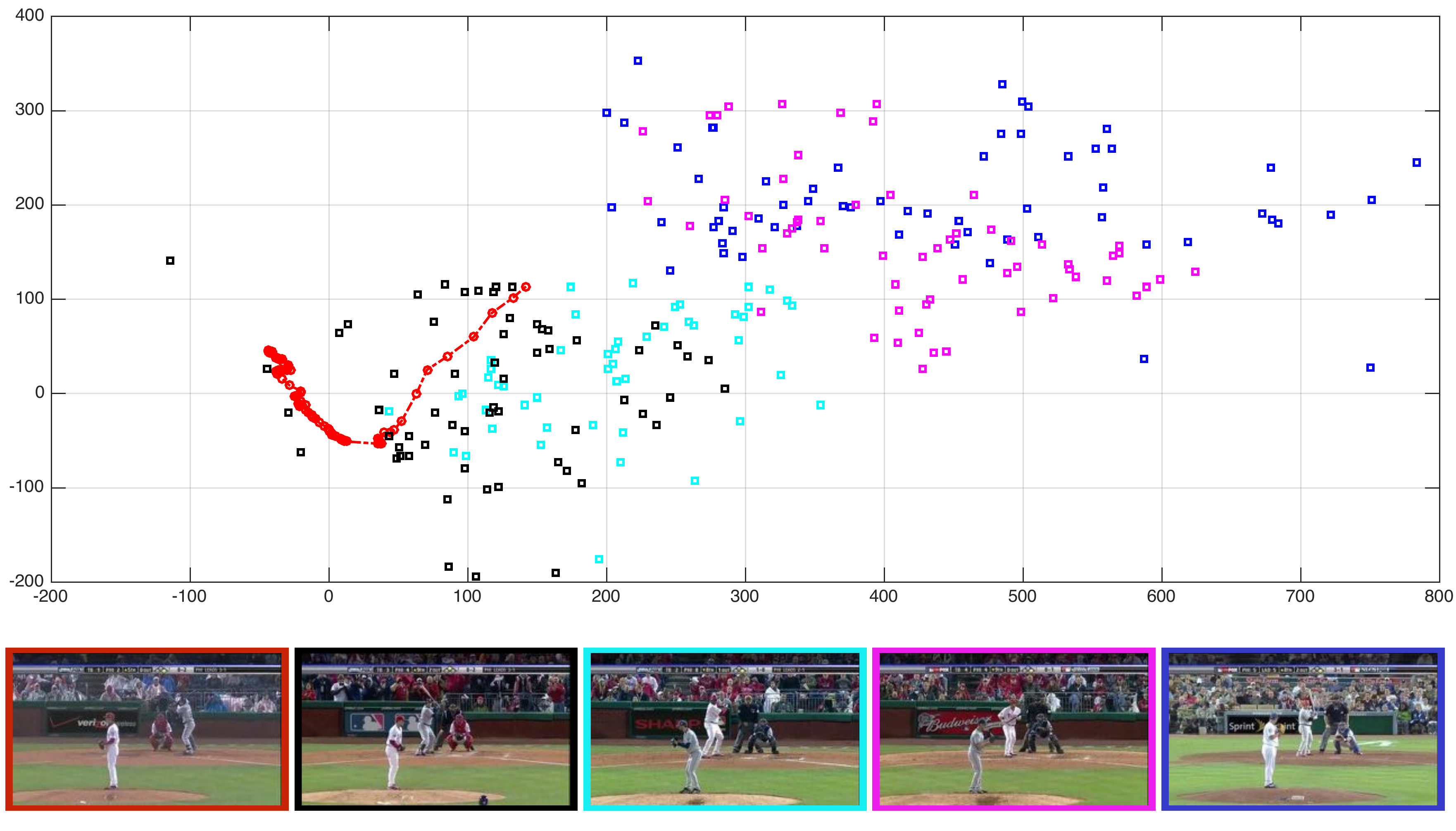}
\caption{\textbf{Perturbation via Semantically Similar Events: } We train a video-specific autoencoder with one baseball game video (a frame of the sequence is shown in the {\color{red}{red box}}). The 2D visualization of this sequence is shown with the {\color{red}{red points}} in the 2D visualization of latent codes. Once trained, we take inputs from other baseball games (shown by views marked with other colors). We observe that the points move farther from the {\color{red}{red points}} as the inputs become less similar to the original sequence.}
\label{fig:sem-pca-01}
\end{figure*}

\begin{figure*}
\centering
\includegraphics[width=\linewidth]{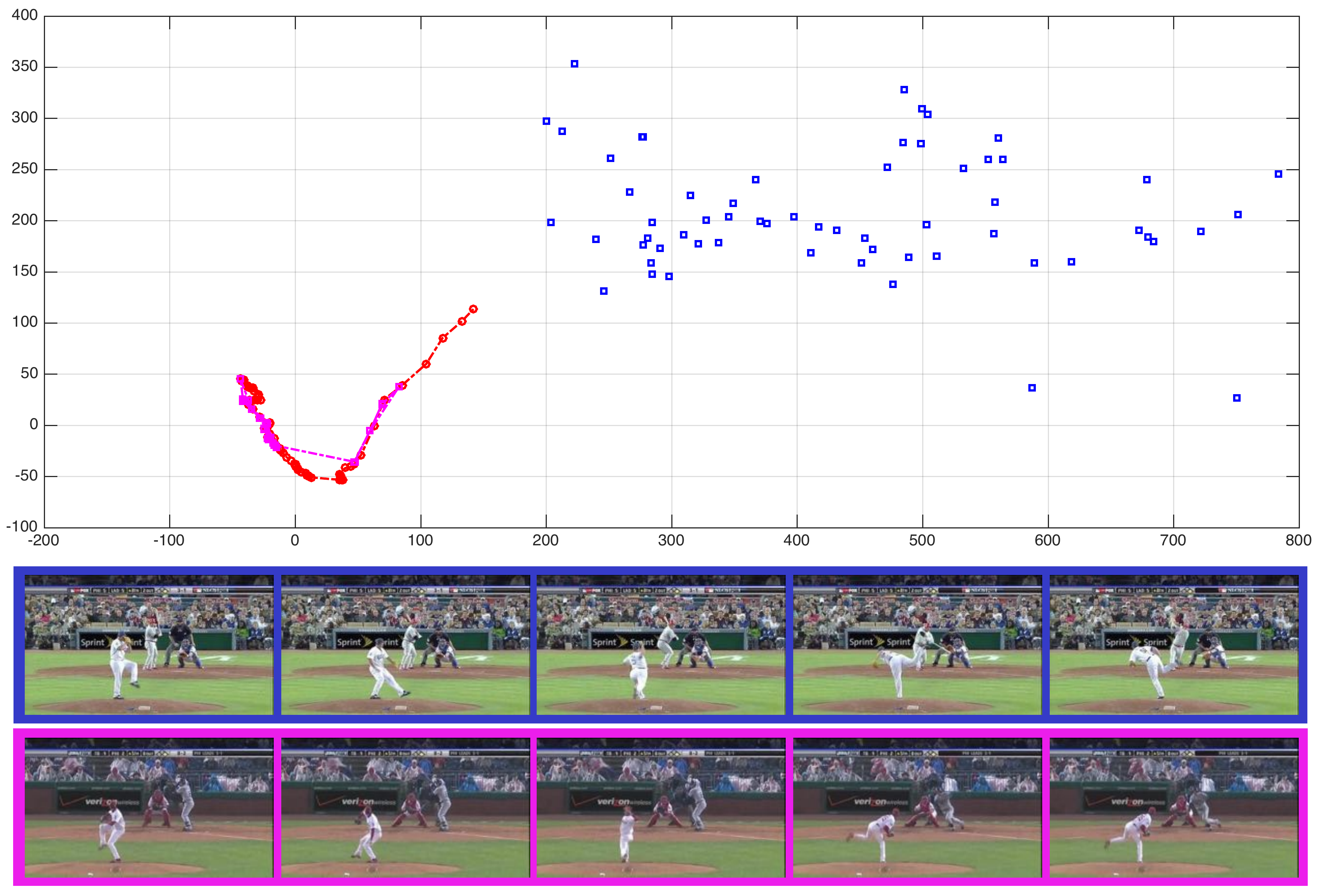}
\caption{\textbf{Iterative Reprojection of Semantically Similar Events: } We train a video-specific autoencoder with one baseball game video. The 2D visualization of this sequence is shown with the {\color{red}{red points}}. Once trained, we take inputs from other another baseball game (shown by {\color{blue}{blue points}}). We iteratively reproject the input and observe that one can align two semantically similar videos (shown by {\color{magenta}{magenta points}}). We also show a few example inputs from the ``unknown'' sequence and the reconstructed images (after $25^{th}$ iteration) from the autoencoder. We observe temporally coherent outputs.}
\label{fig:sem-pca-02}
\end{figure*}

\begin{figure*}
\centering
\includegraphics[width=0.87\linewidth]{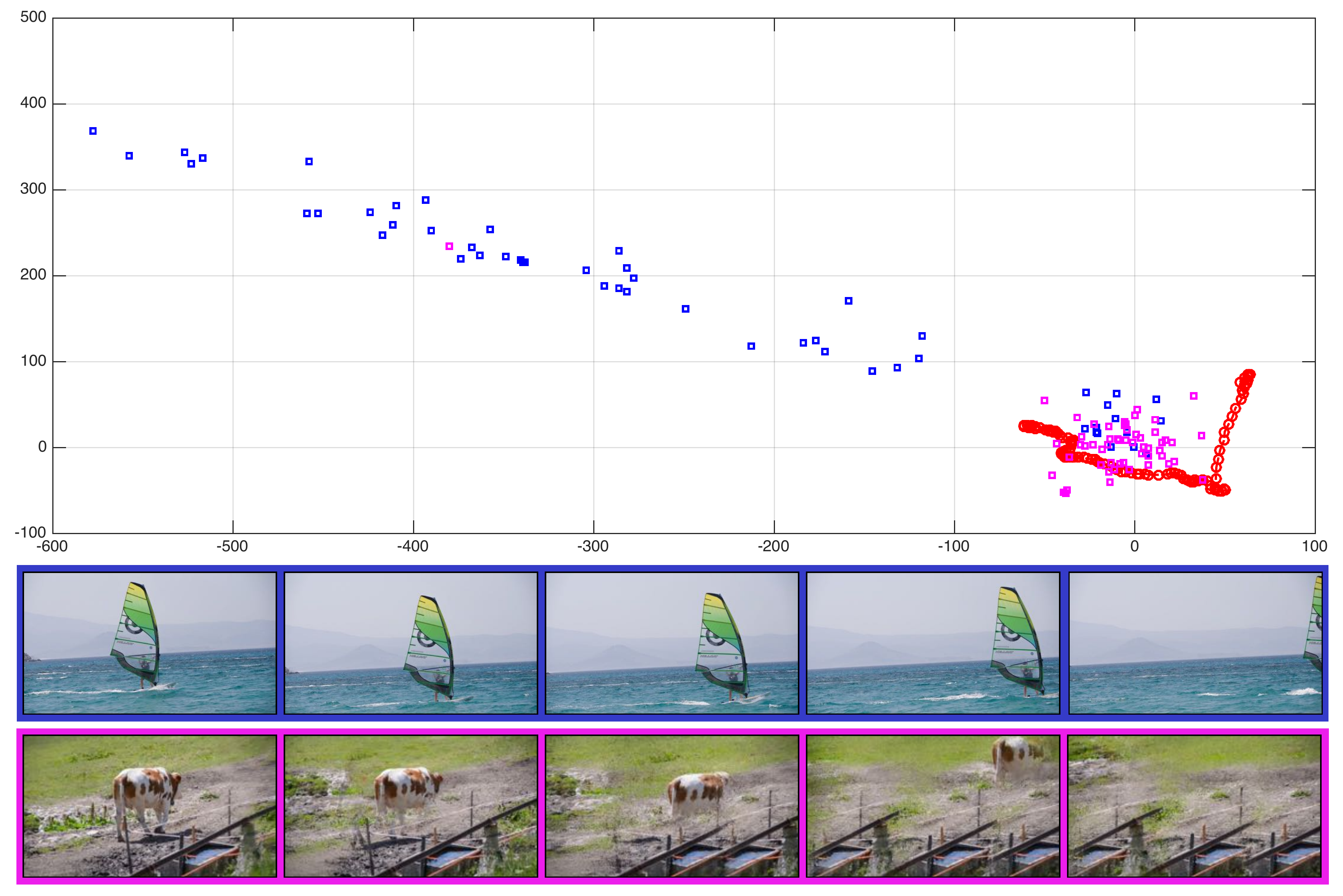}
\includegraphics[width=0.87\linewidth]{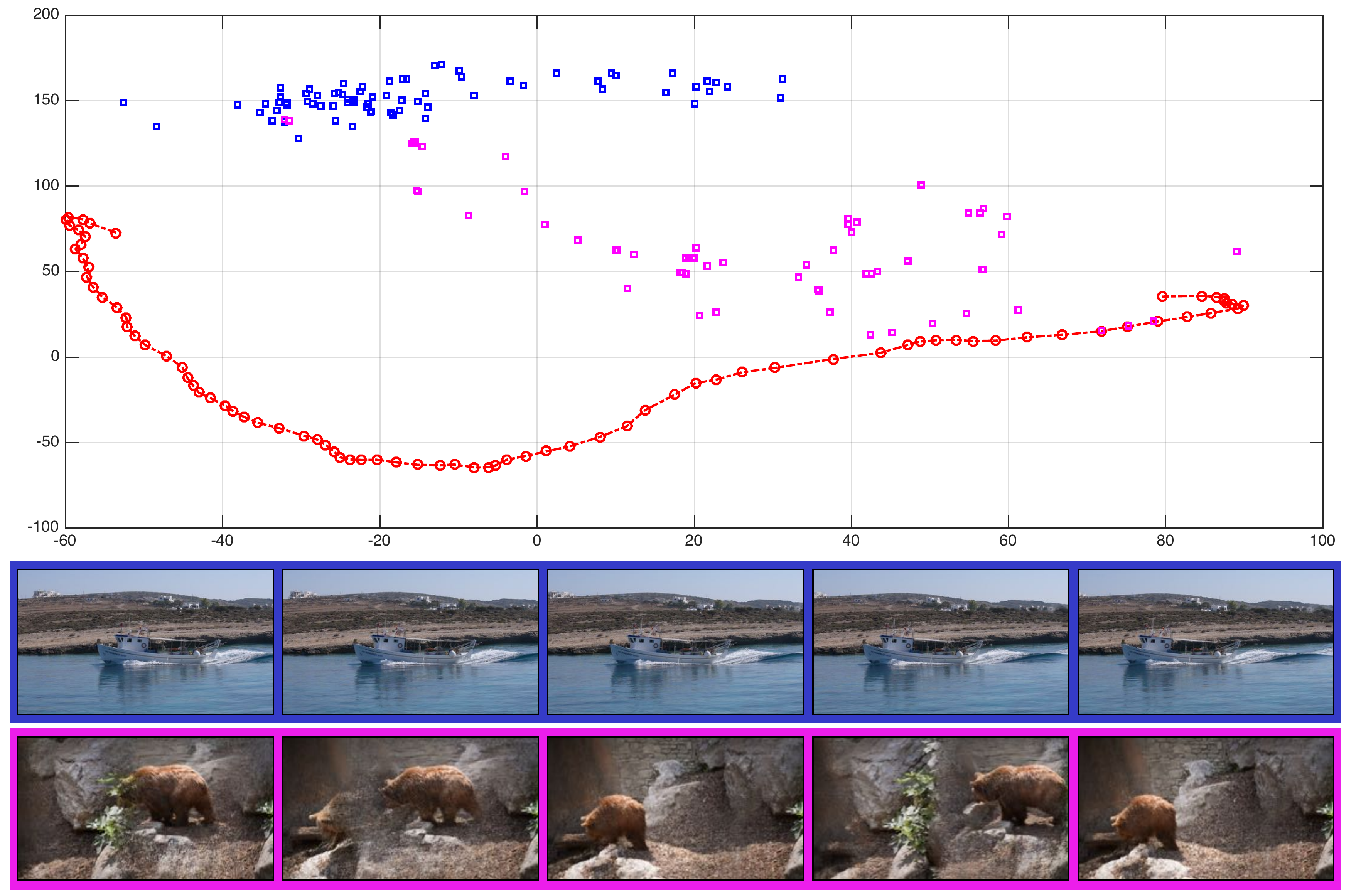}
\caption{\textbf{Extreme Perturbation via Random Videos: } We train a video-specific autoencoder using a cow sequence. The 2D visualization of this sequence is shown with the {\color{red}{red points}}. Once trained, we take inputs from a completely different video. For e.g., a surfing event in top example. The {\color{blue}{blue points}} show the projection of this video in the first iteration, and yields noisy output. Iteratively projecting these inputs bring them close to the original points. Here we show the results of $51^{st}$ iteration using {\color{magenta}{magenta points}}. We also show a few example inputs from the sequence and the reconstructed images (after $51^{st}$ iteration). The outputs are noisy and does not have a temporal coherence.}
\label{fig:diff-vids-pca-01}
\end{figure*}

\begin{figure*}
\centering
\includegraphics[width=0.9\linewidth]{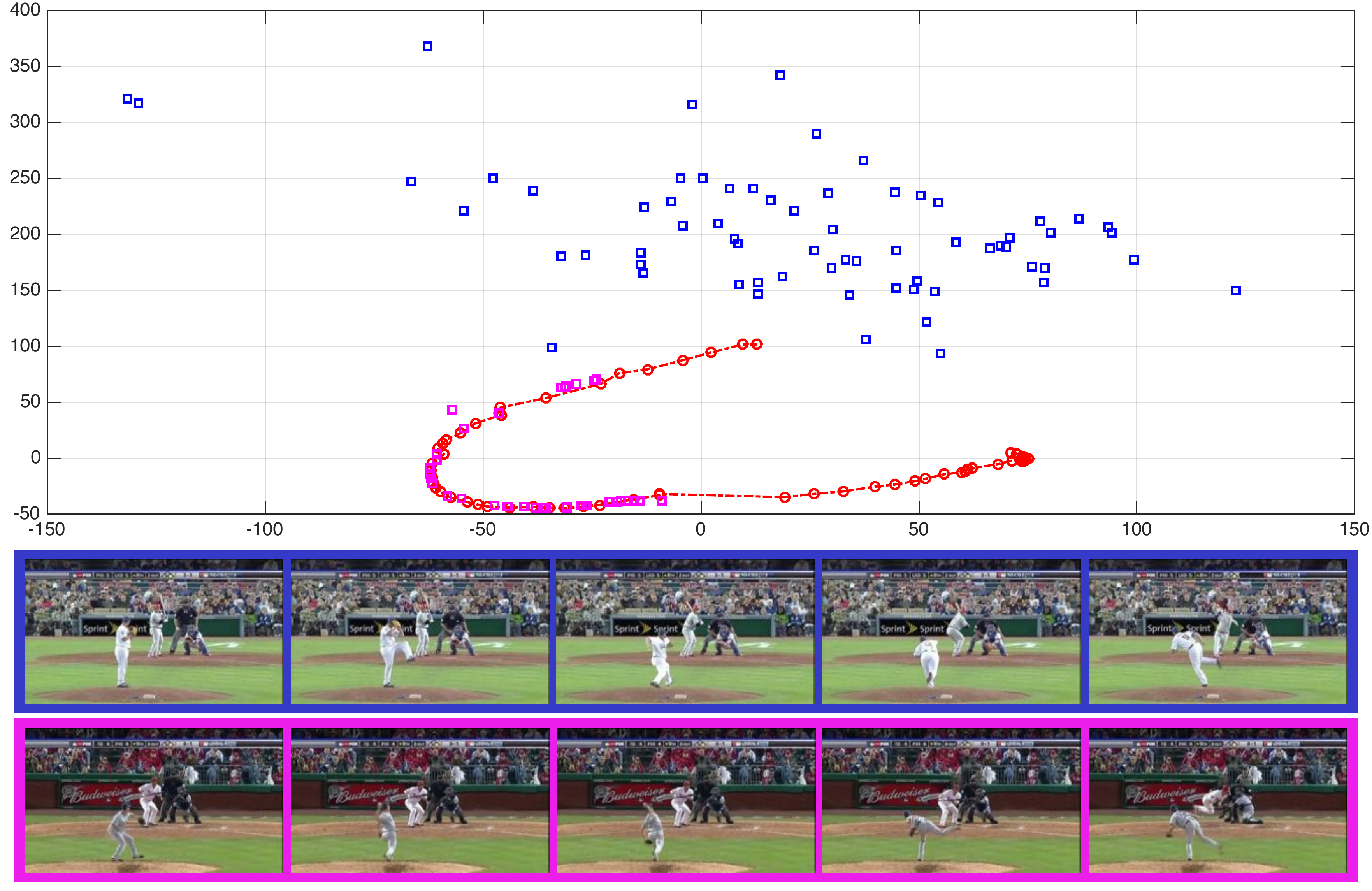}
\includegraphics[width=0.9\linewidth]{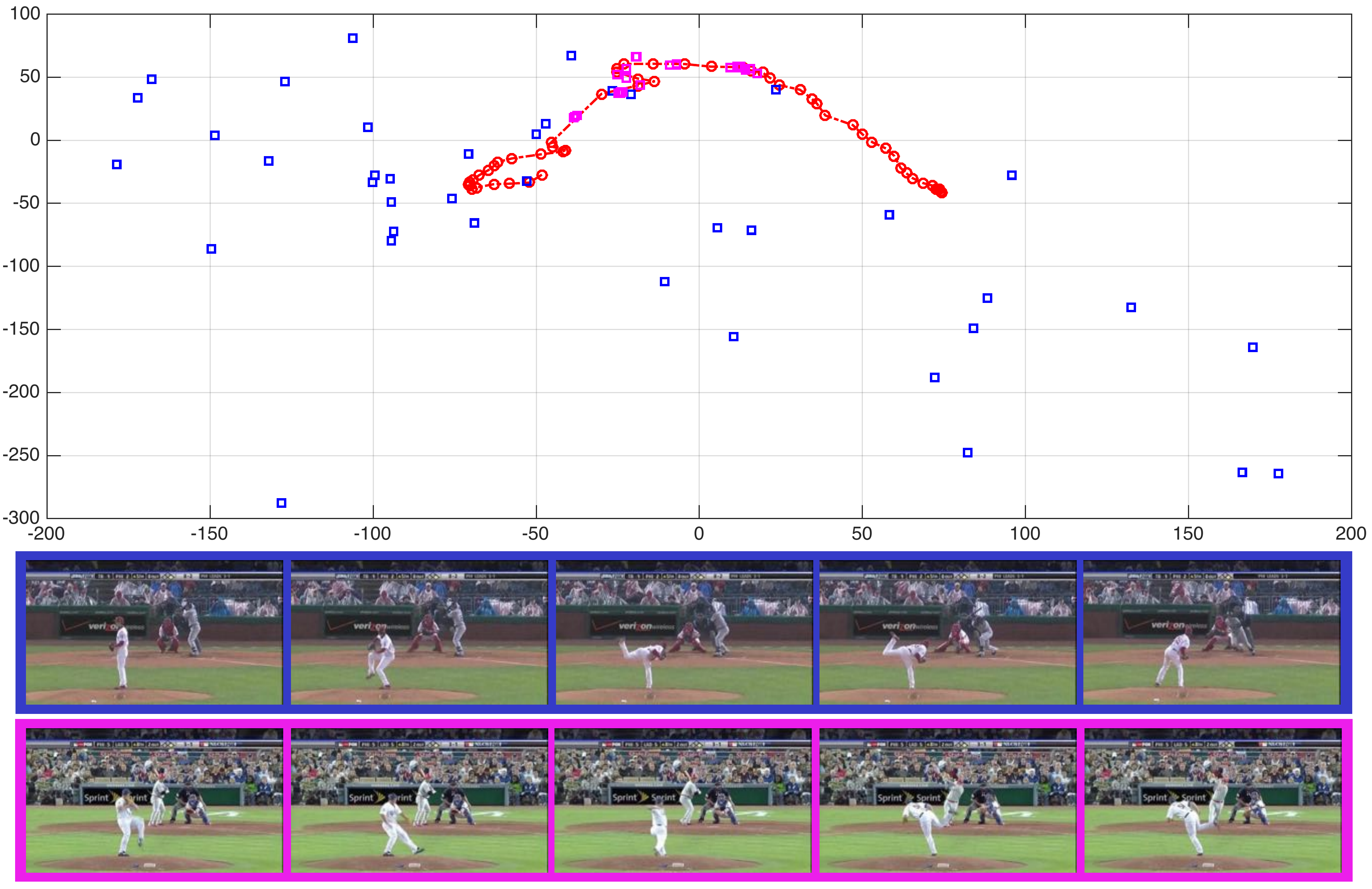}
\caption{\textbf{Frame Correspondence across Videos: } Given a video-specific manifold for a baseball video ({\color{red}{red points}}), we iteratively reproject the unknown video ({\color{blue}{blue points}}) to the manifold (shown using {\color{magenta}{magenta points}}). This allow us to get corresponding frames in two videos via cosine similarity. We also show reconstructed images from the frames of unknown video using the video-specific autoencoder. }
\label{fig:fc-01}
\end{figure*}

\begin{figure*}
\centering
\includegraphics[width=0.8\linewidth]{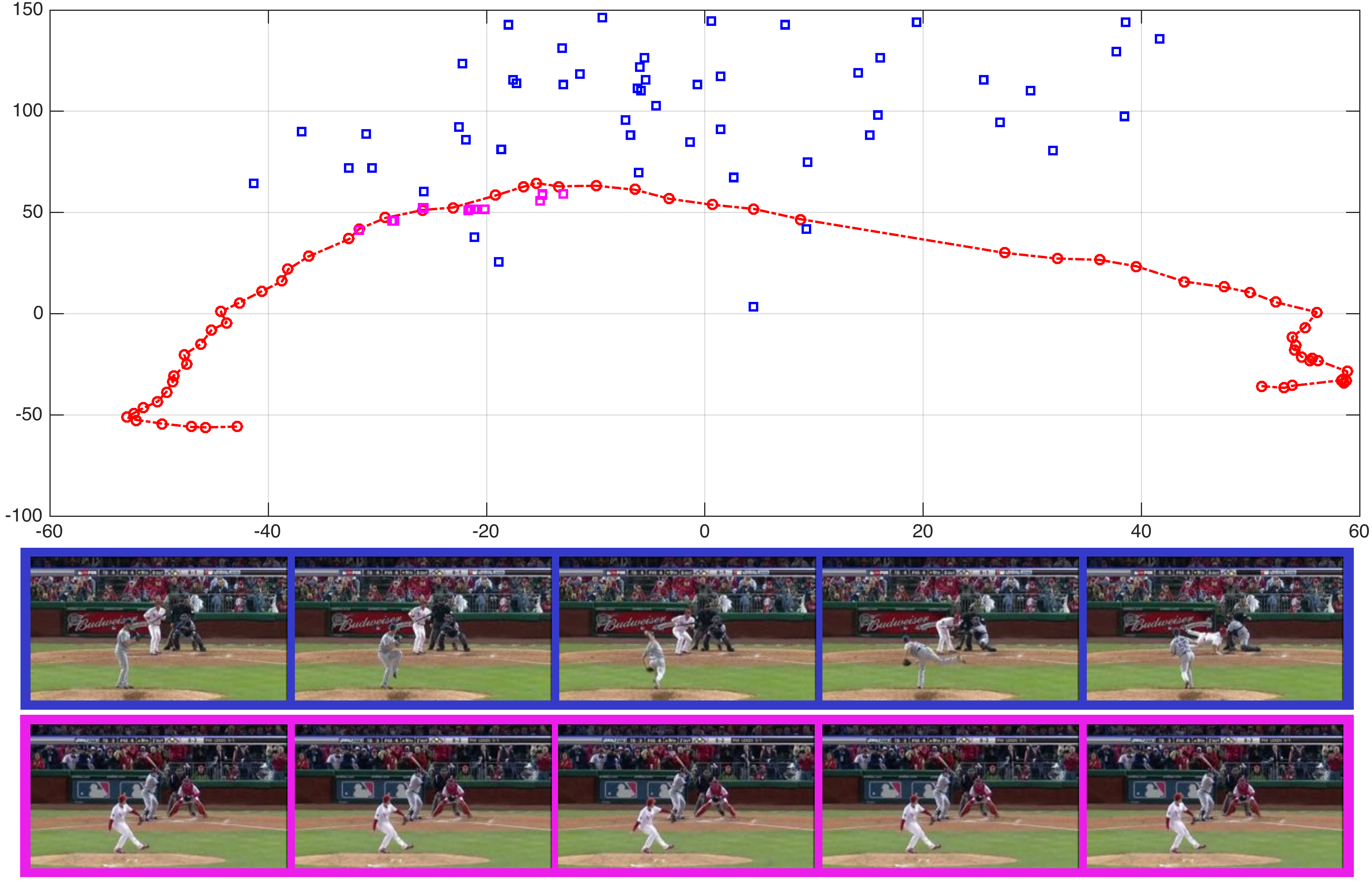}
\caption{\textbf{Failure Case of Frame Correspondence: } We observe mapping ({\color{blue}{blue points}} to {\color{magenta}{magenta points}}) collapse to a few similar points. This means there are no guarantees that we can align two videos using iterative reprojection property. However, one can see the success or failure of frame correspondences via 2D visualization of the latent codes. }
\label{fig:fc-02}
\end{figure*}

\noindent\textbf{Perturbation via Unknown Data-Distribution: } Prior work on audio conversion via exemplar autoencoders~\cite{deng2021unsupervised} showed that one can input an unknown voice sample from a different person (other than training sample) as input and yet be able to get a consistent output. We study if this property holds for the video-specific autoencoders. This property could potentially allow us to establish correspondences across the frames of two videos and do video retargeting~\cite{Recycle-GAN}. We study this behaviour via three controlled experiments: (1) \textbf{multi-view videos}: training a video-specific autoencoder on one stationary camera from a multi-view sequence~\cite{BansalCVPR2020}, and test it on other cameras. The multi-view sequences via stationary cameras allow us to study the role of slight perturbation. Figure~\ref{fig:multi-view-pca} shows the analysis of perturbing data distribution via multi-views. We observe that the points are farther away from the original points (sequence used for training) as we move away from them. This means we cannot naively use a video-specific autoencoder for the inputs that largely vary from original points; (2) \textbf{semantically similar videos}: training a video-specific autoencoder on one baseball game and test it on other baseball games~\cite{zhang2013actemes}. The game videos allow us to study the role of semantic perturbation. Figure~\ref{fig:sem-pca-01} shows the reprojection of various semanitically similar events. We observe that the points move farther as the input becomes less similar. We also observe that we can iteratively bring the points close to the original points by iterative reprojection. After a few iterations, we observe that two semantically similar events align with each other as shown in Figure~\ref{fig:sem-pca-02}. We also get temporally coherent outputs showing alignment between two videos. This property allows us to establish correspondences between two semantically similar videos and do video retargeting; and (3) finally using \textbf{completely different videos} from DAVIS dataset~\cite{Pont-Tuset_arXiv_2017} (for e.g., a model trained on bear sequence and tested it with a surfing event). The completely different videos allow us to see if the autoencoder can learn a reasonable pattern between two videos (for e.g., movement of objects in a similar direction) or leads to indecipherable random projections. We show two examples in  Figure~\ref{fig:diff-vids-pca-01}. In the first example, we train a video-specific autoencoder using a cow sequence. We input the frames of a surfing event to this trained model. The first iteration yields a noisy outputs and far-away from the original points in the video-specific manifold. We reproject the input iteratively multiple times, thereby bringing it close to the original points. We show the results of $51^{st}$ iteration. We also show a few examples showing the mapping from the frames of surfing event and the corresponding reconstructed frames. The outputs are noisy and does not have a temporal coherence. We observe similar behavior in other example.

\begin{figure*}[t]
\centering
\includegraphics[width=0.72\linewidth]{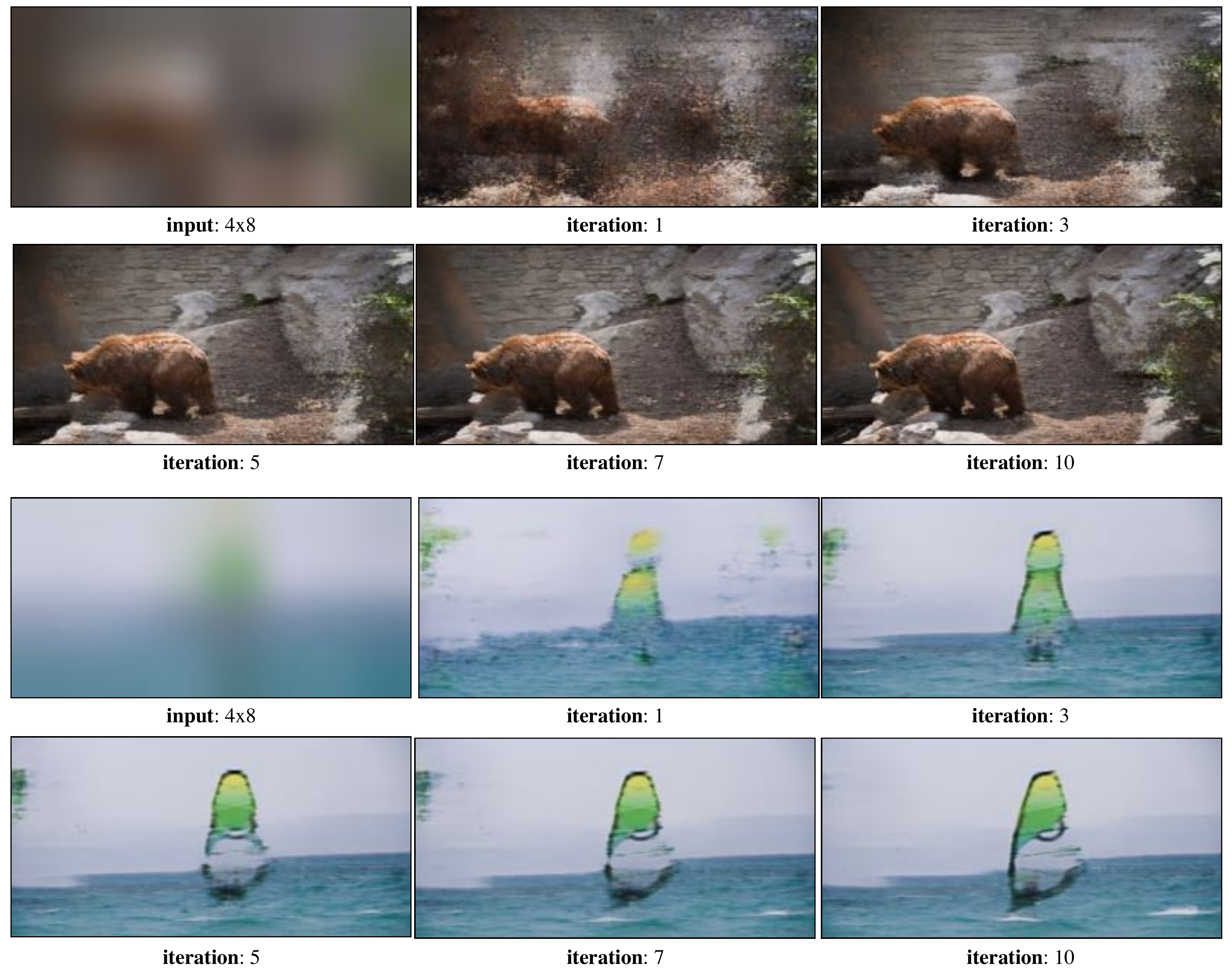}
\caption{\textbf{Iterative Improvement via Reprojection Property}: We input a low-res $4\times8$ image and iteratively improve the quality of outputs. The reprojection property allows us to move towards a good solution with every iteration. At the end of the tenth iteration, we observe a sharp but \emph{plausible} hi-res ($256\times512$) output. However, it may not be an actual solution.}
\label{fig:sres-analysis-04}
\end{figure*}

\begin{figure*}[t]
\centering
\includegraphics[width=0.9\linewidth]{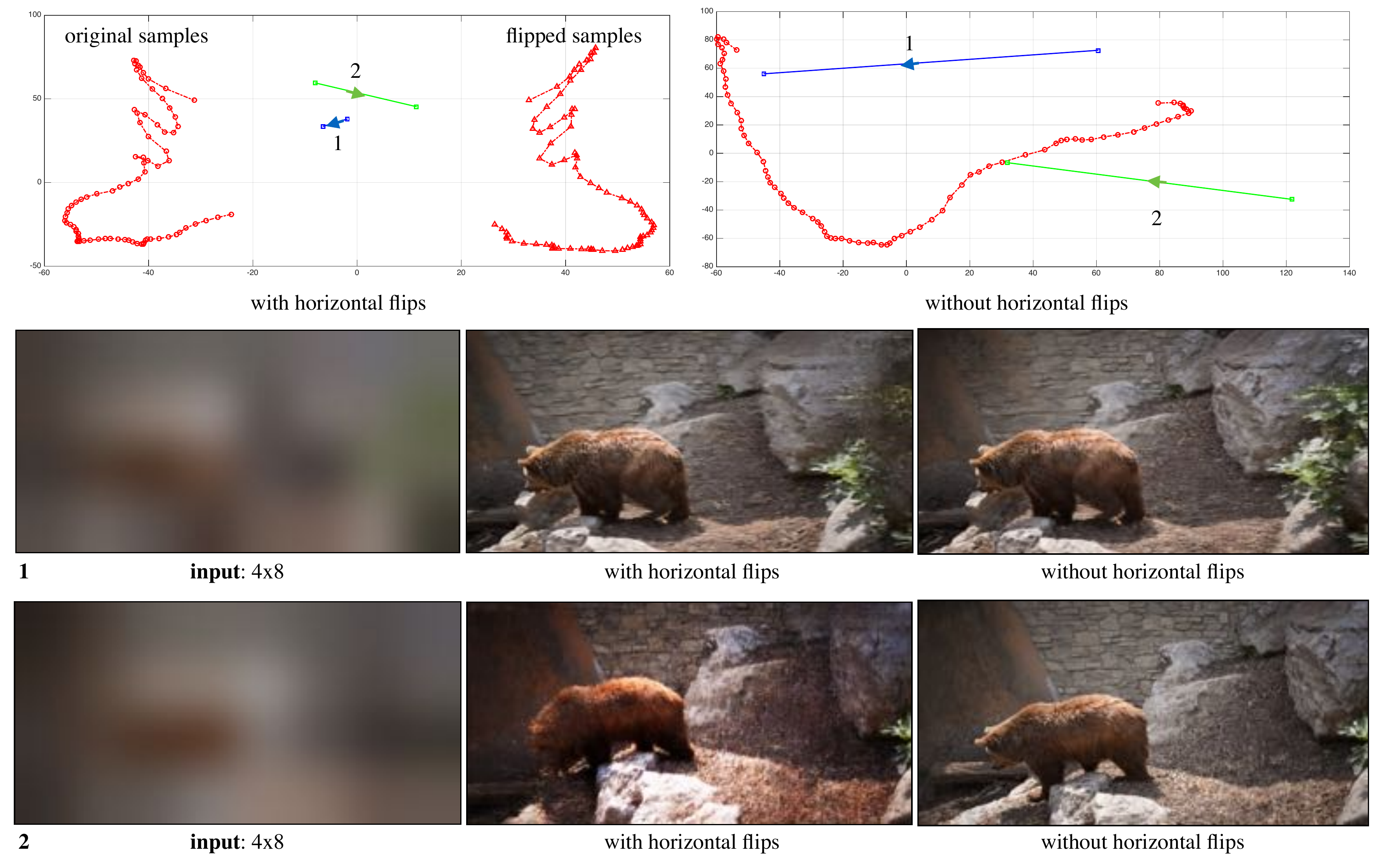}
\includegraphics[width=0.9\linewidth]{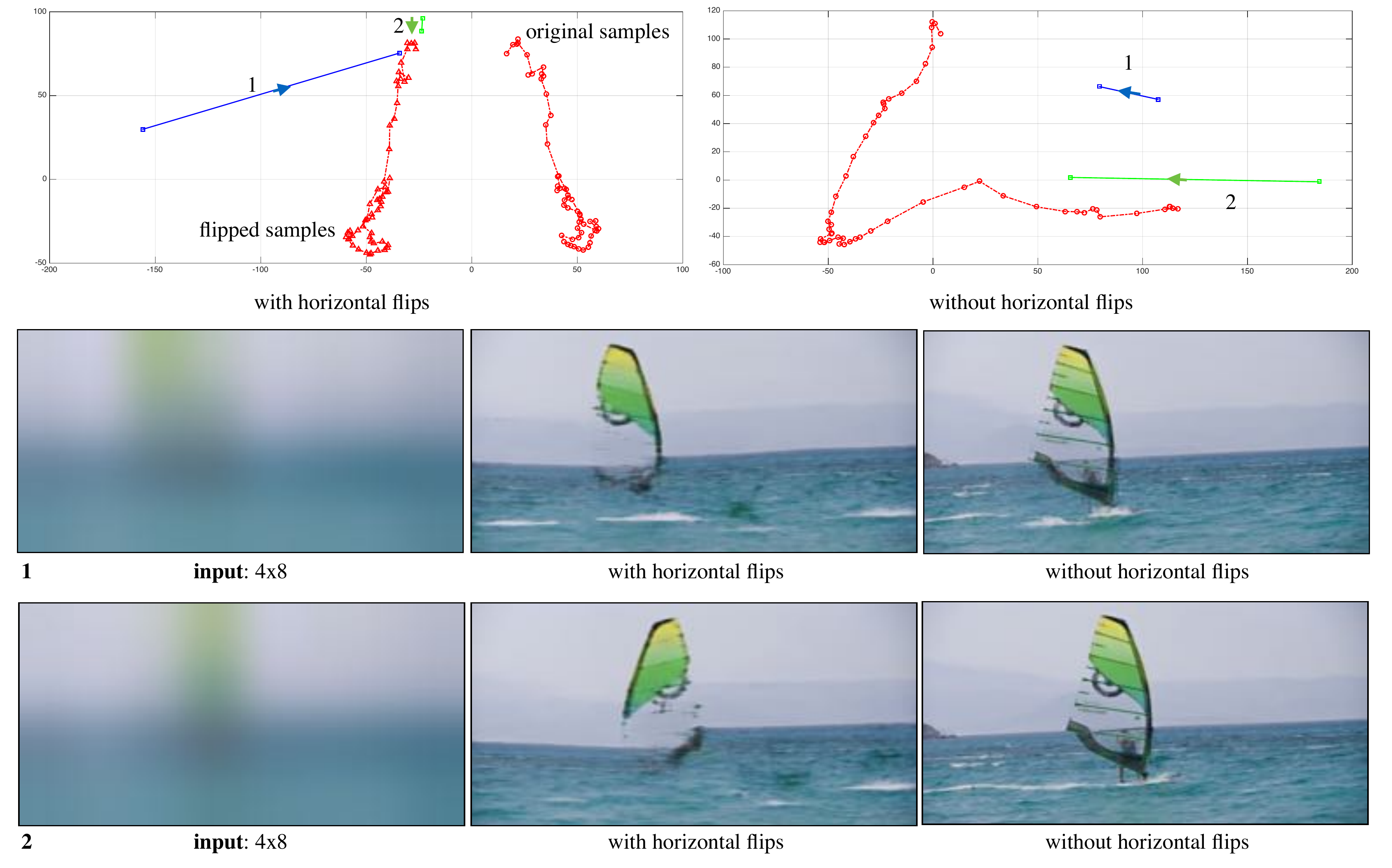}
\caption{\textbf{Influence of random horizontal flips during training on video manifold}: We study the influence of random horizontal flips when training a video-specific autoencoder. We train two autoencoder, one with random horizontal flips and other without horizontal flips. The video manifold of an autoencoder with horizontal flips learns separate spaces for original and flipped samples. This is, however, not true for the model trained without random horizontal flips. We input a low-res $4\times8$ image and iteratively improve the quality of outputs. Due to two separate spaces in the first model, the autoencoder is confused in which direction to move the noisy input sample and thereby leads to slow movement. The other autoencoder is, however, able to move quickly towards a good solution. We observe sharp $256\times512$ output at the end of tenth iteration.}
\label{fig:sres-analysis-05}
\end{figure*}

\noindent\textbf{Negative Influence of Data Augmentation: } We observe an interesting phenomenon in the surfing example shown in Figure~\ref{fig:sres-analysis-04}. The direction of final surfing output is horizontally flipped. We realize that we train the video-specific autoencoder using random horizontal flip that is a standard data augmentation strategy for training deep neural networks. However, when training a model using random horizontal flips creates distinct video-specific manifolds for both original and flipped samples as shown in the top-row of Figure~\ref{fig:sres-analysis-05}. We observe that a model trained with data augmentation confuses a noisy input as to which direction it should move to obtain a hi-res output (shown by the direction of two points in the plot). However, we are able to overcome this issue when a model is trained without random horizontal flips. We also see sharper results when using the model trained without horizontal flips. We observe that autoencoder maps noisy input to the manifold spanned by flipped samples. This is a reason why we see output similar to flipped samples. Additionally, the output of the model trained with horizontal flips suffer averaging artifacts whereas we get good sharp results when not using it. This behaviour specifically holds when the input is quite noisy (e.g. $4\times8$ resolution).

{\small
\bibliographystyle{ieee_fullname}
\bibliography{bibliography}
}

\end{document}